\documentclass[preprint]{elsarticle}

\usepackage{lineno,hyperref}
\modulolinenumbers[5]

\journal{Journal of Computer Vision and Image Understanding}

\usepackage{amsmath,amsbsy,amsfonts,amssymb,amsthm,dsfont,units}
\usepackage{xspace}
\usepackage{graphicx}
\usepackage{algorithm,algorithmic,mathtools}
\usepackage{color,cases}
\usepackage{tikz}
\usetikzlibrary{calc,shapes}
\usepackage{subfigure}
\usepackage{multirow}
\usepackage{makecell}
\usepackage{rotating}

\usepackage{hyperref}
\hypersetup{
colorlinks = true,
citecolor = {blue},
citebordercolor = {white}
}

\usepackage[T1]{fontenc}
\usepackage[utf8x]{inputenc}
\usepackage{units}
\usepackage{amsmath}
\usepackage{amsthm}
\usepackage{graphicx}
\usepackage{esint}
\usepackage{bm}
\usepackage{algorithm}
\usepackage{defs_151115}
\usepackage{algorithmic}
\usepackage{amsthm}
\usepackage{xargs}[2008/03/08]
\usepackage{amsfonts}       % blackboard math symbols
\usepackage{microtype}      % microtypography
\newcommand{\cmark}{\ding{51}}%

\newcommand{\mE}{\mathbb{E}}

\usepackage{relsize}

\newcommand{\figdir}{./pics}

\usepackage{amsmath, amsthm, amssymb}

\usepackage{pifont}
\newcommand{\etal}{\mbox{\emph{et al.\ }}}

\usepackage{url}% simple URL typesetting
\usepackage{amsfonts}% blackboard math symbols
\usepackage{nicefrac}% compact symbols for 1/2, etc.
\usepackage{microtype}% microtypography

\usepackage{graphicx}

\usepackage{colortbl} 
\definecolor{header_color}{rgb}{0.74,0.88,0.91}
\definecolor{even_color}{rgb}{0.9,0.9,0.9}
\definecolor{subheader_color}{rgb}{0.85,0.93,0.95}
\definecolor{childheader_color}{rgb}{1.0,0.93,0.87}

\definecolor{ccolor_best}{rgb}{1.0,0.9,0.9}
\definecolor{ccolor_wrong}{rgb}{1.0,0.85,0.85}

\usepackage{caption}

\usepackage{thmtools}

\declaretheoremstyle[%
  spaceabove=-6pt,%
  spacebelow=6pt,%
  headfont=\normalfont\itshape,%
  postheadspace=1em,%
  qed=\qedsymbol%
]{mystyle}

\global\long\def\expect{\mathbb{E}}

\global\long\def\given{\mid}

\global\long\def\gv{\given}

\global\long\def\vt#1{\mathbf{#1}}

\global\long\def\N{\mathrm{N}}

\global\long\def\bx{\mathbf{\mathbf{x}}}

\global\long\def\bx{\mathbf{x}}

\global\long\def\bz{\mathbf{z}}

\global\long\def\y{\mathrm{y}}

\global\long\def\by{\boldsymbol{y}}

\global\long\def\bw{\vt w}

\global\long\def\and{\cap}

\global\long\def\ess{\mathbb{E}}

\newcommandx\ESS[2][usedefault, addprefix=\global, 1=]{\underset{#2}{\ess}\left[#1\right]}

\vspace{-2mm}

\def\ie{\emph{i.e.~}}
\def\eg{\emph{e.g.~}}
\def\etal{{\em et al.~}}
\def\aka{{\em a.k.a~}}

\begin{document}

\begin{frontmatter}

\title{Pros and Cons of GAN Evaluation Measures}

%\author{Shervin Ardeshir}%\fnref{myfootnote}}
%\address{ardeshir@cs.ucf.edu}
\author{Ali Borji}%\fnref{myfootnote}}
\address{aliborji@gmail.com}



%\address{Center for Research in Computer Vision (CRCV)\\ University of Central Florida (UCF)\\ Orlando, FL, USA.}

\begin{abstract}
Generative models, in particular generative adversarial networks (GANs), have gained significant attention in recent years. A number of GAN variants have been proposed and have been utilized in many applications. Despite large strides in terms of theoretical progress, evaluating and comparing GANs remains a daunting task. While several measures have been introduced, as of yet, there is no consensus as to which measure best captures strengths and limitations of models and should be used for fair model comparison. As in other areas of computer vision and machine learning, it is critical to settle on one or few good measures to steer the progress in this field. In this paper, I review and critically discuss more than 24 quantitative and 5 qualitative measures for evaluating generative models with a particular emphasis on GAN-derived models. I also provide a set of 7 desiderata followed by an evaluation of whether a given measure or a family of measures is compatible with them. 
\end{abstract}

\begin{keyword}
Generative Adversarial Nets, Generative Models, Evaluation, Deep Learning, Neural Networks
\end{keyword}

\end{frontmatter}

%\linenumbers

\section{Introduction}

Generative models are a fundamental component of a variety of important machine learning and computer vision algorithms. They are increasingly used to estimate the underlying statistical structure of high dimensional signals
%distribution of a dataset
and artificially generate various kinds of data including high-quality images, videos, and audio. 
%These models are used to estimate the underlying distribution of a dataset and randomly generate realistic samples according to their estimated distribution.] 
They can be utilized for purposes such as 
%unsupervised
representation learning and semi-supervised learning~\cite{radford2015unsupervised,odena2016conditional,salimans2016improved}, 
%semi-supervised learning~\cite{}, feature learning~\cite{}, 
domain adaptation~\cite{ganin2016domain,tzeng2017adversarial}, text to image synthesis~\cite{reed2016generative}, compression~\cite{theis2017lossy}, 
%density estimation~\cite{}
%denoising~\cite{}, 
super resolution~\cite{ledig2016photo}, inpainting~\cite{pathak2016context,yeh2017semantic}, saliency prediction~\cite{pan2017salgan}, image enhancement~\cite{zhang2017image}, style transfer and texture synthesis~\cite{gatys2016image,johnson2016perceptual}, image-to-image translation~\cite{isola2017image,zhu2017unpaired}, and video generation and prediction~\cite{vondrick2016generating}. A recent class of generative models known as Generative Adversarial Networks (GANs) by Goodfellow \etal~\cite{goodfellow2014generative} has attracted much attention. A sizable volume of follow-up papers have been published since introduction of GANs in 2014. There has been substantial progress in terms of theory and applications and a large number of GAN variants have been introduced. However, relatively less effort has been spent in evaluating GANs and grounded ways to quantitatively and qualitatively assess them are still missing. % remains to be a serious issue. %While there have been numerous works on the subject, a remaining problem is a grounded way to compare GANs. 

%These include feature compression, image synthesis and completion, , to name a few. Lots of applications including 

%Probabilistic generative models can be used for 
%and other tasks. <Style transfer ; etc> <See xx for a review of GAN models!> 

%Despite this substantial progress, quantitative model assessment remains to be a serious issue. % in the machine learning and computer vision communities.

%A certain class of generative models known as 

%[[from https://arxiv.org/pdf/1705.05263.pdf
%There has been a substantial recent interest in deep generative models. 
Generative models can be classified into two broad categories of \textit{explicit} and \textit{implicit} approaches. The former class assumes access to the model likelihood function, whereas the latter uses a sampling mechanism to generate data. %, but do not need to explicitly define a likelihood function. 
Examples of explicit models are variational auto-encoders (VAEs)~\cite{kingma2013auto,kingma2014semi} and PixelCNN~\cite{van2016conditional}. Examples of implicit generative models are GANs. Explicit models are typically trained by maximizing the likelihood or its lower bound. 	
% they are popular in probabilistic modeling as the training procedure optimizes a well-defined quantity and the likelihood can be used for model comparison and selection. However likelihood is not about human perception (Theis et al., 2015). ]] 
GANs aim to approximate a data distribution $P$, using a parameterized model distribution $Q$. They achieve this by jointly optimizing two adversarial networks: a generator and a discriminator. The generator $G$ is trained to synthesize from a noise vector an image that is close to the true data distribution. The discriminator $D$ is optimized to accurately distinguish between the synthesized images coming from the generator and the real images from the data distribution. 
GANs have shown a dramatic ability to generate realistic high resolution images.
%sampled images.

%[[Generative adversarial networks (GANs) are a class of generative models that use neural networks for realistic data generation Goodfellow et al. (2014). 

%[In the evaluation of image quality (e.g., reality and resolution),how to design a reliable metric for generative models has been an open issue. Existing metrics (e.g., inception score [20] and related methods [15]), although successful in certain cases, have been demonstrated to be problematic in others [12]. 
%If a perfect metric exists, the training of generative models would be much easier because we could use such metric as loss directly without training a discriminator.]

Several evaluation measures have surfaced with the emergence of new models. Some of them attempt to quantitatively evaluate models while some others emphasize on qualitative ways such as user studies or analyzing internals of models.
%showing hand/cherry-picked samples. 
Both of these approaches have strengths and limitations. For example, one may think that fooling a person in distinguishing generated images from real ones can be the ultimate test. Such a measure, however, may favor models that concentrate on limited sections of the data (\ie~overfitting or memorizing; low diversity; mode dropping). Quantitative measures, while being less subjective, may not directly correspond to how humans perceive and judge generated images. These, along with other issues such as the variety of probability criteria and the lack of a perceptually meaningful image similarity measures, have made evaluating generative models notoriously difficult~\cite{theis2015note}. In spite of no agreement regarding the best GAN evolution measure, few works have already started to benchmark GANs (\eg~\cite{lucic2017gans,kurach2018gan,shmelkov2018good}). While such studies are indeed helpful, further research is needed to understand GAN evaluation measures and assess their strengths and limitations (\eg~\cite{theis2015note,huang2018an,arora2017gans,chen2017metrics,wu2016quantitative,anonymous2018an}). 

%Currently, comparisons are largely qualitative and based upon cherry-picked samples from a GAN. Cherry-picked examples emphasize small sections of the distribution learned by a GAN, while a more meaningful evaluation would utilize entire GAN distributions. Theis et al. (2015) describe many of the difficulties with evaluating generative models.  Not much work is being done on evlauting models or metric in this field. Although some have already started working on the benchmarking GANs [Google].> Its better to have an idea of which scores are better before conducting comprehensive benchmarking in this field. some comparison works ~\cite{wu2016quantitative}

%We are interested in approximating a natural image distribution not approximating a handful of points from said distribution.]]

%Due to their many uses, evaluating and comparing generative models is a problem-specific task (Theis et al., 2015). ]] 

%Despite this, quantitative model assessment remains to be a serious issue in the machine learning and computer vision communities. [GANs are moderate harder to evaluate than other generative models because it can be difficult to estimate the likelihood for GANs]. 

%ON THE QUANTITATIVE ANALYSIS OF DECODERBASED GENERATIVE MODELS

%Learning from Simulated and Unsupervised Images through Adversarial Training~\cite{shrivastava2017learning}
%~\cite{makhzani2017pixelgan}

%report.pdf An Empirical Study of Generative Adversarial Networks for Computer Vision Tasks
%[a nice overview of GANs]

My main goal in this paper is to critically review available GAN measures and help the researchers objectively assess them. At the end, I will offer some suggestions for designing more efficient measures for fair GAN evaluation and comparison.% models in better ways. %in the hope to create consensus in models evaluation.% talk about the main goals, etc here. Maybe I can explain it based on Ming Liu's tutorial!! Both quantitative and qualitative. % list some benchmark works.

\section{GAN Evaluation Measures}

I will enumerate the GAN evaluation measures while discussing their pros and cons. They will be organized in two categories: \textit{quantitative} and \textit{qualitative}. Notice that some of these measures (\eg~Wasserstein distance, reconstruction error or SSIM) can also be used for model optimization during training. In the next subsection, I will first provide a set of desired properties for GAN measures (\aka  meta measures or desiderata) followed by an evaluation of whether a given measure or a family of measures is compatible with them. 

Table~\ref{tab:sum} shows the list of measures. Majority of the measures return a single value while few (GAM~\cite{im2016generating} and NRDS~\cite{zhang2018decoupled}) perform relative comparison. The rationale behind the latter is that if it is difficult to obtain the perfect measure, at least we can evaluate which model generates better images than others.

%[[ The current submission seems to mix these scopes and conflates criticism of a given setting with criticism of a given approach. ]]

%Tell them first I introduce these properties and then eventually  an evaluation of whether a given approach or family of approaches is compatible with them. in the discussion section 

%[[Instead, a list of properties / disiderata should be enumerated followed by an evaluation of whether a given approach or family of approaches is compatible with them.]]

%Notice that while I tried to include as many measures as possible, it is possible that I have missed some. %they might be some others that are missing. % Here I focus on the most popular measures used in several papers.

%Each paper adopts his own measure! which is not good! We need standards! Some common ones include KL and Inception score!! ]]

\subsection{Desiderata}
\label{desiderata}

Before delving into the explanation of evaluation measures, first I list a number of desired properties that an efficient GAN evaluation measure should fulfill. These properties can serve as meta measures to evaluate and compare the GAN evaluation measures. Here, I emphasize on the qualitative aspects of these measures. As will be discussed in Section~\ref{disc}, some recent works have attempted to compare the meta measures quantitatively (\eg computational complexity of a measure). 
%Such an assessment can be done both quantitative and qualitative. While some works have attempted the former, here %I provide a qualitative comparison of meta measures. Based on previous studies, 
An efficient GAN evaluation measure should:

%[[From the above discussions, 
%From FusedGAN
%A few necessary conditions for a meaningful GAN evaluation metric are as follows:

\begin{enumerate}
\item favor models that generate high fidelity samples (\ie ability to distinguish generated samples from real ones; discriminability),
\item favor models that generate diverse samples (and thus is sensitive to overfitting, mode collapse and mode drop, and can undermine trivial models such as the memory GAN),
%\item be able to identify mode drop and mode collapse, and detect overfitting (i.e., generate diverse samples), 
\item favor models with disentangled latent spaces as well as space continuity (\aka controllable sampling),
\item have well-defined bounds (lower, upper, and chance),
%\item undermine trivial models such as the memory GAN,
\item be sensitive to image distortions and transformations. % (See Figs.~\ref{fig:robustness} and~\ref{fig:FID}),
GANs are often applied to image datasets where certain transformations to the input do not change semantic meanings. Thus, an ideal measure should be invariant to such transformations. For instance, score of a generator trained on CelebA face dataset should not change much if its generated faces are shifted by a few pixels or rotated by a small angle. 
\item agree with human perceptual judgments and human rankings of models, and 
\item have low sample and computational complexity.
\end{enumerate}

In what follows, GAN measures will be discussed and assessed with respect to the above desiderata, and a summary will be presented eventually in Section~\ref{disc}. See Table~\ref{tab:meta}.

\begin{figure}[t]
\includegraphics[width=\linewidth]{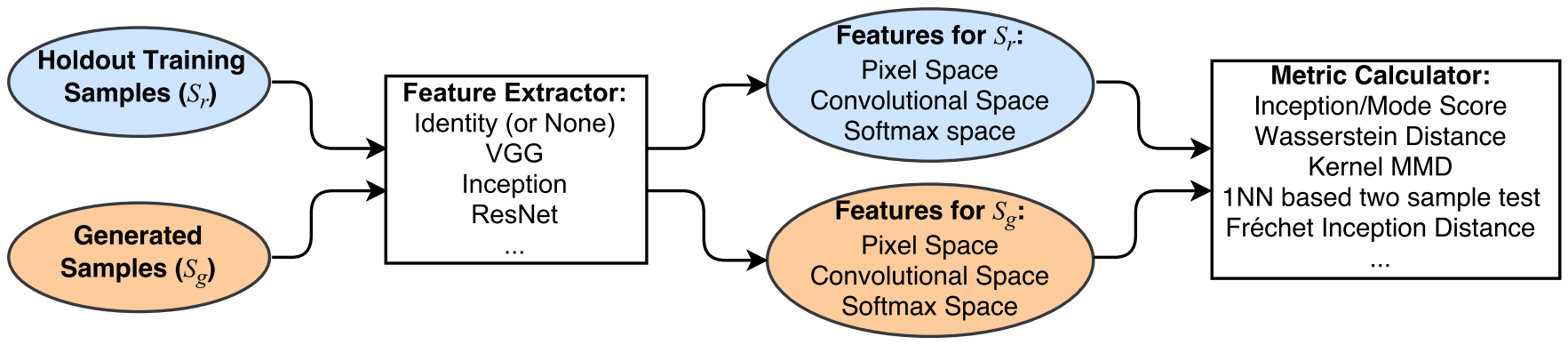}
\caption{A schematic layout of the typical approach for sample based GAN evaluation. $S_r$ and $S_g$ represent real and generated samples, respectively. Figure from~\cite{huang2018an}. } 
\label{fig:layout}
\end{figure}

\subsection{Quantitative Measures}

A schematic layout for sample based GAN evaluation measures is shown in Fig.~\ref{fig:layout}. Some measures discussed in the following are “model agnostic'' in that the generator is used as a black box to sample images and they do not require a density estimation from the model. On the contrary, some other measures such as average log-likelihood demand estimating a probability distribution from samples.

%We briefly describe xx measures. Some regard x while some others regard y. both quantitatively and qualitatively.

\begin{sidewaystable}
%\begin{table}
\begin{center}
\hspace{-30pt}
\renewcommand{\tabcolsep}{.7mm}
\renewcommand{\arraystretch}{1.1}%

\begin{footnotesize}
\begin{tabular}{| lll |} 
\hline
{\bf } & \multicolumn{1}{c}{\bf Measure}  &  \multicolumn{1}{c|}{\bf Description}  \\%& {\bf Pros} & {\bf Cons}  \\
\hline
\hline
\multirow{35}{*}{\rotatebox{90}{\bf Quantitative}} & 1. Average Log-likelihood~\cite{goodfellow2014generative,theis2015note} &  \makecell[l]{$\bullet$ Log likelihood of explaining realworld held out/test data using a density estimated from the generated data \\ (\eg~using KDE or Parzen window estimation). $L = \frac{1}{N} \sum_i \log P_{model}(\bx_i)$} \\ \cline{3-3}
& 2. Coverage Metric~\cite{tolstikhin2017adagan} &  \makecell[l]{$\bullet$ The probability mass of the true data “covered” by the model distribution \\ ${C:=P_{data}( dP_{model} > t)}$ with $t$ such that $P_{model}(dP_{model} > t) = 0.95$}  \\ \cline{3-3}
& 3. Inception Score (IS)~\cite{salimans2016improved} &  $\bullet$ KLD between conditional and marginal label distributions over generated data. $ \exp\left(\expect_{\bx}\left[\mathbb {KL}\left(p\left(\y\gv\bx\right)\parallel p\left(\y\right)\right)\right]\right) $ \\ \cline{3-3}
& 4. Modified Inception Score (m-IS)~\cite{gurumurthy2017deligan}  &  \makecell[l]{$\bullet$ Encourages diversity within images sampled from a particular category. $ \exp({\mathbb{E}_{\bx_{i}}[\mathbb{E}_{\bx_{j}}[(\mathbb {KL}(P(y|\bx_{i})||P(y|\bx_{j})) ]]})$ }  \\ \cline{3-3}
& 5. Mode Score (MS)~\cite{che2016mode}  &  \makecell[l]{$\bullet$ Similar to IS but also takes into account the prior distribution of the labels over real data. \\ $ \exp\left(\expect_{\bx}\left[\mathbb {KL}\left(p\left(y\gv\bx\right)\parallel{p}\left(y^{train}\right)\right)\right]-\mathbb {KL}\left(p\left(y\right)\parallel{p}\left(y^{train} \right)\right)\right) $ }    \\ \cline{3-3}
& 6. AM Score~\cite{zhou2018activation}  &  \makecell[l]{$\bullet$ Takes into account the KLD between distributions of training labels vs. predicted labels, \\as well as the entropy of predictions. $\mathbb {KL}( p(y^\text{train})\parallel p(y))\!+\! \mE_{\bx} \big[H(y|\bx)\big] $} \\ \cline{3-3}% but with reverse order of terms in KLD.  %\mE_{\bx} \big[\mathbb {KL} \big(p(y^\text{train}) \parallel p(y|x) \big) \big]\!-\!\mathbb {KL}( p(y^\text{train}) \parallel p(y)) $  }    

& 7. Fr\'echet Inception Distance (FID)~\cite{heusel2017gans}   & \makecell[l]{$\bullet$ Wasserstein-2 distance between multi-variate Gaussians fitted to data embedded into a feature space\\ $ FID(r,g) = || \mu_r - \mu_g||^2_2 + Tr(\Sigma_r + \Sigma_g - 2(\Sigma_r\Sigma_g)^\frac{1}{2})$} \\ \cline{3-3}
& \makecell[l]{8. Maximum Mean Discrepancy (MMD) \\ \cite{gretton2012kernel}} & \makecell[l]{$\bullet$
Measures the dissimilarity between two probability distributions $P_r$ and $P_g$ using samples drawn independently \\ from each distribution. $M_k(P_r, P_g) =  \mE_{\bx, \bx' \sim P_r}[ k(\bx,\bx') ] -2\mE_{\bx \sim P_r, \by \sim P_g}[k(\bx,\by)] + \mE_{\by,\by' \sim P_g}[ k(\by,\by')]$} \\ \cline{3-3}
%$M_k(P_r, P_g) =  \mE_{P_r}[ k(x,x') ] -2\mE_{P_r,P_g}[k(x,y)] + \mE_{P_g}[ k(y,y') ]$}  \\ \cline{3-3}
& 9. The Wasserstein Critic~\cite{arjovsky2017wasserstein} & \makecell[l]{$\bullet$ The critic (\eg~an NN) is trained to produce high values at real samples and low values at generated samples \\ $\hat{W}(\bx_\mathit{test}, \bx_g) = \frac{1}{N}\sum^{N}_{i=1} \hat{f}(\bx_\mathit{test}[i]) - \frac{1}{N}\sum^{N}_{i=1} \hat{f}(\bx_g[i]) $}  \\ \cline{3-3}
& 10. Birthday Paradox Test~\cite{arora2017gans} &  \makecell[l]{$\bullet$ Measures the support size of a discrete (continuous) distribution by counting the duplicates (near duplicates)}  \\ \cline{3-3}
& 11. Classifier Two Sample Test (C2ST)~\cite{lehmann2006testing} &  \makecell[l]{$\bullet$ Answers whether two samples are drawn from the same distribution (\eg~by training a binary classifier)}   \\ \cline{3-3}
& 12. Classification Performance~\cite{radford2015unsupervised,isola2017image}  &  \makecell[l]{$\bullet$ An indirect technique for evaluating the quality of unsupervised representations \\ (\eg~feature extraction; FCN score). See also the GAN Quality Index (GQI)~\cite{ye2018gan}.}  \\  \cline{3-3}
& 13. Boundary Distortion~\cite{santurkar2018classification}
&  \makecell[l]{$\bullet$ Measures diversity of generated samples 
and covariate shift using classification methods.}  \\  \cline{3-3}
& \makecell[l]{14. Number of Statistically-Different Bins \\ (NDB)~\cite{Richardson2018}} &  \makecell[l]{$\bullet$ Given two sets of samples from the same distribution, the number of samples that \\ fall into a given bin should be the same up to sampling noise}   \\ \cline{3-3}
& 15. Image Retrieval Performance~\cite{wang2016ensembles}  &  \makecell[l]{$\bullet$ Measures the distributions of distances to the nearest neighbors of some query images (\ie~diversity)} \\ \cline{3-3}
& \makecell[l]{16. Generative Adversarial Metric (GAM) \\ \cite{im2016generating}}  & \makecell[l]{$\bullet$ Compares two GANs by having them engaged in a battle against each other by swapping discriminators \\ or generators. $p(\bx|y = 1; M^{`}_1) / p(\bx|y = 1; M^{`}_2) =  \big(p(y = 1|\bx; D_1)p(\bx;G_2)\big) / \big(p(y = 1|\bx; D_2)p(\bx;G_1)$\big) }  \\ \cline{3-3}
& \makecell[l]{17. Tournament Win Rate and Skill \\ Rating~\cite{olsson2018skill}} 
 & \makecell[l]{$\bullet$ Implements a tournament in which a player is either a discriminator that attempts to distinguish between\\ real and fake data or a generator that attempts to fool the discriminators into accepting fake data as real.} \\ \cline{3-3}
& \makecell[l]{18. Normalized Relative Discriminative \\ Score (NRDS)~\cite{zhang2018decoupled}}  & \makecell[l]{$\bullet$ Compares $n$ GANs based on the idea that if the generated samples are closer to real ones, \\ more epochs would be needed to distinguish them from real samples.} \\ \cline{3-3} 
& \makecell[l]{19. Adversarial Accuracy and Divergence \\ \cite{yang2017lr}} &  \makecell[l]{$\bullet$ Adversarial Accuracy. Computes the classification accuracies achieved by the two classifiers, one trained \\ on real data and another on generated data, on a labeled validation set to approximate $P_g(y|\bx)$ and $P_r(y|\bx)$. \\ Adversarial Divergence: Computes $\mathbb {KL}(P_g(y|\bx), P_r(y|\bx))$ }  \\  \cline{3-3}
& 20. Geometry Score~\cite{khrulkov2018geometry}
&  \makecell[l]{$\bullet$ Compares geometrical properties of the
underlying data manifold between real and generated data.}  \\  \cline{3-3}
& 21. Reconstruction Error~\cite{xiang2017effects}  &  \makecell[l]{$\bullet$ Measures the reconstruction error (\eg~$L_2$ norm) between a test image and its closest \\ generated image by optimizing for $z$ (\ie~$min_\bz || G(\bz) - \bx^{(test)} ||^2 $) } \\ \cline{3-3}
& 22. Image Quality Measures~\cite{wang2004image,ridgeway2015learning,juefei2017gang}  &  \makecell[l]{$\bullet$ Evaluates the quality of generated images using measures such as SSIM, PSNR, and sharpness difference} \\  \cline{3-3}
& 23. Low-level Image Statistics~\cite{zeng2017statistics,karras2017progressive}  &  \makecell[l]{$\bullet$ Evaluates how similar low-level statistics of generated images are to those of natural scenes \\ in terms of mean power spectrum,  distribution of random filter responses, contrast distribution, etc.} \\ \cline{3-3}
%& Sample and Runtime Efficiency~\cite{huang2018an} &  \makecell[l]{$\bullet$ Regards the number of samples needed for a metric to discriminate a set of generated samples \\ from real ones. It also considers computational efficiency and runtime}  \\
& 24. Precision, Recall and $F_1$ score~\cite{lucic2017gans}  & \makecell[l]{$\bullet$ These measures are used to quantify the degree of overfitting in GANs, often over toy datasets.} \\ \cline{3-3}
\hline  
\hline

\multirow{6}{*}{\rotatebox{90}{\bf Qualitative}} & 1. Nearest Neighbors  &  \makecell[l]{$\bullet$ To detect overfitting, generated samples are shown next to their nearest neighbors in the training set} \\ \cline{3-3}
& 2. Rapid Scene Categorization~\cite{goodfellow2014generative}  &  \makecell[l]{$\bullet$ In these experiments, participants are asked to distinguish generated samples from real images \\ in a short presentation time (\eg~100 ms); \ie~real v.s fake }  \\ \cline{3-3}
& 3. Preference Judgment~\cite{huang2017stacked,zhang2017stackgan,xiao2018generating,yi2017dualgan} &  \makecell[l]{$\bullet$ Participants are asked to rank models in terms of the fidelity of their generated images (\eg~pairs, triples)}  \\ \cline{3-3}
& 4. Mode Drop and Collapse~\cite{srivastava2017veegan,lin2017pacgan} &  \makecell[l]{$\bullet$ Over datasets with known modes (\eg~a GMM or a labeled dataset), modes are computed as by measuring \\ the distances of generated data to mode centers}  \\ \cline{3-3}
& 5. Network Internals~\cite{radford2015unsupervised,chen2016infogan,higgins2016beta,mathieu2016disentangling,zeiler2014visualizing,bau2017network} &  \makecell[l]{$\bullet$ Regards exploring and illustrating the internal representation and dynamics of models (\eg~space continuity) \\ as well as visualizing learned features}  \\ 
\hline
\end{tabular}
\caption{A summary of common GAN evaluation measures.} 
%Maybe I can make a table summarizing the pros and cons of model!!
%e.g., needs labels, can it identify mode collapse? x
%How to classify the metrics?! scores? similar to saliency research?!!}
\label{tab:sum}
\end{footnotesize}
\end{center}

%\end{table}
\end{sidewaystable}

% https://openreview.net/pdf?id=Sy1f0e-R-
%All of the metrics above are, what we refer to as “model agnostic": they use the generator as a
%black box to sample the generated images $S_g$. Model agnostic metrics should not require a density
%estimation from the model. 

\begin{enumerate}

\item {\bf Average Log-likelihood.} 
% https://openreview.net/pdf?id=Sy1f0e-R-
Kernel density estimation (KDE or Parzen window estimation) is a well-established method for estimating
the density function of a distribution from samples\footnote{Each sample is a vector shown in boldface (\eg~$\bx$).}. For a probability kernel $K$ (most often an
isotropic Gaussian) and i.i.d samples $\{\bx_1, \cdots , \bx_n\}$, a density function at $\bx$ is defined as 
$p(\bx) \approx \frac{1}{z} \Sigma_{i=1}^n K(\bx - \bx_i)$, where $z$ is a normalizing constant. This allows the use of classical measures such as KLD and JSD (Jensen Shannon divergence). However, despite its widespread use, its suitability for estimating the density of GANs has been questioned by Theis \etal~\cite{theis2015note}. %  since the probability kernel depends on the Euclidean distance between images (which is very sensitive to minor image changes; see fig. xx).

%from https://arxiv.org/pdf/1702.03307.pdf   Generative Mixture of Networks
%[[from AdaGAN:  
Log-likelihood (or equivalently Kullback-Leibler divergence) has been the
de-facto standard for training and evaluating generative models~\cite{tolstikhin2017adagan}. It measures the likelihood of the true data under the generated distribution on $N$
samples from the data, \ie~$L = \frac{1}{N} \sum_i \log P_{model}(\bx_i)$.  %It is commonly used for generative models is the average log-likelihood of the test set (a.k.a Parzen estimation). 
Since estimating likelihood in higher dimensions is not feasible, generated samples can be used to infer something about a model's log-likelihood. The intuition is that a model with maximum likelihood (zero KL divergence) will produce perfect samples. 

    %Note that \cite{1611.04273} proposes a more general and elegant approach (but less straightforward to implement) to have an objective measure of GAN. On the simple problems we tackle here, we can precisely estimate the likelihood. ]]

The Parzen window approach to density estimation works by taking a finite set of samples generated by a model and then using those as the centroids of a Gaussian mixture. The constructed Parzen windows mixture is then used to compute a log-likelihood score on a set of test examples. Wu \etal~\cite{wu2016quantitative} proposed to use annealed importance sampling (AIS)~\cite{neal2001annealed} to estimate
log-likelihoods using a Gaussian observation model with a fixed variance. The key drawback of this approach is the assumption of the Gaussian observation model which may not work quite well in high-dimensional spaces. They found that AIS is two orders of magnitude more accurate than KDE, and is accurate enough for comparing generative models.

While likelihood is very intuitive, it suffers from several drawbacks~\cite{theis2015note}: %The classic approach towards evaluating generative models is based on model likelihood which is often intractable. While the log-likelihood can be approximated for distributions on low-dimensional vectors, in the context of complex high-dimensional data the task becomes extremely challenging.]] 
\begin{enumerate}
\item For a large number of samples, Parzen window estimates fall short in approximating a model's true log-likelihood when the data dimensionality is high. Even for the fairly low dimensional space of $6 \times 6$ image patches, it requires a very large number of samples to come close to the true log-likelihood of a model. See Fig.~\ref{fig:theis}.B.

\item Theis \etal showed that the likelihood is generally uninformative about the quality of samples and vice versa. In other words, log-likelihood and sample quality are moderate unrelated. A model can have poor log-likelihood and produce great samples, or have great log-likelihood and produce poor samples. An example in the former case is a mixture of Gaussian distributions where the means are training images (\ie~akin to a look-up table). Such a model will generate great samples but will still have very poor log-likelihood. An example of the latter is a mixture model combined of a good model, with a very low weight $\alpha$ (\eg~$\approx 0.01$), and a bad model with a high weight  $1 - \alpha$. Such a model has a large average log-likelihood but generates very poor samples (See~\cite{theis2015note} for the proof).
  
    %which is what GAN practitioners care about.
\item Parzen window estimates of the likelihood produce rankings different from other measures (See Fig.~\ref{fig:theis}.C). 

\end{enumerate}

Due to the above issues, it becomes difficult to answer basic questions such as whether GANs are simply memorizing training examples, or whether they are missing important modes of the data distribution. For further discussions on other drawbacks of average likelihood measures consult~\cite{huszar2015not}.

\item {\bf Coverage Metric.} %[[ which paper: AdaGAN: Boosting Generative Models
%To evaluate how well the generated distribution matches the target distribution,
%we use a \emph{coverage} metric~$C$. 
Tolstikhin \etal~\cite{tolstikhin2017adagan} proposed to use the probability mass of the real data ``covered''
by the model distribution $P_{model}$ as a metric. They compute ${C:=P_{data}( dP_{model} > t)}$ with $t$ such that
$P_{model}(dP_{model} > t) = 0.95$. A kernel density estimation
method was used to approximate the density of $P_{model}$. They claim that  %Note that we could also use the discriminator $D$ to approximate the coverage as well, using the relation from \eqref{eq:prob-ratio-and-d}. 
this metric is more interpretable than the likelihood, making it easier to assess the difference in performance of the algorithms.

\item {\bf Inception Score (IS).} Proposed by Salimans \etal~\cite{salimans2016improved}, it is perhaps the most widely adopted score for GAN evaluation (\eg~in~\cite{fedus2017many}). It uses a pre-trained neural network (the Inception Net~\cite{szegedy2016rethinking} trained on the ImageNet~\cite{deng2009imagenet}) to capture the desirable properties of generated samples: \textit{highly classifiable} and \textit{diverse} with respect to class labels. It measures the average KL divergence between the conditional label distribution $p(y|\bx)$ of samples (expected to have low entropy for easily classifiable samples; better sample quality) and the marginal distribution $p(y)$ obtained from all the samples (expected to have high entropy if all classes are equally represented in the set of samples; high diversity). It favors low entropy of $p(y|\bx)$ but a large entropy of $p(y)$. 
%This metric, proposed by \citet{improved-gans}, is based on the classification output $p(y \mid x)$ of the Inception model \citep{inception}. Defined as $\exp\left( \E_x \KL{p(y \mid x)}{p(y)} \right)$,
%it is highest when each image's predictive distribution has low entropy,
%but the marginal $p(y) = \E_x p(y \mid x)$ has high entropy.
%This score correlates moderate with human judgement of sample quality on natural images,
%but it has some issues, especially when applied to domains which do not represent a variety of the types of classes in ImageNet. In particular, it knows nothing about the desired distribution for the model.

%As the name suggests, the classifier network used to compute the inception score was originally an Inception network [37] trained on the ImageNet dataset. 

% propose an
%image-quality measure which they find to be highly correlated with human visual judgment. 
%They propose to feed the samples x of the model to the “inception” model to obtain a conditional label
%distribution p(y|x), and evaluate the score defined by 

 \begin{equation}
   \exp\left(\expect_{\bx}\left[\mathbb {KL}\left(p\left(\y\gv\bx\right)\parallel p\left(\y\right)\right)\right]\right) = \exp\left(H(y) - \expect_{\bx}\left[H(y|\bx)\right]\right),
   \label{eq:1}
\end{equation}   
   
where $p\left(\y\gv\bx\right)$ is the conditional label distribution
for image $\bx$ estimated using a pretrained Inception model \cite{szegedy2016rethinking},
and $p\left(\y\right)$ is the marginal distribution: $p\left(\y\right)\approx\nicefrac{1}{\N}\sum_{n=1}^{\N}p\left(\y\gv\bx_{n}=G\left(\bz_{n}\right)\right)$. $H(\bx)$ represents entropy of variable $\bx$.

The Inception score shows a reasonable correlation with the quality and diversity of generated
images~\cite{salimans2016improved}. IS over real images can serve as the upper bound. Despite these appealing properties, IS has several limitations: 

\begin{enumerate}
\item First, similar to log-likelihood, it favors a “memory GAN” that stores all training samples, thus is unable to detect overfitting (\ie~can be fooled by generating centers of data modes~\cite{yang2017lr}). This is aggravated by the fact that it does not make use of a holdout validation set. 

\item Second, it fails to detect whether a model has been trapped into one bad mode (\ie~is agnostic to mode collapse). Zhou \etal~\cite{zhou2018activation}, however, shows results on the contrary.

\item Third, since IS uses Inception model that has been trained on ImageNet with many object classes, it may 
favor models that generate good objects rather realistic images.
%towards measuring \emph{``objectness''} rather than \emph{``realisticity''} the GAN is intended to strive towards.

\item Fourth, IS only considers $P_g$ and ignores $P_r$. Manipulations such as mixing in natural images from an entirely different distribution could deceive this score. As a result, it may favor models that simply learn sharp and diversified images, instead of $P_r$~\cite{huang2018an}\footnote{This also applies to the Mode Score.}.

\item Fifth, it is an asymmetric measure. 

\item Finally, it is affected by image resolution. See Fig.~\ref{fig:odena}.

\end{enumerate}

\begin{figure}[t]
\centering
\includegraphics[width=.9\linewidth]{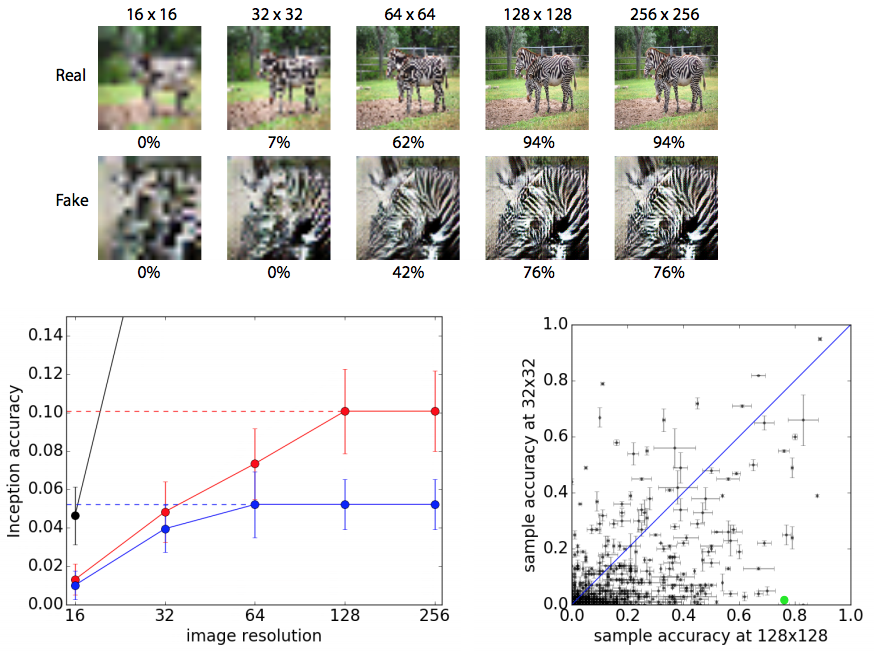}
\caption{Sensitivity of the inception score to image resolution. Top: Training data and synthesized images from the zebra class resized to a lower spatial resolution and subsequently artificially resized to the original resolution (128 × 128 for the red and black lines; $64 \times 64$ for the blue line). Bottom Left: IS score across varying spatial resolutions for training data and image samples from $64 \times 64$ and $128 \times 128$ models. Error bars show standard deviation across 10 subsets of images. Dashed lines highlight the accuracy at the output spatial resolution of the model. %The training data (clipped) achieves accuracies of 24\%, 54\%, 81\% and 81\% at resolutions of 32, 64, 128, and 256, respectively. 
Bottom Right: Comparison of accuracy scores at $128 \times 128$ and $32 \times 32$ spatial resolutions. Each point represents an ImageNet class. 84.4\% of the classes are below the diagonal. The green dot corresponds to the zebra class. Figure from~\cite{odena2016conditional}. %We also artificially resized 128 × 128 and 64 × 64 images to 256 × 256 as a sanity check to demonstrate that simply increasing the number of pixels will not increase discriminability.
} 
\label{fig:odena}
\end{figure}

% from https://arxiv.org/pdf/1703.02000.pdf 
Zhou \etal~\cite{zhou2018activation} provide an interesting analysis of the Inception score. They experimentally measured the two components of the IS score, entropy terms in Eq.~\ref{eq:1}, during training and showed that $H(y|\bx)$ behaves as expected (\ie~decreasing) while $H(y)$ does not. See Fig.~\ref{fig:Zhou} (top row). They found that 
CIFAR-10 data are not evenly distributed over the classes under the Inception model trained on ImageNet. See Fig.~\ref{fig:Zhou}(d). Using the Inception model trained over ImageNet or CIFAR-10, results in two different values for $H(y)$. Also, the value of $H(y|\bx)$ varies for each specific sample in the training data (\ie~some images are deemed less real than others). Further, a mode-collapsed generator usually gets a low Inception score (See Fig. 5 in~\cite{zhou2018activation}), which is a good sign. Theoretically, in an extreme case when all the generated samples are collapsed into a single point (thus $p(y) = p(y|\bx)$), then the minimal Inception score of 1.0 will be achieved. 
Despite this, it is believed that the Inception score can not reliably measure whether a model has collapsed. For example, a class-conditional model that simply memorizes one example per each ImageNet class, will achieve high IS values.
Please refer to~\cite{barratt2018note} for further analysis on the inception score.
%, which is the exp result of zero.

%xx et al., show that for an ImageNet Classifier, both the two indicators of Inception Score cannot work correctly.
%We experimentally find that, as in Figure 3b, the value of H(C¯G) is usually going down during
%the training process, however, which is expected to increase. 

%H(C(x)) for each specific sample in the training data, we find the value of H(C(x)) score is also
%variant, as illustrated in Figure 4b, which means, even in real data, it would still strongly prefer some
%samples than some others. 

\begin{figure}[t]
\centering
\includegraphics[width=\linewidth]{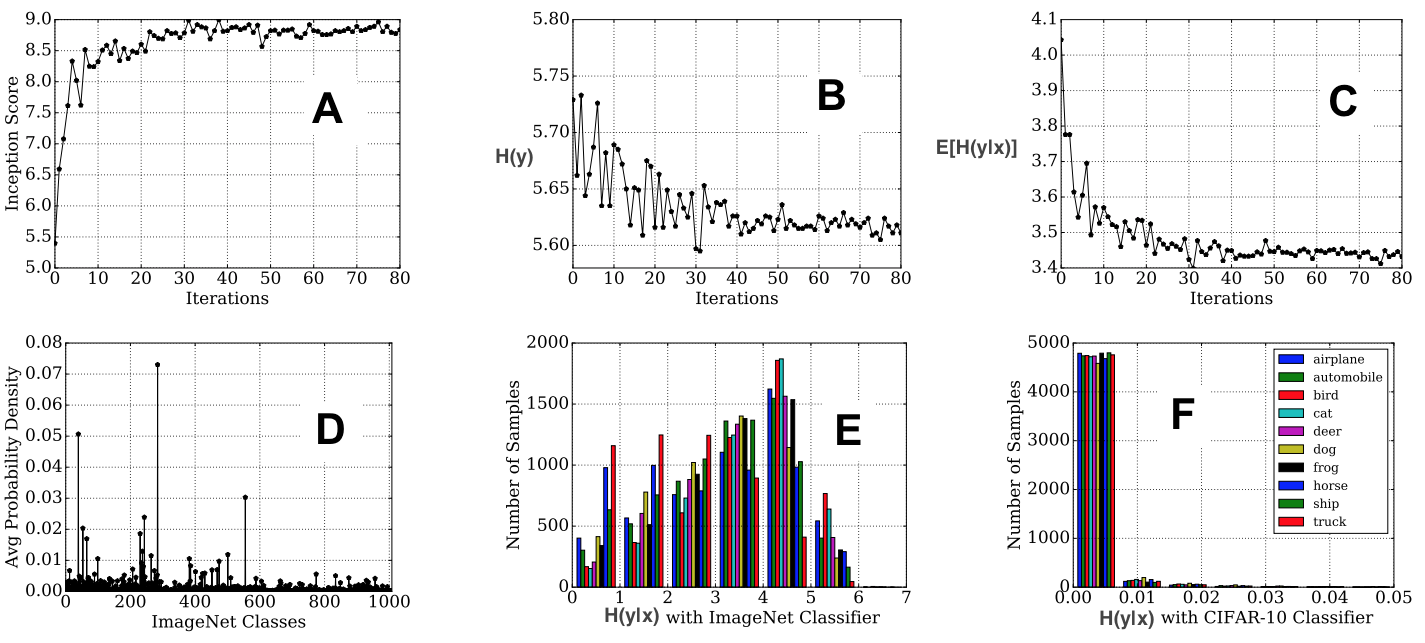}
\caption{Top: Training curves of Inception Score and its decomposed terms. A) IS during training, B) First term in rhs of Eq.~\ref{eq:1}, $H(y)$, goes down with training which is supposed to go up, C) The second term, $H(y|\bx)$ decreases in training, as expected. Bottom: Statistics of the CIFAR-10 training images. D) $p(y)$ over ImageNet classes, E) $H(y|\bx)$ distribution with ImageNet classifier of each class, and F) $H(y|\bx)$ distribution with CIFAR-10 classifier
of each class. Figure compiled from~\cite{zhou2018activation}.} 
\label{fig:Zhou}
\end{figure}

\item {\bf  Modified Inception Score (m-IS).}
%[[from DeLiGAN : Generative Adversarial Networks for Diverse and Limited Data
%Passing a generated image $x=G(z)$ through a trained classifier with an ``inception" architecture~\cite{szegedy2015going} results in a conditional label distribution $p(y|x)$. If $x$ is realistic enough, it should result in a ``peaky" label distribution i.e. $p(y|x)$ should have low entropy. We also want all categories to be covered uniformly among the generated samples, i.e. $ p(y) = \int_z p(y|x = G(z)) p_{z}(z)dz$ should have high entropy. These two requirements are unified into a single measure called ``inception-score" as  $e^{\mathbb{E}_x KL( p(y|x) || p(y) )}$ where $KL$ stands for KL-divergence and expectation $\mathbb{E}$ is taken over generated samples $x$. 
%\textbf{Our modification:} In its original formulation,
Inception score assigns a higher value to models with a low entropy class conditional distribution over all generated data $p(y|\bx)$. However, it is desirable to have diversity within samples in a particular category. To characterize this diversity, Gurumurthy \etal~\cite{gurumurthy2017deligan} suggested to use a cross-entropy style score $-p(y|\bx_{i})log(p(y|\bx_{j}))$ where $\bx_{j}$s are samples from the same class as $\bx_{i}$ based on the inception model's output. Incorporating this  term into the original inception-score results in:

\begin{equation}
\exp({\mathbb{E}_{\bx_{i}}[\mathbb{E}_{\bx_{j}}[(\mathbb {KL}(P(y|\bx_{i})||P(y|\bx_{j})) ]]}),
\end{equation}

which is calculated on a per-class basis and is then averaged over all classes. Essentially, m-IS can be viewed as a proxy for measuring both intra-class sample diversity as well as sample quality.

\item {\bf Mode Score.} %To address this problem, for labeled datasets, we further recruit the so-called 
Introduced in \cite{che2016mode}, this score addresses an important drawback of the Inception score which is ignoring the the prior distribution of the ground truth labels (\ie~disregarding the dataset):

\begin{equation}
\exp\left(\expect_{\bx}\left[\mathbb {KL}\left(p\left(y\gv\bx\right)\parallel{p}\left(y^{train}\right)\right)\right]-\mathbb {KL}\left(p\left(y\right)\parallel{p}\left(y^{train} \right)\right)\right),
\end{equation}

where ${p}\left(\y^{train}\right)$ is the empirical distribution of labels computed from training data. Mode score  adequately reflects the variety and visual quality of generated images~\cite{che2016mode}. It has been, however, proved that Inception and MODE scores are in fact equivalent. See~\cite{zhou2017inception} for the proof.

\item {\bf AM Score.} Zhou \etal~\cite{zhou2018activation} argue that the entropy term on $y$ in the Inception score is not suitable when the data is not evenly distributed over classes. To take $y^{train}$ into account, they proposed to replace $H(y)$ with the KL divergence between $y^{train}$ and $y$. The AM score is then defined as
%They proposed to leverage conditional and prior data distributions to address this. 
%[[We show that Inception score is actually equivalent to Mode score, both consisting of two entropy
%terms, which would be incompetent when the data is not evenly distributed over classes. We thus
%propose AM score as an alternative that leverages cross-entropy and takes the prior data distribution
%into account. 

\begin{equation} \label{eq:am-score}
\mathbb {KL}( p(y^\text{train})\parallel p(y))\!+\! \mE_{\bx} \big[H(y|\bx)\big].
\end{equation}

%They swap the order of $y^\text{train}$ and $p(y|\bx)$ in the two KL divergence terms in the inception score which leads to a more sensible metric (arguing that $y^\text{train}$ should be the reference term): 

%	\begin{align} \label{eq:am-score}
%	&\,\,\mE_{\bx} \big[\mathbb {KL} \big(p(y^\text{train}) \parallel p(y|\bx) \big) \big]\!-\!\mathbb {KL}( p(y^\text{train}) \parallel p(y)) \nonumber \\[2pt]
%	&= \mE_{\bx} \big[ H \big( y^\text{train}, y|\bx \big) \big]\!-\!H (y^\text{train} )\!-\!H ( y^\text{train}, y )\!+\!H ( y^\text{train} ) \nonumber \\[2pt]
%	&= \mE_{\bx} \big[ H\big( y^\text{train}, y|\bx \big)  \big] - H ( y^\text{train}, y ) % \triangleq \text{AM score}. 
%	\end{align}

The AM score consists of two terms. The first one is minimized when $y^{train}$ is close to $y$.  
%each sample is far away from the training data overall class distribution. 
The second term is minimized when the predicted class label for sample $\bx$ (\ie~$y|\bx$) has low entropy. Thus, the smaller the AM score, the better.

%the generated samples' average distribution is the same as training data. The overall class distribution indicated by the training data, i.e., $y^\text{train}$, has thus been taken into account, which is important when training data is not evenly distributed. 

It has been shown that the Inception score with $p(y|\bx)$ being the Inception model trained with ImageNet, correlates with human evaluation on CIFAR10. CIFAR10 data, however, is not evenly distributed over the ImageNet Inception model. The entropy term on average distribution of the Inception score may thus not work well (See Fig.~\ref{fig:Zhou}). With a pre-trained CIFAR10 classifier, the AM score can well capture the statistics of the average distribution. Thus, $p(y|\bx)$ should be a pre-trained classifier on a given dataset.

%Empirical results indicate that AM score outperforms Inception score.]]

%[[from https://arxiv.org/pdf/1703.02000.pdf
%6.4 AM SCORE WITH ACCORDINGLY PRETRAINED CLASSIFIER
%6.3 INCEPTION SCORE AS AN DIVERSITY MEASUREMENT
%
%a nice discussion there about both 
%
%ACTIVATION MAXIMIZATION GENERATIVE ADVERSARIAL NETS
%
%]]

%\textit{Pretrained Classifier:} 

%Note that the Inception score and the MODE score adopt an exponential transformation based on the above-calculated scores in Eq.~(\ref{eq:inception=mode}). With a pre-trained classifier on the given dataset, we will, however, show in the experiment that without the exponential transformation, AM score is informative enough. 

% From https://arxiv.org/pdf/1801.01401.pdf
\item {\bf Fr\'echet Inception Distance (FID).} 
Introduced by Heusel \etal~\cite{heusel2017gans}, FID embeds a set of generated samples into a feature space given by a specific layer of Inception Net (or any CNN). Viewing the embedding layer as a continuous multivariate Gaussian, the mean and covariance are estimated for
both the generated data and the real data. The Fr\'echet distance between these two Gaussians (\aka Wasserstein-2 distance) is then used to quantify the quality of generated samples, \ie~, 

%[[FID calculates the
%Wasserstein-2 distance between the generated images and the real images in the feature space of
%an Inception-v3 network. Lower FID values mean closer distances between synthetic and real data
%distributions. I]]

\begin{equation}
FID(r,g) = || \mu_r - \mu_g||^2_2 + Tr\left(\Sigma_r + \Sigma_g - 2(\Sigma_r\Sigma_g)^\frac{1}{2}\right),
\end{equation}

where $(\mu_r, \Sigma_r)$ and $(\mu_g, \Sigma_g)$ are the mean and covariance of the real data and model distributions, respectively. Lower FID means smaller distances between synthetic and real data
distributions. %sample embeddings from the real data and model distributions, respectively.  

%Note that under the Gaussian assumption on both Pr and Pg, the Fr\'echet distance is equivalent to the Wasserstein-2 distance.

% https://openreview.net/pdf?id=Sy1f0e-R-
FID performs well in terms of discriminability, robustness and computational efficiency. It
appears to be a good measure, even though it only takes into consideration the first two
order moments of the distributions. However, it assumes that features are of Gaussian distribution which is often not guaranteed. 
%The main advantage is the speed. 
%The calculating of FID distance is much faster than our evaluation methods.
%The disadvantage of FID is that it only considers difference in first two moments of the samples,
%which can be insufficient unless the feature maps are Gaussian distributed. On the other hand, the
%four metrics that we consider does not make any assumptions about the distribution of the samples.
It has been shown that FID is consistent with human judgments and is more robust to noise than IS~\cite{heusel2017gans} (\eg~negative correlation between the FID and visual quality of generated samples). Unlike IS however, it is able to detect intra-class mode dropping\footnote{On the contrary, Sajjadi et al.~\cite{sajjadi2018assessing} show that FID is sensitive to both the addition of spurious modes as well as to mode dropping.}, \ie~a model that generates only one image per class can score a high IS but will have a bad FID. Also, unlike IS, the FID worsens as various types of artifacts are added to images (See Fig.~\ref{fig:FID}). IS and AM scores measure the diversity and quality of generated samples, while FID measures the distance between the generated and real distributions. An empirical analysis of FID can be found in~\cite{lucic2017gans}. See also~\cite{liu2018improved} for a class-aware version of FID.

\begin{figure}[t]
\centering
\includegraphics[width=\linewidth]{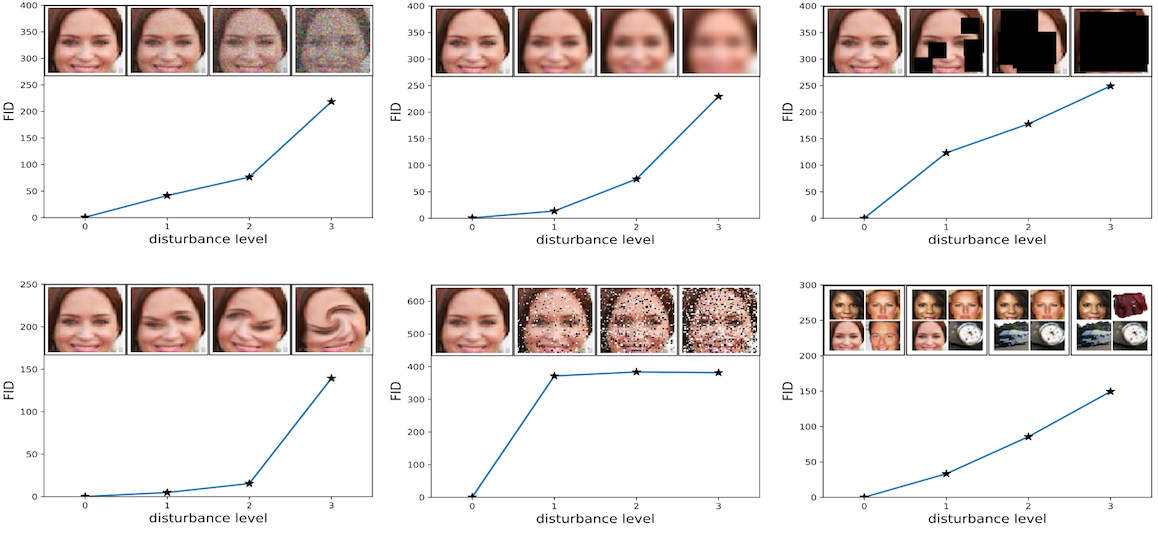}
\caption{FID measure is sensitive to image distortions. From upper left to lower right: Gaussian noise, Gaussian blur, implanted black rectangles, swirled images, salt and pepper noise, and CelebA dataset contaminated by ImageNet images. Figure from~\cite{heusel2017gans}.}
\label{fig:FID}
\end{figure}

%The disturbance level rises
%from zero and increases to the highest level. The FID captures the disturbance level very well by
%monotonically increasing. [Figure https://arxiv.org/pdf/1706.08500.pdf   Heusel et al.,~\cite{heusel2017gans}].
%[[Next we show that the FID is consistent with increasing disturbances and human judgment. Fig. 3
%evaluates the FID for Gaussian noise, Gaussian blur, implanted black rectangles, swirled images,
%salt and pepper noise, and CelebA dataset contaminated by the ImageNet images. The FID captures the
%disturbance level very well.  They also showed that where we show that FID is more consistent with the noise level than the Inception Score \cite{denton2015deep}]]} 
%\label{fig:denton}
%\end{figure}

% formula from https://openreview.net/pdf?id=Sy1f0e-R-
\item {\bf Maximum Mean Discrepancy (MMD).} 
This measure computes the dissimilarity between two probability distributions\footnote{Distinguishing two distributions by finite samples is known as \emph{Two-Sample Test} in statistics.} $P_r$ and $P_g$ using samples drawn independently from each~\cite{fortet1953convergence}. A lower MMD hence means that $P_g$ is closer to $P_r$. MMD can be regarded as two-sample testing since, as in classifier two samples test, it tests whether one model or another is closer to the true data distribution~\cite{muandet2017kernel,sutherland2016generative,bounliphone2015test}. Such hypothesis tests allow choosing one evaluation measure over another.

The kernel MMD~\cite{gretton2012kernel} measures (square) MMD between $P_r$ and $P_g$ for some fixed characteristic kernel function $k$ (\eg~Gaussian kernel $k(\bx,\bx')=\exp(\|\bx-\bx'\|^2)$) as follows\footnote{Please beware that here $\by$ represents the generated samples, and not the class labels.}: %Given two distributions $P_r$ and $P_g$, and a kernel $k$, the square of MMD distance is defined as
\begin{equation}
	%\vspace{-5pt}
	M_k(P_r, P_g) =  \mE_{\bx, \bx' \sim P_r}[ k(\bx,\bx') ] -2\mE_{\bx \sim P_r, \by \sim P_g}[k(\bx,\by)] + \mE_{\by,\by' \sim P_g}[ k(\by,\by') ].
\end{equation}
%theorem: characteristic kernel

%where k is a characteristic kernel (e.g., Gaussian kernel $k(x, x')=\exp(\|x-x'\|^2)$). 

In practice, finite samples from distributions are used to estimate MMD distance. Given $X = \{\bx_1, \cdots, \bx_n\} \sim P_r$ and $Y = \{\by_1, \cdots, \by_n\} \sim P_g$, one estimator of $M_k(P_r, P_g)$ is:
\begin{equation}
	\hat{M}_k(X, Y) = \frac{1}{{n \choose 2}}\sum_{i\neq i'} k(\bx_i, \bx_i') - \frac{2}{{n \choose 2}}\sum_{i\neq j} k(\bx_i,
	\by_j) + \frac{1}{{n \choose 2}}\sum_{j\neq j'} k(\by_j, \by_{j'}).
\end{equation}
Because of the sampling variance, $\hat{M}(X, Y)$ may not be zero even when $P_r=P_g$. Li \etal~\cite{li2017mmd} put forth a remedy to address this. Kernel MMD works surprisingly well when it operates in the feature space of a pre-trained CNN. It is able to distinguish generated images from real images, and both its sample complexity
and computational complexity are low~\cite{huang2018an}.

%Given two sets of samples from $P_r$ and $P_g$, the empirical MMD between the two distributions can be computed with finite sample approximation of the expectation. 
%The Parzen window estimate (Gretton et al., 2007) can be viewed as a specialization of Kernel MMD.
% https://openreview.net/pdf?id=Sy1f0e-R-

%Our goal is to measure the distance from P to Q using samples drawn independently from each distribution

% https://arxiv.org/pdf/1705.08584.pdf
Kernel MMD has also been used for training GANs. For example, the Generative Moment Matching Network (GMMN)~\cite{li2015generative,dziugaite2015training,li2017mmd} replaces the discriminator in GAN with a two-sample test based on kernel MMD. See also~\cite{binkowski2018demystifying} for more analyses on MMD and its use in GAN training.

\item {\bf The Wasserstein Critic.} 
%[[from Variational Approaches for Auto-Encoding Generative Adversarial Networks
%Independent Wasserstein critic: 
%Danihelka et al. [5] proposed training an independent Wasserstein GAN critic to distinguish between held out validation data and generated samples. We will train another critic to assign high values to validation samples and low values to generated samples. %Wasserstein GAN (WGAN) \citep{arjovsky2017wasserstein} uses a critic instead of a discriminator.
The Wasserstein critic~\cite{arjovsky2017wasserstein} provides an approximation of the Wasserstein distance between the real data distribution $P_r$ and the generator distribution $P_g$: 
\begin{align}
\label{eq:duality}
W(P_\mathit{r}, P_\mathit{g}) \propto \max_f \mathbb{E}_{\bx \sim P_\mathit{r}}\left[f(\bx)\right] - \mathbb{E}_{\bx \sim P_\mathit{g}}\left[f(\bx)\right],
\end{align}
where $f: \mathbb{R}^D \to \mathbb{R}$ is a Lipschitz continuous function. In practice, the critic $f$ is a neural network with clipped weights to have bounded derivatives. It is trained to produce high values at real samples and low values at generated samples (\ie~is an approximation):
%The difference of the expected critic values then approximates the Wasserstein distance.
%The approximation is scaled by the Lipschitz constant for the critic \citep{arjovsky2017wasserstein}.
%After training the critic for the current generator distribution,
%the generator can be trained to minimize the approximate Wasserstein distance.
%We will now look at the approximate Wasserstein distance between the validation data and the generator distribution. To approximate the Wasserstein distance, we will use the duality in Equation~\ref{eq:duality}. We will train another critic to assign high values to validation samples and low values to generated samples. 
%This \textit{independent} critic will be used only for evaluation. The generator will not see the gradients from the independent critic.
\begin{align}
\label{eq:indep_critic}
  \hat{W}(\bx_\mathit{test}, \bx_g) &= \frac{1}{N}\sum^{N}_{i=1} \hat{f}(\bx_\mathit{test}[i]) - \frac{1}{N}\sum^{N}_{i=1}, \hat{f}(\bx_g[i])
\end{align}
where $\bx_\mathit{test}$ is a batch of samples from a test set, $\bx_g$ is a batch of generated samples, and $\hat{f}$ is the independent critic. For discrete distributions with densities $P_r$ and $P_g$, the Wasserstein distance is often referred to as the Earth Mover's Distance (EMD) which intuitively is the minimum mass displacement to transform one distribution into the other. A variant of this score known as sliced Wasserstein distance (SWD) approximates the Wasserstein-1 distance between real and generated images, and is computed as the statistical similarity between local image patches extracted from Laplacian pyramid representations of these images~\cite{karras2017progressive}. We will discuss SWD in more detail later under scores that utilize low-level image statistics.

%]]

This measure addresses both overfitting and mode collapse. If the generator memorizes the training set, the critic trained on test data can distinguish between samples and data. If mode collapse occurs, the critic will have an easy job in distinguishing between data and samples. Further, it does not saturate when the two distributions do not overlap. The magnitude of the distance indicates how easy it is for the critic to distinguish between samples and data. 
%[[Using an Independent critic for evaluation has been proposed and used in practice before, see “Comparison of Maximum Likelihood and GAN-based training of Real NVPs”, Danihelka et all, as well as Variational Approaches for Auto-Encoding Generative Adversarial Networks, Rosca at all. 

The Wasserstein distance works well when the base distance is computed in a suitable feature space. A key limitation of this distance is its high sample and time complexity. These make Wasserstein distance less appealing as a practical evaluation measure, compared to other ones (See~\cite{arora2017gans}).

\item {\bf Birthday Paradox Test.} 
%from https://arxiv.org/pdf/1706.08224.pdf]]
This test approximates the support size\footnote{The support of a real-valued function $f$ is the subset of the domain containing those elements which are not mapped to zero.} of a discrete distribution.
%Suppose a distribution has support N. 
Arora and Zhang~\cite{arora2017gans} proposed to use the birthday paradox\footnote{The ``Birthday theorem'' states that with probability at least 50\%, a uniform sample (with replacement) of size $S$ from a set of  $N$ elements will have a duplicate given $S > \sqrt{N}$.} test to evaluate GANs as follows:

\begin{enumerate}
\item Pick a sample of size $S$ from the generated distribution
\item Use an automated measure of image similarity to flag the $k$ (\eg~$k=20$) most similar pairs in the sample
\item Visually inspect the flagged pairs and check for duplicates
\item Repeat.

\end{enumerate}

The suggested plan is to manually check for duplicates in a sample of size $S$. If a duplicate exists, then the estimated support size is $S^2$. It is not possible to find exact duplicates as the distribution of generated images is continuous. Instead, a distance measure can be used to find near-duplicates (\eg~using the $L_2$ norm). 
In practice, they first created a candidate pool of potential near-duplicates by choosing the 20 closest pairs according
to some heuristic measure, and then visually identified the near duplicated. Following this procedure and using Euclidean distance in pixel space, Arora and Zhang~\cite{arora2017gans} found that with probability $\geq 50\%$, a batch of about $400$ samples generated from the CelebA dataset~\cite{liu2015faceattributes} contains at least one pair of duplicates for both DCGAN and MIX+DCGAN (thus leading to support size of $400^2$). The birthday theorem assumes uniform sampling. Arora and Zhang~\cite{arora2017gans}, however, claim that the birthday paradox holds even if data are distributed in a highly nonuniform way. This test can be used to detect mode collapse in GANs.% ) ]]

%In the GAN setting, the distribution is continuous, not discrete. When support size is infinite then in a finite sample, we should not expect exact duplicate images where every pixel is identical. Thus a priori one imagines the birthday paradox test to completely not work. But surprisingly, it still works if we look for near-duplicates. 

%Given a finite sample, we select the 20 closest pairs according to some heuristic metric, thus obtaining a candidate pool of potential near-duplicates inspect. Then we visually identify if any of them would be considered duplicates byhumans. For example,  over the CelebA dataset, we find that with probability >= 50\%, a batch of about 400 samples contains at least one pair of duplicates for both DCGAN and MIX+DCGAN.

%Our test were done using two datasets, CelebA (faces) and CIFAR-10.

%https://openreview.net/forum?id=BJehNfW0-

%[[The proposed approach relies on so-called birthday paradox which allows to estimate the number of objects in the support by counting number of matching (or very similar) pairs in the generated sample
%
%This paper proposes a clever new test based on the birthday paradox for measuring diversity in generated samples. 

%The main idea used to estimate the support is the Birthday theorem that says that with probability at least 1/2, a uniform sample (with replacement) of size S from a set of  N elements will have a duplicate given $S > \sqrt{N}$. 

\item {\bf Classifier Two-sample Tests (C2ST).}
%from https://arxiv.org/pdf/1610.06545.pdf
%[[From Revisiting Classifier Two-Sample Tests for GAN Evaluation and Causal Discovery.
%David Lopez-Paz, Maxime Oquab.
The goal of two-sample tests is to assess whether two samples are drawn from the same distribution~\cite{lehmann2006testing}. In other words, 
%[The goal of two-sample tests is to 
decide whether two probability distributions, denoted by P and Q, are equal. % (Lehmann \& Romano, 2006).]
The generator is evaluated on a held out test set. This set is split into a test-train and
test-test subsets. The test-train set is used to train a fresh discriminator, which tries to distinguish generated images
from the real images. Afterwards, the final score is computed as
the performance of this new discriminator on the test-test set and the freshly generated images. More formally,  %Perhaps intriguingly, one relatively unexplored method to build two-sample tests is the use of binary classifiers. In particular, construct a dataset by pairing the $n$ examples in $S_P$ with a positive label, and by pairing the $m$ examples in $S_Q$ with a negative label. If the null hypothesis ``$P =Q$'' is true, then the classification accuracy of a binary classifier on a held-out subset of this dataset should remain near chance-level. 
%As we will show, such \emph{Classifier Two-Sample Tests} (C2ST) learn a suitable representation of the data on the fly, return test statistics in interpretable units, have a simple null distribution, and their predictive uncertainty allow to interpret where $P$ and $Q$ differ.
%The goal of this paper is to establish the properties, performance, and uses of C2ST.  First, we analyze their main theoretical properties.  Second, we compare their performance against a variety of state-of-the-art alternatives.  Third, we propose their use to evaluate the sample quality of generative models with intractable likelihoods, such as Generative Adversarial Networks (GANs).  Fourth, we showcase the novel application of GANs together with C2ST for causal discovery.
  %Without loss of generality, 
assume we have access to two samples $S_P = \{\bx_1, \cdots, \bx_n\} \sim P^n(X) $ and $S_Q = \{\by_1, \dots, \by_n\} \sim Q^n (Y)$ where $\bx_i, \by_i \in \mathcal{X}$, for  all $i = 1, \ldots, n$. % and $j = 1, \ldots, m$, and $m=n$. 
To test whether the null hypothesis $H_0 : P=Q$ is true, these five steps need to be completed:
  
  \begin{enumerate}
    
 \item Construct the following dataset
  \begin{equation*}
    \mathcal{D} = \{(\bx_i, 0)\}_{i=1}^n \cup \{(\by_i, 1)\}_{i=1}^n =: \{(\bz_i, l_i)\}_{i=1}^{2n}.
  \end{equation*}
  
  \item Randomly shuffle $\mathcal{D}$, and split it into two disjoint \emph{training} and \emph{testing}
  subsets $\mathcal{D}_\text{train}$ and $\mathcal{D}_\text{test}$, where 
  $\mathcal{D} = \mathcal{D}_\text{train} \cup \mathcal{D}_\text{test}$ and
  $n_\text{test} := |\mathcal{D}_\text{test}|$. 
  
  \item  Train a binary classifier $f
  : \mathcal{X} \to [0,1]$ on $\mathcal{D}_\text{train}$. In the following, assume
  that $f(\bz_i)$ is an estimate of the conditional probability distribution
  $p(l_i = 1 | \bz_i)$. 
  
  \item  Calculate the  classification accuracy on $\mathcal{D}_\text{test}$:
  \begin{equation}\label{eq:stat}
    \hat{t} = \frac{1}{n_\text{test}} \sum_{(\bz_i,l_i) \in \mathcal{D}_\text{test}}
    \mathbb{I}\left[ \mathbb{I}\left(f(\bz_i) > \frac{1}{2}\right) = l_i \right]
  \end{equation}
  as the \emph{C2ST statistic}, where $\mathbb{I}$ is the
  indicator function. The intuition here is that if $P=Q$, the test accuracy in
  Eq.~\ref{eq:stat} should remain near
  chance-level. In contrast, if binary classifier performs better than chance then it implies that $P \neq Q$. % and the binary classifier unveils  distributional differences between the two samples, the test classification  accuracy in Eq.~\ref{eq:stat} should be \emph{greater} than chance-level.
  
  \item  To accept or reject the null hypothesis, compute a
  $p$-value using the null distribution of the C2ST. %, as discussed next.

  \end{enumerate}

%C2ST has been used in this paper: GANs for Biological Image Synthesis
%~\cite{osokin2017gans}

% https://openreview.net/pdf?id=Sy1f0e-R-
%One advantage of the C2ST measure is that it is bounded in the interval [0.5, 1]. 
In principle, any binary classifier can be adopted for computing C2ST. Huang \etal~\cite{huang2018an} introduce a variation of this measure known as the \textit{1-Nearest Neighbor} classifier. The advantage of using 1-NN over other classifiers is that it requires no special training and little hyperparameter tuning. Given two sets of real $S_r$ and generated $S_g$ samples with the same size (\ie~$|S_r| = |S_g|$), one can compute
the leave-one-out (LOO) accuracy of a 1-NN classifier trained on $S_r$ and $S_g$ with positive labels
for $S_r$ and negative labels for $S_g$. The LOO accuracy can vary from 0\% to 100\%. 
If the GAN memorizes samples in $S_r$ and re-generate them exactly, \ie~$S_g = S_r$, then the accuracy would be 0\%. This is because every sample from $S_r$ would have its nearest neighbor from $S_g$ with zero distance (and vice versa). If it generates samples that are widely different than real images (and thus completely separable), then the performance would be 100\%. Notice that chance level here is 50\% which happens when a label is randomly assigned to an image. Lopez-Paz and Oquab~\cite{lopez2016revisiting} offer a revisit of classifier two-sample tests in~\cite{lopez2016revisiting}.

Classifier two-sample tests can be considered a different form of two-sample test to MMD. MMD has the advantage of a U-statistic estimator with Gaussian asymptotic distribution, while the classifier 2-sample test has a different form (cf. [75]).  MMD can be better when the U-statistic convergence outweighs the potentially more powerful classifier (\eg from a deep network), while a classifier based test could be better if the classifier is better than the choice of kernel. 

%[75] D. Lopez-Paz, M. Oquab, Revisiting classifier two-sample tests, arXiv
%preprint arXiv:1610.06545.

%The presentation should rather be organized around 2-sample testing, thus grouping points 8 and 11, but that kind of clustering of metrics has not been appropriately done.  The current presentation certainly does have a coverage of methods, but misses these important connections.
%

\item {\bf Classification Performance.} 
One common indirect technique for evaluating the quality of unsupervised representation learning algorithms is to apply them as feature extractors on labeled datasets and evaluate the performance of linear models fitted on top of the learned features. %However, this approach is rather indirect and relies heavily on the choice of classifier.  For example, in the work by Radford et al, they used nearest neighbor classifiers, which suffers from the problem that Euclidean distance is not a good dissimilarity measure for images. 
For example, to evaluate the quality of the representations learned by DCGANs, Radford \etal~\cite{radford2015unsupervised} trained their model on ImageNet dataset and then used the discriminator’s convolutional features from all layers to train a regularized linear L2-SVM to classify CIFAR-10 images. They achieved 82.8\% accuracy on par with or better than several baselines trained directly on CIFAR-10 data. 

%Also,  Guo Jun Qi used image classification as a surrogate to quantitatively evaluate the resultant LS-GAN model.

A similar strategy has also been followed in evaluating conditional GANs (\eg~the ones proposed for style transfer). For example, an off-the-shelf classifier is utilized by Zhang \etal~\cite{zhang2016colorful} to assess the realism of synthesized images. 
% from https://arxiv.org/pdf/1603.08511.pdf
%Semantic interpretability (VGG classification): Does our method
%produce realistic enough colorizations to be interpretable to an off-the-shelf object
They fed their fake colorized images to a VGG network that was trained on real color photos.
If the classifier performs well, this indicates that the colorizations are accurate enough
to be informative about object class. They call this ``semantic interpretability''. Similarly, Isola \etal~\cite{isola2017image} proposed the ``FCN score'' to measure the quality of the generated images conditioned on an input segmentation map. They fed the generated images to the fully-convolutional semantic segmentation network (FCN)~\cite{long2015fully} and then measured the error between the output segmentation map and the ground truth segmentation mask.

%from [21], and use it to evaluate the Cityscapes labels -> photo task.
%The FCN metric evaluates how interpretable the generated photos are according to an off-the-shelf semantic segmentation algorithm (the fully-convolutional network, FCN, from [31]). The FCN predicts a label map for a generated
%photo. This label map can then be compared against the input ground truth labels using standard semantic segmentation
%metrics described below. The intuition is that if we generate a photo from a label map of “car on road”, then we
%have succeeded if the FCN applied to the generated photo detects “car on road”.
%Semantic segmentation metrics

% The intuition is that if the generated images are realistic, classifiers trained on real images will be able
%to classify the synthesized image correctly as well.

%FCN score!!! (from Cycle GAN)~\cite{zhu2017unpaired}

%It has also been used here:  ~\cite{liang2017generative} [[In this work, we focus on a more challenging semantic
%manipulation task, which aims to modify the semantic meaning of an object
%while preserving its own characteristics (e.g. viewpoints and shapes), such as
%cow!sheep, motor! bicycle, cat!dog. To]]

%\item{\bf GAN Quality Index (GQI).}
%\item{GAN QUALITY INDEX (GQI) BY GAN-INDUCED CLASSIFIER}
Ye et al.~\cite{ye2018gan} proposed an objective measure known as the {\bf GAN Quality Index (GQI)} to evaluate GANs. First, a generator $G$ is trained on a labeled real dataset with $N$ classes. Next, a classifier $C_{real}$ is trained on the real dataset. The generated images are then fed to this classifier to obtain labels. A second classifier, called the GAN-induced classifier $C_{GAN}$, is trained on the generated data. Finally, the GQI is defined as the ratio of the accuracies of the two classifiers:
\begin{equation} 
GQI = \frac{ACC(C_{GAN})}{ACC(C_{real})} \times 100
\end{equation}
GQI is an integer in the range of 0 to 100. Higher GQI means that the GAN distribution better matches the real data distribution. 

{\bf Data Augmentation Utility:} Some works measure the utility of GANs for generating additional training samples. This can be interpreted as a measure of the diversity of the generated images. 
%influence of training the classifier with a combination of real images from the training set and a number of generated images. 
Similar to Ye et al.~\cite{ye2018gan}, Lesort et al.~\cite{lesort2018evaluation} proposed to use a mixture of real and generated data to train a classifier and then test it on a labeled test dataset. The result is then compared with the score of the same classifier trained on the real training data mixed with noise. Along the same line, recently, Shmelkov et al.~\cite{shmelkov2018good} proposed to compare class-conditional GANs with GAN-train and GAN-test scores using a neural net classifier. GAN-train is a network trained on GAN generated images and is evaluated on real-world images. GAN-test, on the other hand, is the accuracy of a network trained on real images and evaluated on the generated images. They
%[[ from how good is my GAN
%In addition to these two measures, we study the 
%utility of images generated by GANs for augmenting training data. This can be interpreted as a measure of the diversity of the generated images.
%This paper is similar to this one: x
%https://openreview.net/forum?id=HJ1HFlZAb
%Evaluation of generative networks through their data augmentation capacity 
%We analyze the utility of GANs for data augmentation, i.e., for generating additional training samples, with the best-performing GAN model (SNGAN) under two settings. 
%influence of training the classifier with a combination of real images from the training set and a number of generated images. 
analyzed the diversity of the generated images by evaluating GAN-train accuracy with varying amounts of generated data. The intuition is that a model with low diversity generates redundant samples, and thus increasing the quantity of data generated in this case does not result in better GAN-train accuracy. In contrast, generating more samples from a model with high diversity produces a better GAN-train score.

Above mentioned measures are indirect and rely heavily on the choice of the classifier. Nonetheless, they are useful for evaluating generative models based on the notion that a better generative model should result in better representations for surrogate tasks (\eg~supervised classification). This, however, does not necessary imply that generated images have high diversity.

\item{\bf Boundary Distortion.}
Santurkar et al.~\cite{santurkar2018classification} aimed to measure diversity of generated samples using classification methods. This phenomenon can be viewed as a form of covariate shift in GANs wherein the generator concentrates a large probability mass on a few modes of the true distribution. % (somewhat similar to mode collapse!!). 
It is illustrated using two toy examples in Fig.~\ref{fig:covariate}. The first example regards learning a unimodal spherical Gaussian distribution using a vanilla GAN~\cite{goodfellow2014generative}. As can be seen in Fig.~\ref{fig:covariate}.A, the spectrum (eigenvalues of the covariance matrix) of GAN data shows a decaying behavior (unlike true data). The second example considers binary classification using logistic regression where the true distribution for each class is a unimodal spherical Gaussian. The synthetic distribution for one of the classes undergoes boundary distortion which causes a skew between the classifiers trained on true and synthetic data (Fig.~\ref{fig:covariate}.B). Naturally, such errors would lead to poor generalization performance on true data as well. Taken together, these examples show that a) boundary distortion is a form of covariate shift that GANs can realistically introduce, and b) this form of diversity loss can be detected and quantified even using classification.

Specifically, Santurkar et al. proposed the following method to measure boundary distortion introduced by a GAN:
\begin{enumerate}
\item{Train two separate instances of the given unconditional GAN, one for each class in true dataset D (assume two classes).} 
\item{Generate a balanced dataset by drawing N/2 from each of these GANs.} 
\item{Train a binary classifier based on the labeled GAN dataset obtained in Step 2 above.} 
\item{Train an identical, in terms of architecture and hyperparameters, classifier on the true data D for comparison.}
\end{enumerate}

Afterwards, the performance of both classifiers is measured on a hold-out set of true data. Performance of the classifier trained on synthetic data on this set acts as a proxy measure for diversity loss through covariate shift.
Notice that this measure is akin to the classification performance discussed above.
%Only show their fig. 1

\begin{figure}[t]
\centering
\includegraphics[width=.85\linewidth]{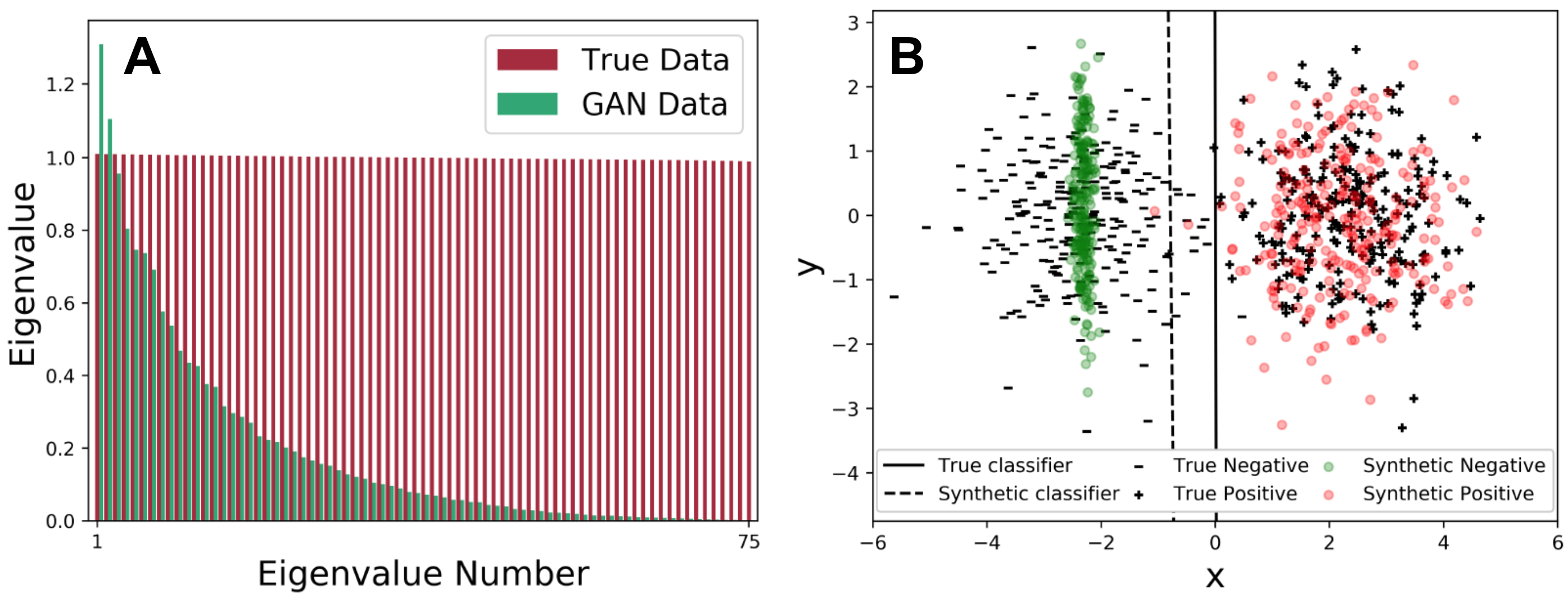}
\caption{A) Spectrum of the learned distribution of a vanilla GAN (a 200D latent space),
compared to that of the true distribution (a 75D spherical unimodal Gaussian). B) An example of 
covariate shift between synthetic and true distributions leading to a distortion in the learned decision boundary of a (linear) logistic regression classifier. Here the synthetic
distribution for one class suffers from boundary distortion. Figure compiled from~\cite{zhang2018decoupled}.} 
\label{fig:covariate}
\end{figure}

%from https://arxiv.org/pdf/1805.12462.pdf
\item {\bf Number of Statistically-Different Bins (NDB).}
To measure diversity of generates samples and mode collapse, Richardson and Weiss~\cite{Richardson2018} propose
%a simple method to evaluate generative models based on relative proportions of samples that fall into predetermined bins. Unlike previous automatic methods for evaluating models, our method does not rely on an additional neural network nor does it require approximating intractable computations. 
an evaluation method based on the following observation: Given two sets of samples from the same distribution, the number of samples that fall into a given bin should be the same up to sampling noise. More formally, let $I_B(\bx)$ be the indicator function for bin $B$. $I_B(\bx)=1$ if the sample $\bx$ falls into the bin $B$ and zero otherwise. Let $\{\bx^p_i\}$ be $N_p$ samples from distribution $p$ (\eg~training samples) and $\{\bx^q_j\}$ be $N_q$ samples from distribution $q$ (\eg~testing samples), then if $p=q$,
it is expected that $\frac{1}{N_p} \sum_i I_B(\bx^p_i) \approx \frac{1}{N_q} \sum_j I_B(\bx^q_j)$.
%The decision whether the number of samples in a given bin are \emph{statistically different} is a classic \emph{two-sample problem} for Bernoulli variables~\cite{daniel1995biostatistics}. 
The \textit{pooled sample proportion} $P$ (the proportion that falls into $B$ in the joined sets) and its standard error: $\mathit{SE} = \sqrt{P(1-P)[{1}/{N_p} + {1}/{N_q}]}$ are calculated. The test statistic is the $z$-score: $z = \frac{P_p-P_q}{SE}$, where $P_p$ and $P_q$ are the proportions from each sample that fall into bin $B$. If $z$ is smaller than a threshold (\ie~significance level) then the number is \emph{statistically different}. This test is performed on all bins and then the \emph{number of statistically-different bins} (NDB) is reported.

%There is still the question of which bin to use to compare the two distributions. 
To perform binning, one option is to use a uniform grid. The drawback here is that in high dimensions, a randomly chosen bin in a uniform grid is very likely to be empty. Richardson and Weiss proposed to use \emph{Voronoi cells} to guarantee that each bin will contain some samples. Fig.~\ref{fig:ndb} demonstrates this procedure using a toy example in $\mathbb{R}^2$. 
%A set of $N_p$ training samples from the reference distribution $p$ and a set of $N_q$ samples with distribution $q$, generated by the model we wish to evaluate. 
To define the Voronoi cells, $N_p$ training samples are clustered into K ($K \ll N_p,N_q$) clusters using 
K-means. Each training sample $\bx^p_i$ is assigned to one of the $K$ cells (bins). Each generated sample $\bx^q_j$ is then assigned to the nearest ($L_2$) of the $K$ centroids. %We perform the two-sample test on each cell separately and report the \emph{number of statistically-different bins} (NDB). 

Unlike IS and FID, NDB measure is applied directly on the image pixels rather than pre-learned deep representations.
This makes NDB domain agnostic and sensitive to different image artifacts (as opposed to using pre-trained deep models). One advantage of NDB over MS-SSIM and Birthday Paradox Test is that NDB offers a measure between the data and generated distributions and not just measuring the general diversity of the generated samples. One concern regarding NDB is that using $L_2$ distance in pixel space as a measure of similarity may not be meaningful. 

%  suffers from 
%A possible concern about using Voronoi cells as bins is that this essentially treats images as vectors in pixel spaces, where $L_2$ distance may not be meaningful. %[[This score can thus measure the diversity and overfitting. i.e., mode collapse]]

%Unlike previous automatic methods for evaluating models, our method does not rely on an additional neural network nor does it require approximating intractable computations.

\begin{figure}[t]
\centering
\includegraphics[width=\linewidth]{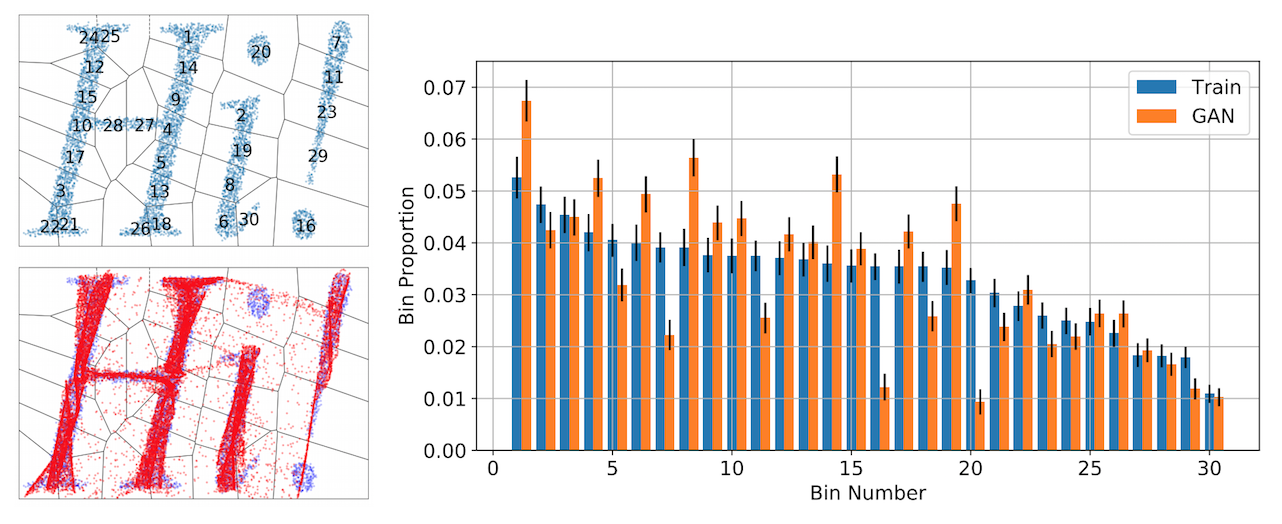}
\caption{Illustration of the NDB evaluation method on a toy example in $\mathbb{R}^2$. Top-left: The training data (blue) and binning result - Voronoi cells (numbered by bin size). Bottom-left: Samples (red) drawn from a GAN trained on the data. Right: Comparison of bin proportions between the training data and the GAN samples. Black lines = standard error ($\textit{SE}$) values. Figure from~\cite{Richardson2018}.}
\label{fig:ndb}
\end{figure}

\item {\bf Image Retrieval Performance.} %from Ensembles of Generative Adversarial Networks
Wang \etal~\cite{wang2016ensembles} proposed an image retrieval measure to evaluate GANs.  
The main idea is to investigate images in the dataset that are badly modeled by a network. 
%<We are especially interested if some images in the dataset are badly modeled by the network, which would lead to high nearest neighbor distances.> 
Images from a held-out test set as well as generated images are represented using a 
%Ns, with an image descriptor based on 
discriminatively trained CNN~\cite{lecun1998gradient}. The nearest neighbors of generated images in the test dataset are then retrieved. 
% , they look at <retrieve> their nearest neighbor in the generated image dataset.
To evaluate the quality of the retrieval results, they proposed two measures:

\begin{enumerate}
\item Measure 1: Consider $d_{i,j}^k$ to be the distance of the $j^{th}$ nearest image generated by method $k$ to test  image $i$, and ${\bf{d}}^k_{j}  = \left\{ {d_{1,j}^k, \cdots, d_{n,j}^k } \right\}$ the set of $j^{th}$-nearest distances to all $n$ test images ($j$ is often set to 1). The Wilcoxon signed-rank test is then used to test the hypothesis that the median of the difference between two nearest distance distributions by two generators is zero, in which case they are equally good (\ie~the median of the distribution ${\bf{d}}^k_{1}-{\bf{d}}^m_{1}$). If they are not equal, the test can be used to assess which method is statistically better. 
%This method is for example popular to compare illuminant estimation methods~\cite{hordley2006reevaluation}.

%Consider these distance distribution for two different generators.

\item Measure 2: Consider ${\bf{d}}^t_{j}$ to be the distribution of the $j^{th}$ nearest distance of the train images to the test dataset. Since train and test sets are drawn from the same dataset, the distribution ${\bf{d}}^t_{j}$ can be considered the optimal distribution that a generator could attain (assuming it generates an equal number of images present in the train set). To model the difference with this ideal distribution, the relative increase in mean nearest neighbor distance is  computed as:
\begin{equation}
\hat{d}_{j}^{k} = \frac{ \bar{d}_{j}^{k}- \bar{d}_{j}^{t}}{ \bar{d}_{j}^{t}}, \ \bar{d}_{j}^{k} = \frac{1}{N}\sum_{i = 1}^{N}d_{i,j}^{k}, \  \bar{d}_{j}^{t} = \frac{1}{N}\sum_{i = 1}^{N}d_{i,j}^{t},
\end{equation}
%\begin{equation}
%\bar{d}_{j}^{t} = \frac{1}{N}\sum_{i = 1}^{N}d_{i,j}^{t}
%\end{equation}
where $N$ is the size of the test dataset. As an example, $\hat{d}_{1}=0.1$ for a model means that the average distance to the nearest neighbor of a query image is 10\% higher than for data drawn from the real distribution.

\end{enumerate}

\item {\bf Generative Adversarial Metric (GAM).} Im \etal~\cite{im2016generating} proposed to compare two GANs by having them engaged in a battle against each other by swapping discriminators or generators across the two models (See Fig.~\ref{fig:battle_train}). GAM measures the relative performance of two
GANs by measuring the likelihood ratio of the two models. Consider two GANs with their respective
trained partners, $M_1 = (D_1, G_1)$ and $M_2 = (D_2, G_2)$, where $G_1$ and $G_2$ are the generators, and $D_1$ and $D_2$ are the discriminators. The hypothesis $\mathcal{H}1$ is that $M_1$ is better than $M_2$ if $G_1$ fools $D_2$ more than $G_2$ fools $D_1$, and vice versa for the hypothesis $\mathcal{H}0$. The likelihood-ratio is defined as:

\begin{equation}
\frac{p(\bx|y = 1; M'_1)}{p(\bx|y = 1; M'_2)}  =  \frac{p(y = 1|\bx; D_1)p(\bx; G_2)}{p(y = 1|\bx; D_2)p(\bx; G_1)},
\end{equation}

where $M'_1$ and $M'_2$ are the swapped pairs $(D_1, G_2)$ and $(D_2, G_1)$, $p(\bx|y = 1; M)$ is the likelihood
of $\bx$ generated from the data distribution $p(\bx)$ by model $M$, and $p(y = 1|\bx; D)$ indicates that discriminator $D$ thinks $\bx$ is a real sample.

%Hence, the procedure continues by simply swapping the partners of the two GANs, where the discriminator and generator from the first GAN is partnered with the discriminator and generator from the second GAN. 
Then, one can measure which generator fools the opponent’s discriminator more, $ \frac{D_1(\bx_2)}{D_2(\bx_1)}$
where $\bx_1 \sim G_1$ and $\bx_2 \sim G_2$. To do so, Im \etal proposed a sample ratio test to declare a winner or a tie. 
%% by defining the winning model

A variation of GAM known as generative multi-adversarial metric
(GMAM), that is amenable to training with multiple discriminators, has been proposed in~\cite{durugkar2016generative}.

GAM suffers from two main caveats:  
%First, it does not give an absolute value as it offers a relative comparison between two models. 
a) it has a constraint where the two discriminators must have an approximately similar performance on a calibration dataset, which can be difficult to satisfy in practice, and b) it is expensive to compute because it has to be computed for all pairs of models (\ie~pairwise comparisons between independently trained GAN).

%[[Rather, I would hope that a proper hypothesis test should be the goal of any system.  E.g. the meaning of a Wasserstein distance or MMD of 0.17 is not of itself really meaningful.  If we know the distribution of that metric's empirical expectation, we can then create proper hypothesis tests for model selection, just as we e.g. require that competing classifiers are evaluated using a Wilcoxon signed rank test. ]]

\begin{figure}[t]
\centering
\includegraphics[width=.7\linewidth]{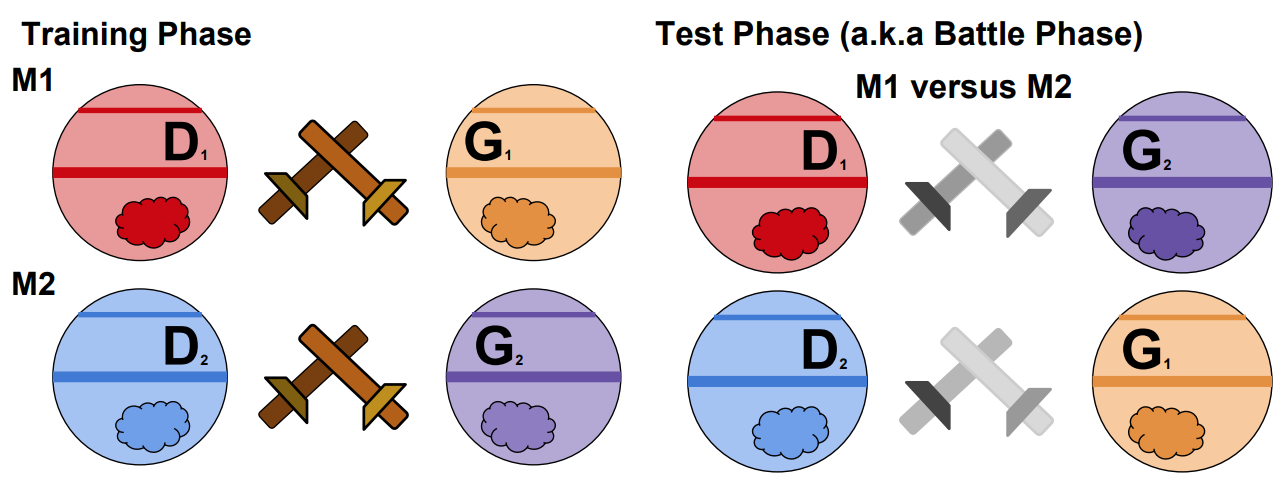}
\caption{Illustration of the Generative Adversarial Metric (GAM). 
During the training phase, $G_1$ and $G_2$ compete with $D_1$ and $D_2$, respectively. At test time, model M1 plays against M2 by having $G_1$ try to fool $D_2$, and vice-versa. M1 is better than M2 if $G_1$ fools $D_2$ more
than $G_2$ fools $D_1$ (and vice versa). Figure from~\cite{im2016generating}.} 
\label{fig:battle_train}
\end{figure}

\item{\bf Tournament Win Rate and Skill Rating.}
%The most closely-related metric to our tournament-based approach are the generative adversarial metric (GAM) [Im et al., 2016], and the generative multi-adversarial metric (GMAM) [Durugkar et al., 2017] which is an extension to generators whose training process involved multiple discriminator networks.
Inspired by GAM and GMAM scores (mentioned above) as well as skill rating systems in games such as chess or tennis, Olsson et al.~\cite{olsson2018skill} utilized tournaments between generators and discriminators for GAN evaluation. 	
They introduced two methods for summarizing tournament outcomes: tournament win rate and skill rating. Evaluations are useful in different contexts, including a) monitoring the progress of a single model as it learns during the training process, and b) comparing the capabilities of two different fully trained models. The former regards a single model playing against past and future versions of itself producing a useful measure of training progress (\aka within trajectory tournament). The latter regards multiple separate models (using different seeds, hyperparameters, and architectures) and provides a useful relative comparison between different trained GANs (\aka multiple trajectory tournament). Each player in a tournament is either a discriminator that attempts to distinguish between real and fake data or a generator that attempts to fool the discriminators into accepting fake data as real.

{\bf Tournament Win Rate:} To determine the outcome of a match between discriminator $D$ and generator $G$, the discriminator $D$ judges two batches: one batch of samples from generator $G$, and one batch of real data. Every
sample $\bx$ that is not judged correctly by the discriminator (\eg~$D(\bx) \geq 0.5$ for the generated data or
$D(\bx) \leq 0.5$ for the real data) counts as a win for the generator and is used to compute its win rate. 
A match win rate of $0.5$ for $G$ means that $D$'s performance against $G$ is no better than chance. The tournament win rate for generator $G$ is computed as its average win rate over all discriminators in $D$. Tournament
win rates are interpretable only within the context of the tournament they were produced from, and cannot be directly compared with those from other tournaments.

Olsson et al. ran a tournament between 20 saved checkpoints of discriminators and generators from the same
training run of a DCGAN trained on SVHN~\cite{netzer2011reading} using an evaluation batch size of 64. Fig.~\ref{fig:skill}.A shows the raw tournament outcomes from the within-trajectory tournament, alongside the same tournament outcomes summarized using tournament win rate and skill rating, as well as SVHN classifier score and SVHN Fréchet distance computed from 10,000 samples, for comparison\footnote{To compute these score, a pre-trained SVHN classifier is used rather than an ImageNet classifier.}. It shows that tournament win rate and skill rating both provide a comparable measure of training progress to SVHN classifier score.

{\bf Skill Rating:} Here the idea is to use a skill rating
system to summarize tournament outcomes in a way that takes into account the amount of new
information each match provides. Olsson et al. used the Glicko2 system~\cite{glickman1995comprehensive}. In a nutshell, a player's skill rating is represented as a Gaussian distribution, with a mean and standard deviation,
representing the current state of the evidence about their ``true'' skill rating. See \cite{olsson2018skill} for details of the algorithm.

Olsson et al. constructed a tournament from saved snapshots from six SVHN GANs that differ slightly from one
another, including different loss functions and architectures. They included 20 saved checkpoints of discriminators and generators from each GAN experiment, a single snapshot of 6-auto, and a generator player that produces batches of real data as a benchmark. Fig.~\ref{fig:skill}.B shows the results compared to Inception and Fr\'echet distances.

One advantage of these scores is that they are not limited to fixed feature sets and players can learn to attend to any features that are useful to win. Another advantage is that human judges are eligible to play as discriminators, and could participate to receive a skill rating. This allows a principled method to incorporate human perceptual judgments in model evaluation. %  to be incorporated into the evaluation of generative models in a more nuanced fashion, by taking into account the variation in judgment among human raters (See Section 2). 
The downside is providing relative rather than absolute score of a model's ability thus making reproducing results challenging and expensive.

%Finally, our method is more adaptable than moment matching approaches, because it does not require the experimenter to specify a fixed feature set; players in the tournament can learn to attend to any features that are useful to win.
% Some downsides to our approach include that it provides a relative rather than absolute score of a model’s ability, that tournaments among many types of models involve greater software complexity than other metrics, and that reproducing scores requires reproducing the population of models used in the tournament

%
%
%As a proof-of-concept for an
%unexplored domain in which a standard feature space is not available, we evaluate a GAN trained on
%70,000 hand-drawn images of apples from the QuickDraw [Ha and Eck, 2017] dataset. Although we
%represent the drawings as images rather than strokes, they are not “natural” images (i.e. photographs
%of the physical world). We compare within-trajectory skill rating to evaluation methods that use a
%natural image embedding space from an unrelated dataset (SVHN).
%Figure 2 shows that, subjectively, sample quality increases consistently with more iterations. SVHN
%Classifier score is a poor judge of quality for these samples. Fréchet distance is a better fit, but
%saturates at iteration 1300 whereas sample quality continues improving. Of these three methods,
%within-trajectory skill rating is the best fit, providing preliminary evidence that skill rating can succeed
%in unexplored domains.

\begin{figure}[htbp]
\centering
\includegraphics[width=1\linewidth]{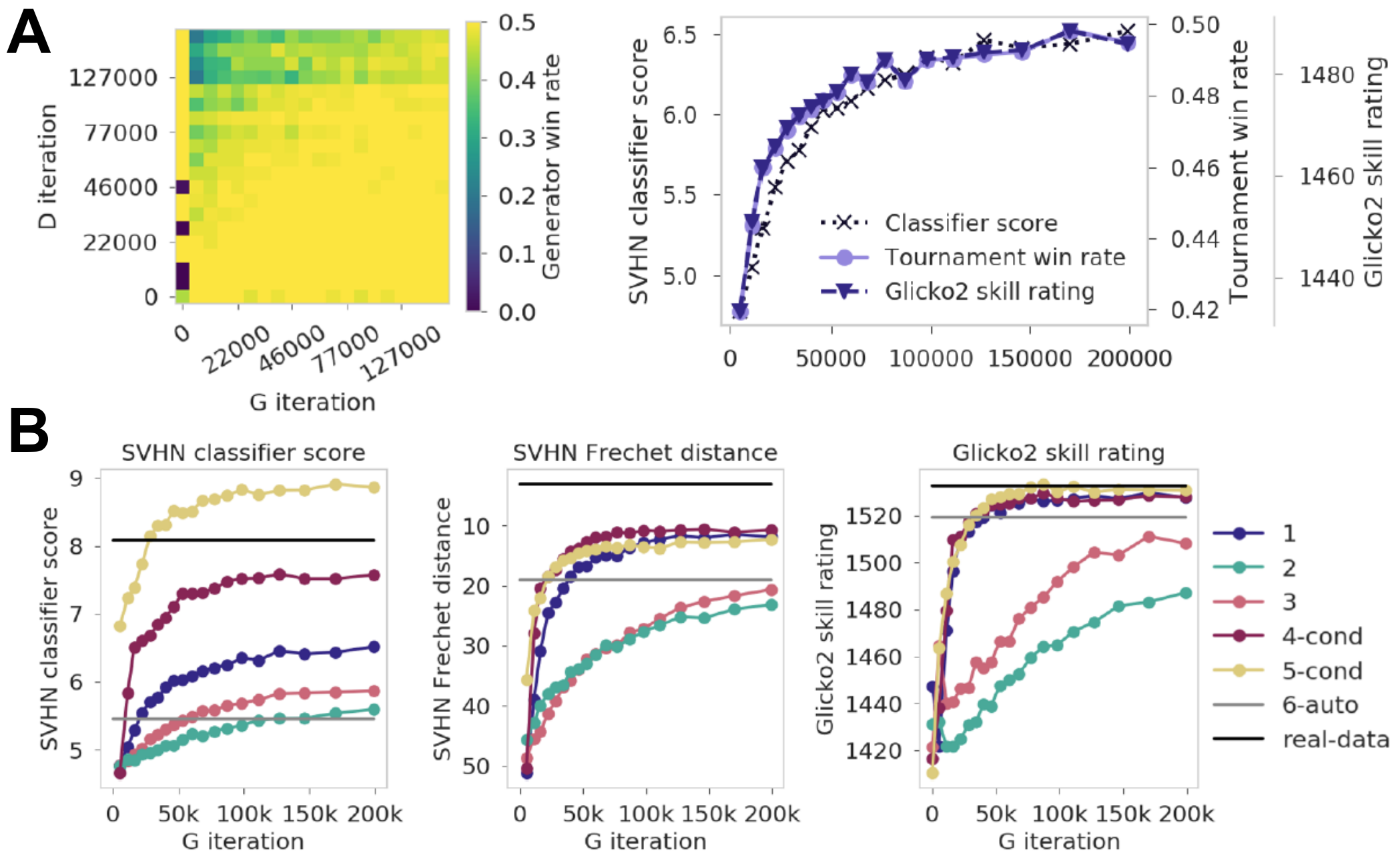}
\caption{A) A within-trajectory tournament. Left panel shows raw tournament outcomes. Each pixel represents the average win rate between one generator and one discriminator from different iterations. Brighter pixel
values represent stronger generator performance. Right panel compares tournament summary
measures to SVHN classifier score. Tournament win rate in this figure is the column-wise average of
the pixel values in the heatmap. 
B) Multiple-trajectory tournament outcomes among six models and real data. The tournament contains SVHN
generator and discriminator snapshots from models with different seeds, hyperparameters, and
architectures. Models are evaluated using SVHN classifier score (left), SVHN
Fr\'echet distance (center), and skill rating method (right). Each point represents
the score of one iteration of one model. The overall trajectories show the improvement of each
model with increasing training. Note the inverted y-axis on the Fr\'echet distance plot, such that lower
distance (better quality) is plotted higher on the plot. The learning curves produced by skill rating broadly agree with those produced by Fr\'echet distance, and disagree with classifier score only in the case of the conditional
models 4-cond and 5-con. Figure compiled from~\cite{olsson2018skill}. Please see text for more details on these experiments.} 
\label{fig:skill}
\end{figure}

% from https://arxiv.org/pdf/1801.06790.pdf
\item {\bf Normalized Relative Discriminative Score (NRDS).} 
%A new measure known as the Normalized Relative Discriminative Score
%has also been introduced recently by 
%More specifically, we
%train a single discriminator/classifier to separate real samples
%from generated samples, and those generated samples
%closer to real ones will be more difficult to be separated. For
%example, given two generative models G1 and G2, which
%define the distributions of generated samples pg1 and pg2,
%respectively. 
The main idea behind this measure proposed by Zhang \etal~\cite{zhang2018decoupled} is that more epochs would be needed to distinguish good generated samples from real samples (compared to separating poor ones from real samples). They used a binary classifier (discriminator) to separate the real samples from fake ones generated
by all the models in comparison. In each training epoch, the discriminator's
output for each sample is recorded. The average discriminator
output of real samples will increase with epoch
(approaching 1), while that of generated samples from each
model will decrease (approaching 0). However,
the decrement rate of each model varies based on how close
the generated samples are to the real ones. The samples
closer to real ones show slower decrement rate whereas poor samples will show a faster decrement rate. Therefore,
comparing the ``decrement rate'' of each model can be an indication of how well it performs relative to other models. 

%The decrement rate is proportional to the area under the curve of average discriminator output versus epoch. Larger area indicates slower decrement rate, implying that the generated samples are closer to real ones. 

There are three steps to compute the NRDS: 
\begin{enumerate}
\item Obtain the curve $\mathcal{C}_i$ ($i=1,2,\cdots,n$) of discriminator average output versus epoch (or mini-batch) for each model (assuming $n$ models in comparison) during training, 
\item Compute the area under each curve $A(\mathcal{C}_i)$ (as the decrement rate), and 
\item Compute NRDS of the $i$th model by 
\begin{equation}
	NRDS_i = \frac{A(\mathcal{C}_i)}{\sum_{j=1}^{n}A(\mathcal{C}_j)}.
	\label{eq:nrds}
\end{equation}

\end{enumerate}

The higher the NRDS, the better. Fig.~\ref{fig:NRDS} illustrates the computation of NRDS over a toy example.

\begin{figure}[t]
\centering
\includegraphics[width=1\linewidth]{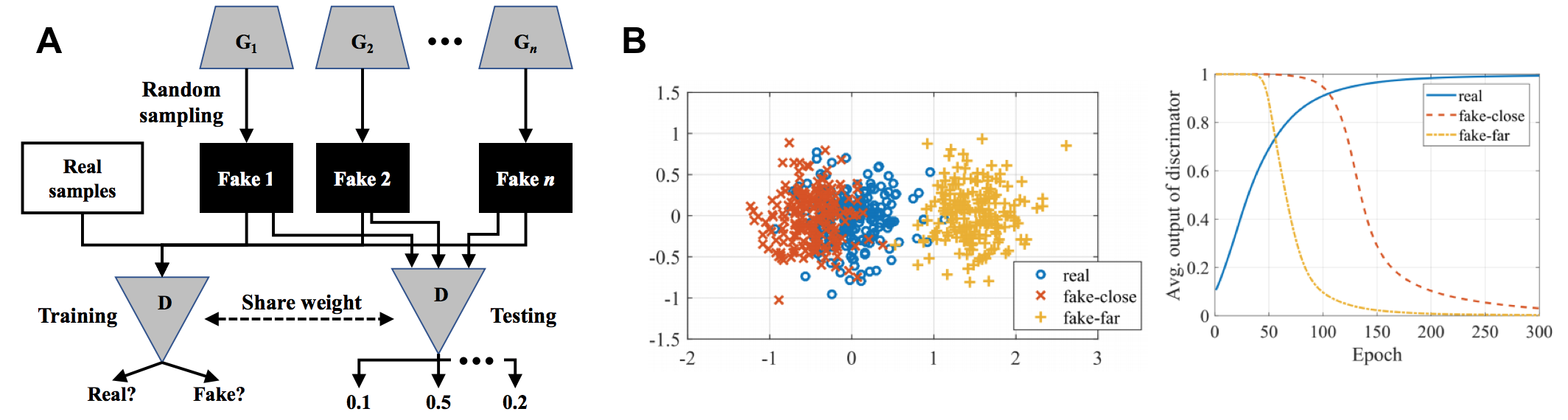}
\caption{A) Illustration of NRDS. $G_n$ indicates the $n$th generative
model. Its corresponding fake samples are Fake $n$, which are
sampled randomly. The fake samples from $n$ models, as well as
the real samples, are used to train the binary classifier $D$ during training (bottom
left). Testing only uses fake samples and performs alternatively
with the training process. The bottom right shows an example of
average output of $D$ for fake samples of each model. B) A toy example of computing NRDS. Left: the real
and fake samples randomly sampled from 2D normal distributions
with different means but with the same (identity) covariance.
The real samples (blue circles) have with zero mean. The red ``x''
and yellow ``+'' denote fake samples with the mean of [0.5, 0] and
[1.5, 0], respectively. The notation fake-close (fake-far) indicates that the
mean of correspondingly fake samples is close to (far from) that of the real samples. Right: the curves of epoch vs. averaged output of discriminator on corresponding sets (colors) of samples. 
In this example, the area under the curves of fake-close ($\mathcal{C}_1$) and fake-far ($\mathcal{C}_2$) are $A(\mathcal{C}_1)=145.4955$ and $A(\mathcal{C}_2)=71.1057$, respectively. From Eq.~\ref{eq:nrds}, $NRDS_1 =\frac{A(\mathcal{C}_1)}{\sum_{i=1}^{2}A(\mathcal{C}_i)}=0.6717$ and $NRDS_2 =\frac{A(\mathcal{C}_2)}{\sum_{i=1}^{2}A(\mathcal{C}_i)}=0.3283$. Therefore, the model generating fake-close is relatively better. Figure compiled from~\cite{zhang2018decoupled}.} 
\label{fig:NRDS}
\end{figure}

\item {\bf Adversarial Accuracy and Adversarial Divergence.}
%[[from LR-GAN: LAYERED RECURSIVE GENERATIVE ADVERSARIAL NETWORKS FOR IMAGE GENERATION
%Several metrics have been proposed to evaluate GANs, such as Gaussian parzen window citep{GAN}, Generative Adversarial Metric (GAM) citep{GRAN} and Inception Score citep{ImprovedGAN}. The common goal is to measure the similarity between the generated data distribution $P_g(\bm{x}) = G(\bm{z}; \theta_z)$ and the real data distribution $P(\bm{x})$. Most recently, Inception Score has been used in several works citep{ImprovedGAN, EBGAN}. However, it is an assymetric metric and could be easily fooled by generating centers of data modes. %In addition to these metrics, 
Yang \etal~\cite{yang2017lr} proposed two measures based on the intuition that a sufficient, but unnecessary, condition for closeness of generated data distribution $P_g(\bx)$ and the real data distribution $P_r(\bx)$ is closeness of $P_g(\bx|y)$ and $P_r(\bx|y)$, \ie~distributions of generated data and real data conditioned on all possible variables of interest $y$, \eg~category labels. One way to obtain the variable of interest $y$ is by asking human participants to annotate the images (sampled from $P_g(\bx)$ and $P_r(\bx)$). %, human  can be asked to label the category of the samples according to some rules. %Note that such human annotation is often easier than comparing samples from the two distributions (e.g., because there is no one to one correspondence between samples to conduct forced-choice tests). 

%After the annotations, we need to verify whether the two distributions are similar in each category. 

Since it is not feasible to directly compare $P_g(\bx|y)$ and $P_r(\bx|y)$, they proposed to compare $P_g(y|\bx)$ and $P_r(y|\bx)$ instead (following the Bayes rule) which is a much easier task. Two classifiers are then trained from human annotations to approximate $P_g(y|\bx)$ and $P_r(y|\bx)$ for different categories. These  classifiers are used to compute the following evaluation measures:

%In this case, we can simply train a discriminative model on the samples from $P_g(\bm{x})$ and $P_r(\bm{x})$ together with the human annotations about categories of these samples. With a slight abuse of notation, we use $P_g(y|\bm{x})$ and $P(y|\bm{x})$ to denote probability outputs from these two classifiers (trained on generated samples vs trained on real samples). We can then use these two classifiers to compute the following two evaluation metrics: 

\begin{enumerate}
\item \textit{Adversarial Accuracy:} Computes the classification accuracies achieved by the two classifiers on a validation set (\ie~another set of real images). If $P_g(\bx)$ is close to $P_r(\bx)$, then similar accuracies are expected.

\item \textit{Adversarial Divergence:} Computes the KL divergence between $P_g(y|\bx)$ and $P_r(y|\bx)$. The lower the adversarial divergence, the closer the two distributions. The lower bound for this measure is exactly zero, which means $P_g(y|\bx) = P_r(y|\bx)$ for all samples in the validation set. 

\end{enumerate}

One drawback of these measures is that a lot of human effort is needed to label the real and generated samples. To mitigate this, Yang \etal~\cite{yang2017lr} first trained one generator per category using a labeled training set and then 
generated samples from all categories. Notice that these measures overlap with classification performance discussed above.
  
%gene
%As discussed above, we need human efforts to label the real and generated samples. Fortunately, we can further simplify this. Based on the labels given on training data, we split the training data into categories, and train one generator for each category. With all these generators, we generate samples of all categories. %This strategy will be used in our experiments on the datasets with labels given.

%]]

% Krishna!
%To evaluate a conditional GAN with input $\bx_i$, and the corresponding output $\by_i$, first the KL divergence between $p(u|\bx)$ and $p(u|\by_i)$ is calculated and then is averaged over all indices $i$ (i.e., all pairs). 

% Analyzing the Training Processes of Deep Generative Models
% http://www.shixialiu.com/publications/dgmtracker/paper.pdf

\item{\bf Geometry Score.}
Khrulkov and Oseledets~\cite{khrulkov2018geometry} proposed to compare geometrical properties of the underlying data manifold between real and generated data. This score, however, involves a lot of technical details making it hard to understand and compute. Here, we provide an intuitive description. 

The core idea is to build a simplicial complex from data using proximity information (\eg~pairwise distances between samples). To investigate the structure of the manifold, a threshold $\epsilon$ is varied and generated simplices are added into the approximation. An example is shown in Fig.~\ref{fig:gm_score}.A. For each value of $\epsilon$, topological properties of the corresponding simplicial complex, namely homologies, are computed. A homology encodes the number of holes of various dimensions in a space. Eventually, a barcode (signature) is constructed reflecting how long generated holes (homologies) persist in simplicial complexes (Fig.~\ref{fig:gm_score}.B). In general, to find the rank of a $k$-homology (\ie~the number of $k$-dimensional holes) at some fixed value $\epsilon$, one has to count intersections of the vertical line $\epsilon =  0$ with the intervals at the desired block $H_k$.

Since computing the barcode using all data is intractable, in practice often subsets of data (\eg~by randomly selecting points) are used. For each subset, Relative Living Times (RLT) of each number of holes is computed which is defined as the ratio of the total time when this number was present and of the value $\epsilon_{max}$ when points connect into a single blob. The RLT over random subsets are then averaged to give the Mean Relative Living Times (MRLT). By construction, they add up to 1. To quantitatively evaluate the topological difference between two datasets, the $L_2$ distance between these distributions is computed. 

Fig.~\ref{fig:gm_score}.C shows an example over synthetic data. Intuitively, the value at location $i$ in the bar chart (on $x$ axis), indicates that for that amount of time, the 1D hole existed by varying the threshold. For example, in the left most histogram, nearly never none, 2 or 3 1D holes were observed and most of the time only one hole appeared. Similarly, for the 4th pattern from the left, most of the time one 1D hole is observed. Comparing the MRLTs of the patterns with the ground truth pattern (leftmost one) reveals that this one is indeed the closest to the ground truth. 

Fig.~\ref{fig:gm_score}.D shows comparison of two GANs, WGAN~\cite{arjovsky2017wasserstein} and WGAN-GP~\cite{gulrajani2017improved}, over the MNIST dataset using the method above over single digits and the entire dataset. It shows that both models produce distributions that are very close to the ground truth, but for almost all classes WGAN-GP shows better performance.

The geometry score does not use auxiliary networks and is not limited to visual data. However, since it only takes topological properties into account (which do not change if for example the entire dataset is shifted by 1) assessing the visual quality of samples may be difficult based only on this score. Due to this, authors propose to use this score in conjunction with other measures such as FID when dealing with natural images.
.

\begin{figure}[htbp]
\centering
\includegraphics[width=.85\linewidth]{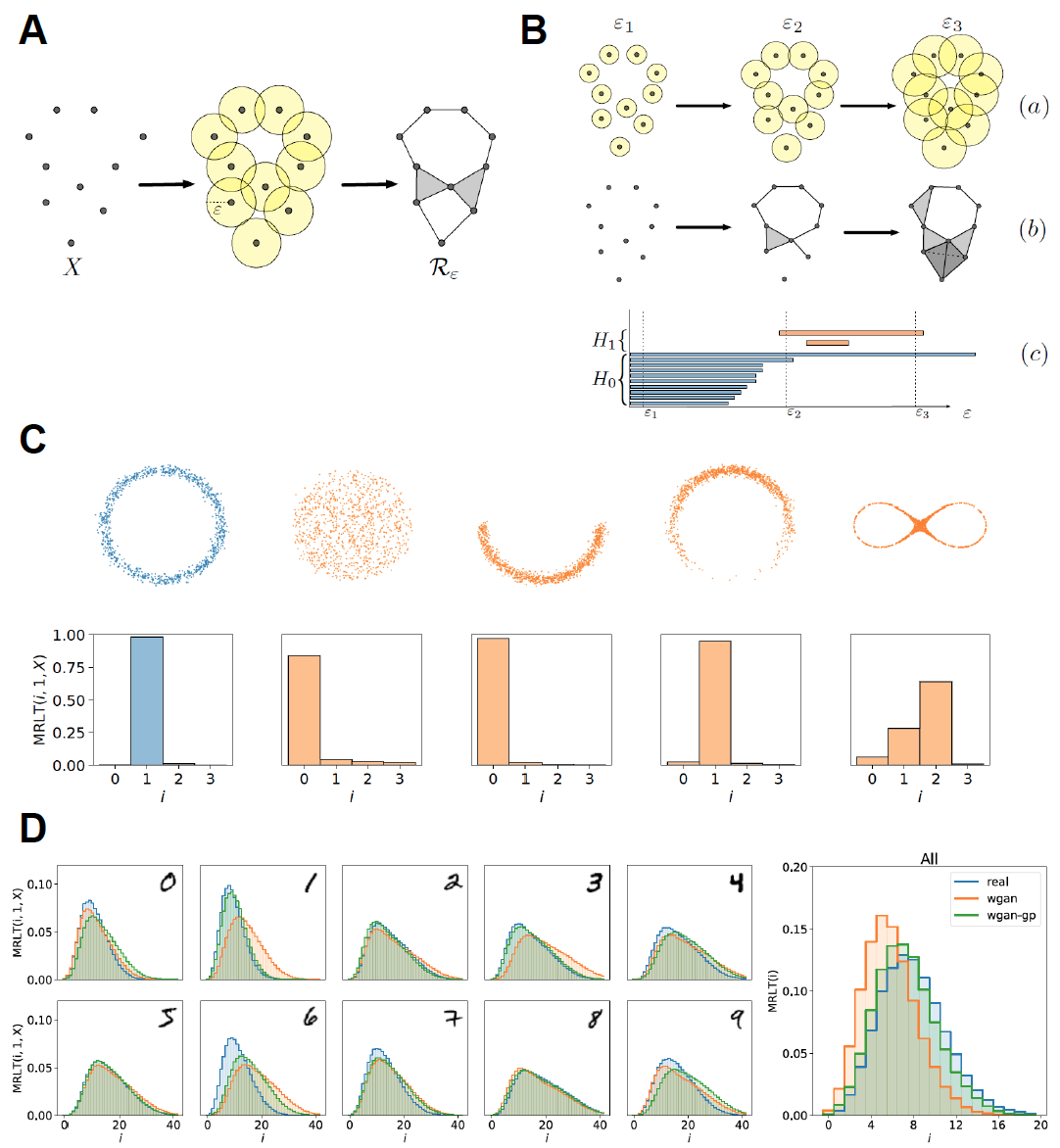}
\caption{A) A simplicial complex constructed on a sample $X$. First,
a fixed proximity parameter $\epsilon$ is chosen. Then, all balls of radius
$\epsilon$ centered at each point are considered, and if for some subset of $X$ of size $k + 1$
all the pairwise intersections of the corresponding balls are nonempty,
then the $k$-dimensional simplex spanning this subset is added to
the simplicial complex $R_x$. 
B) Using different values for $\epsilon$, different simplicial complexes are obtained (a). For $\epsilon=\epsilon_1$ the balls do not intersect and there are just 10 isolated components (b,
[left]). For $\epsilon=\epsilon_2$ several components are merged and one loop
is appeared (b, [middle]). The filled triangle corresponding to the
triple pairwise intersection is topologically trivial and does not
affect the topology (and similarly darker tetrahedron on the right).
For $\epsilon=\epsilon_3$ all the components are merged into one and the same hole
still exists (b, [right]). In the interval $[\epsilon_2, \epsilon_3]$ one smaller hole as
in A is appeared and is quickly disappeared. This information is summarized in the persistence barcode (c). The
number of connected components (holes) in the simplicial complex
for some value $\epsilon_0$ is given by the number of intervals in $H_0 (H_1)$
intersecting the vertical line $\epsilon = \epsilon_0$.
C) Mean Relative Living Times (MRLT) for various synthetic 2D datasets. The number of 1D holes is correctly identified in all the cases. Comparing MRLTs reveals that the second dataset from the left is closest to the `ground truth' (noisy circle on the left).
D) Comparison of MRLT of the MNIST dataset and of samples generated by WGAN and WGAN-GP trained on MNIST. MRLTs match almost perfectly, however, WGAN-GP shows slightly better performance on most of the classes. 
Figure compiled from~\cite{khrulkov2018geometry}.} 
\label{fig:gm_score}
\end{figure}

\item {\bf Reconstruction Error.} 
%[[from On the Effects of Batch and Weight Normalization in Generative Adversarial Networks]]
For many generative models, the reconstruction error on the training set is often explicitly optimized (\eg~Variational Autoencoders~\cite{ledig2016photo}). It is therefore natural to evaluate generative models
using a reconstruction error measure (\eg~$L_2$ norm) computed on a test set. In the case of
GANs, given a generator G and a set of test samples $X = \{ \bx^{(1)}, \bx^{(2)}, \cdots, \bx^{(m)}\}$ the reconstruction
error of G on $X$ is defined as:

\begin{equation}
\mathcal{L}_{rec}(G,X) = \frac{1}{m} \sum_{i=1}^{m} min_\bz || G(\bz) - \bx^{(i)} ||^2.
\end{equation}

Since it is not possible to directly infer the optimal $\bz$ from $\bx$, Xiang and Li~\cite{xiang2017effects} used the following alternative method. Starting from an all-zero vector, they performed gradient descent on the latent code to find the one that minimizes the $L_2$ norm between the sample generated from the code and the target one. Since the code is optimized instead of being computed from a feed-forward network, the evaluation process
is time-consuming. Thus, they avoided performing this evaluation at every training iteration when
monitoring the training process, and only used a reduced number of samples and gradient
descent steps. Only for the final trained model, they performed an extensive evaluation on a larger test
set, with a larger number of steps. 

%]]  <Ref> and results! Once we find the optimal z, we can then use the L2 norm. [[maybe L2 norm GAN paper]]
%
%This can be easily used for evaluating conditional GANs since the correponding groundtruth image exists! [was it used in Isola?! what did they use]

\item {\bf Image Quality Measures (SSIM, PSNR and Sharpness Difference).} Some researchers have proposed to use measures from the image quality assessment literature for training and evaluating GANs. They are explained next.

%Different generative works~\cite{journals/corr/MathieuCL15,DBLP:journals/corr/LedigTHCATTWS16,DBLP:conf/cvpr/ShiCHTABRW16,DBLP:journals/corr/ParkYYCB17} have used Structural-Similarity (SSIM), PSNR (Peak Signal-to-Noise Ratio) and Sharpness Difference to compare between the real data and the generated samples to quantify their work. 

%In this paper [[??]], we train neural nets with the structural-similarity metric (SSIM) \cite{wang2004ssim} and its multi-scale extension (MS-SSIM) \cite{Wang2003_multiscalessim}.
%We chose the SSIM family of metrics because it is well accepted and frequently utilized in the literature.
%Further, its pixel-wise gradient has a simple analytical form and is inexpensive to compute.
%In this work, we focus on the original grayscale SSIM and MS-SSIM, although there are interesting variations and improvements such as colorized SSIM \cite{kolaman2012quaternion,Hassan2012_colorssim}.

%Learning to Generate Images With Perceptual Similarity Metrics

\begin{enumerate}

\item 
The single-scale SSIM measure \cite{wang2004image} is a well-characterized perceptual similarity measure that aims to discount aspects of an image that are not important for human perception. It compares corresponding pixels and their neighborhoods in two images, denoted by $x$ and $y$, using three quantities---luminance ($I$), contrast ($C$), and 
structure ($S$):
\begin{equation*} \label{eq:ssim_components}
\begin{split}
I(x,y) \!=\! \frac{2\mu_x\mu_y + C_1}{\mu_x^2+\mu_y^2+C_1}  ~~~~
C(x,y) \!=\! \frac{2\sigma_x\sigma_y + C_2}{\sigma_x^2+\sigma_y^2+C_2}  ~~~~
S(x,y) \!=\! \frac{\sigma_{xy} + C_3}{\sigma_x\sigma_y+C_3}
\end{split}
\end{equation*}
The variables $\mu_x$, $\mu_y$, $\sigma_x$, and $\sigma_y$ denote
mean and standard deviations of pixel intensity in a local image patch centered at either $x$ or
$y$ (typically a square neighborhood of 5 pixels).  
The variable $\sigma_{xy}$ denotes the sample correlation
coefficient between corresponding pixels in the patches centered at $x$ and $y$. The constants $C_1$, $C_2$, and $C_3$ are small values added for numerical stability. The three quantities are combined to form the SSIM score:
%\vspace{-.035in} 
\begin{equation*}\label{eq:ssim_1}
\text{SSIM}(x,y) = I(x,y)^\alpha  C(x,y)^\beta  S(x,y)^\gamma
%\vspace{-.035in}
\end{equation*}

SSIM assumes a fixed image sampling density and viewing distance. %, and may only be appropriate for certain range of image scales. 
A variant of SSIM operates at multiple scales. The input images $x$ and $y$ are iteratively downsampled by a factor of 2 with a low-pass filter, with scale $j$ denoting the original images downsampled by a factor of $2^{j-1}$. The contrast $C(x,y)$ and
structure $S(x,y)$ components are applied to all scales. The luminance component is applied only to the
coarsest scale, denoted $M$. Further, contrast and structure components can be weighted at each scale. The final measure is:
%\vspace{-.035in}
\begin{equation*}\label{eq:mssim}
\text{MS-SSIM}(x,y) = I_M(x,y)^{\alpha_M}  \prod_{j=1}^{M}{ C_j(x,y)^{\beta_j}  S_j(X,y)^{\gamma_j}  }
%\vspace{-.035in}
\end{equation*}

%In our work, we weight each component and each scale equally ($\alpha = \beta_{1..M} = \gamma_{1..M} =1$), a common simplification of MS-SSIM. Following \cite{Wang2003_multiscalessim}, we use $M=5$ downsampling steps. 

MS-SSIM ranges between 0 (low similarity) and 1 (high similarity). Snell \etal~\cite{ridgeway2015learning} defined a loss function for training GANs which is the sum of structural-similarity scores over all image pixels,
%\vspace{-.05in}
\begin{equation*}\label{eq:objfn}
\mathcal{L}(X,Y) = -\sum_{i} \text{MS-SSIM}(X_i,Y_i) ,
%\vspace{-.05in}
\end{equation*}
where $X$ and $Y$ are the original and reconstructed images, and $i$ is an index over image pixels. This loss function
has a simple analytical derivative \cite{wang2008maximum} which allows performing gradient descent. See Fig.~\ref{fig:Snell} for more details.

\item PSNR measures the peak signal-to-noise ratio between two monochrome images $I$ and $K$ to assess the quality of a generated image compared to its corresponding real image (\eg~for evaluating conditional GANs~\cite{krishnaGAN}). The higher the PSNR (in db), the better quality of the generated image. It is computed as:
\begin{eqnarray}
PSNR (I, K)   &=& 10 \log_{10} \left( \dfrac{MAX^2_{I}}{MSE} \right)  \\
		   &=& 20 \log_{10} \left( MAX_{I}) -  20 \ log_{10} (MSE_{I,K} \right) 
\end{eqnarray}
where
\begin{equation}
MSE_{I, K}   =  \dfrac{1}{mn} \sum_{i = 0}^{m-1} \sum_{i = 0}^{n-1} (I(m,n) - K(m,n))^2  
\end{equation}
and, $MAX_{I}$ is the maximum possible pixel value of the image (\eg~255 for an 8 bit representation). This score can be used when a reference image is available for example in training conditional GANs using paired data (\eg~\cite{isola2017image,krishnaGAN}).

%$MAX_{K}$ = 255 (maximum pixel intensity value). 

%{\bf PSNR.}  P SNR = 20 log10(MAXI ) - 10 log10 MSE
%%from this: http://cs231n.stanford.edu/reports/2017/pdfs/17.pdf
%[[this paper explains them well]] with MAXI - a max level of intensity, usually 255. SSIM metric is calculated on various windows of an image. The measure between two windows x and y y of common size N x N is:
%[[looks like psnr is good for super resolution ]]

\item Sharpness Difference (SD) measures the loss of sharpness during image generation. It is compute as: % the difference of gradients between the images as 
\begin{equation}
SD(I, K)   = 10 \log_{10} \left( \dfrac{MAX^2_{I}}{GRADS_{I,K}} \right),
\end{equation}

where
\begin{equation}
GRADS_{I,K} = \dfrac{1}{N}\sum_{i} \sum_{j}  |(\triangledown_{i}I + \triangledown_{j}I) - (\triangledown_{i}K + \triangledown_{j}K)|,
\end{equation}

and 
\begin{equation}
\triangledown_{i}I = |I_{i,j} - I_{i-1,j}|, \ \triangledown_{j}I = |I_{i,j} - I_{i,j-1}|.
\end{equation}

%We encourage people to evaluate the evaluation measure like we did in xx, yy, and z for saliency modeling.
%Like everywhere else in CV there are lots of scores for different vision problems.!! and there is debate of course

\end{enumerate}

Odena et al.~\cite{odena2016conditional} used~\footnote{or `abused' since the original MS-SSIM measure is intended to measure similarity of an image with respect to a reference image.} MS-SSIM to evaluate the diversity of generated images. The intuition is that image pairs with higher MS-SSIM seem more similar than pairs with lower MS-SSIM. They measured the MS-SSIM scores of 100 randomly chosen pairs of images within a given class. The higher (lower) diversity within a class, the lower (the higher) mean MS-SSIM score (See Fig.~\ref{fig:MS-SSIM}.A). 
Training images from the ImageNet training data contain a variety of mean MS-SSIM scores across the
classes indicating the variability of image diversity in ImageNet classes. 
%Note that the highest mean MS-SSIM score (indicating the least variability) is 0.25 for the training data.
Fig.~\ref{fig:MS-SSIM}.B plots the mean MS-SSIM values for image samples versus training data for each class (after training was completed). It shows that 847 classes, out of 1000, have mean sample MS-SSIM scores below that of the maximum MS-SSIM for the training data. To identify whether the generator in AC-GAN~\cite{odena2016conditional} collapses during training, Odena et al. tracked the mean MS-SSIM score for all 1000 ImageNet classes (Fig.~\ref{fig:MS-SSIM}.C). Fig.~\ref{fig:MS-SSIM}.D shows the joint distribution of Inception accuracies versus MS-SSIM across all 1000 classes. It shows that Inception score and MS-SSIM are anti-correlated ($r^2$ = −0.16).

Juefei-Xu \etal~\cite{juefei2017gang} used the SSIM and PSNR measures to evaluate GANs in image completion tasks. The advantage here is that having 1-vs-1 comparison between the ground-truth and the completed image allows very straightforward visual examination of the GAN quality. It also allows head-to-head comparison between various GANs. In addition to the above mentioned image quality measures, some other measures such as Universal Quality Index (UQI)~\cite{WangBovik} and Visual Information Fidelity (VIF)~\cite{sheikh2006image} have also been adopted for assessing the quality of synthesized images.
%1706.09138.pdf%]]
%from http://research.nvidia.com/sites/default/files/publications/karras2017gan-paper.pdf
It has been reported that MS-SSIM finds large-scale mode collapses reliably but fails to diagnose smaller effects such as loss of variation in colors or textures. Its drawback is that it does not directly assess image quality in terms of similarity to the training set~\cite{odena2016conditional}.

%They also employed this metric to compare the diversity of the training images to
%the samples from the GAN model after training was completed.

%In other words, 84.7\% of classes have sample variability that exceeds that of the least variable class from the ImageNet training data.

\begin{figure}[htbp]
\centering
\includegraphics[width=1\linewidth]{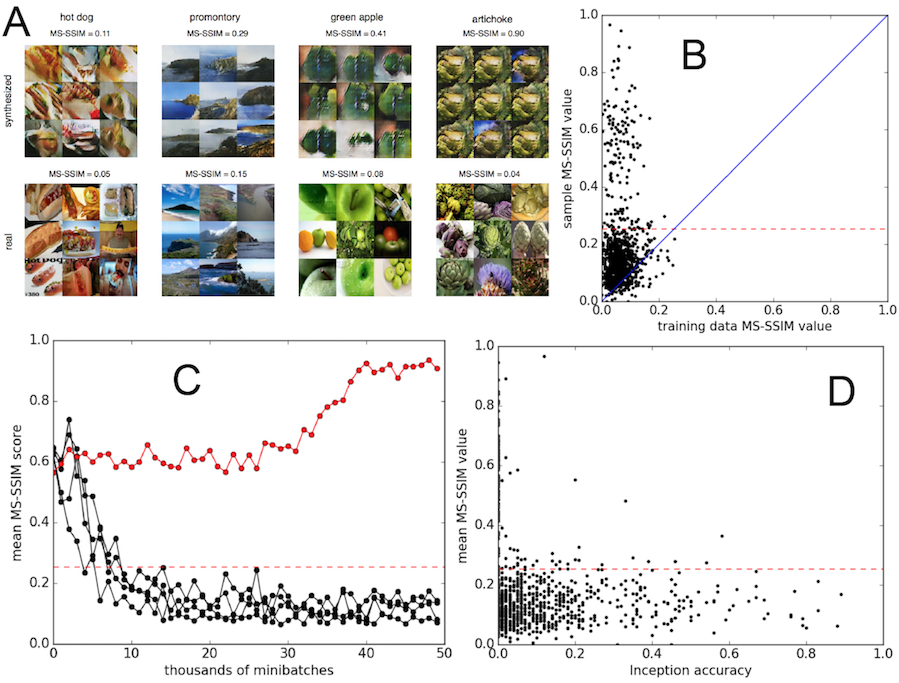}
\caption{MS-SSIM score used for measuring image diversity. A) MS-SSIM scores for samples generated by the AC-GAN~\cite{odena2016conditional} model (top row) and training samples (bottom row). B) The mean MS-SSIM scores between pairs of images within a given class of the ImageNet dataset versus AC-GAN samples. The horizontal red line marks the maximum MS-SSIM across all ImageNet classes (over training data). Each data point represents the mean MS-SSIM value for samples from one class. C) Intra-class MS-SSIM for five ImageNet classes throughout a training run. Here, decreasing trend means successful training (black lines) whereas increasing trend indicates that the generator `collapses' (red line). D) Comparison of Inception score vs. MS-SSIM for all 1000 ImageNet
classes ($r^2$ = −0.16). AC-GAN samples do not achieve variability at the expense of discriminability. Figure compiled from~\cite{odena2016conditional}.} 
\label{fig:MS-SSIM}
\end{figure}

\item {\bf Low-level Image Statistics.}  %Natural scene statistics Borji. Add formula
Natural scenes make only a tiny fraction of the space of all possible images and have certain characteristics (\eg~\cite{geisler2008visual,simoncelli2001natural,torralba2003statistics,ruderman1994statistics}). It has been shown that statistics of natural images remain the same when the images are scaled (\ie~scale invariance)~\cite{srivastava2003advances, zhu2003statistical}. The average power spectrum magnitude $A$ over natural images has the form $A(f) = 1/f^{-\alpha},~\alpha \approx 2$~\cite{deriugin1956power, cohen1975image, burton1987color, field1987relations}. Another important property of natural image statistics is the non-Gaussianity~\cite{srivastava2003advances, zhu2003statistical, wainwright1999scale}. This means that marginal distribution of almost any zero mean linear filter response on virtually any dataset of images is sharply peaked at zero, with heavy tails and high kurtosis (greater than 3)~\cite{lee2001occlusion}. Recent studies have shown that the contrast statistics of the majority of natural images follows a Weibull distribution~\cite{ghebreab2009biologically}.

Zeng \etal~\cite{zeng2017statistics} proposed to evaluate generative models in terms of low-level statistics of their generated images with respect to natural scenes. They considered 
%measured the low-level statistics of images generated by state-of-the-art deep generative models compared to natural images in terms of 
four statistics including 1) the mean power spectrum, 2) the number of connected components in a given image area, 3) the distribution of random filter responses, and 4) the contrast distribution.
%The images generated by GANs as well as natural scenes and cartoons, statistics including 
Their results show that although generated images by DCGAN~\cite{radford2015unsupervised}, WGAN~\cite{arjovsky2017wasserstein}, and VAE~\cite{kingma2013auto} resemble natural scenes in terms of low level statistics, there are still significant differences. For example, generated images do not have scale invariant mean power spectrum magnitude, which indicates existence of extra structures in these images caused by deconvolution operations.

%Our analyses on training images support current findings on scale invariance, non-Gaussianity, and Weibull contrast distribution of natural scenes. We find that although similar results hold over cartoon
%images, there is still a significant difference between statistics of natural scenes and images generated by both DCGAN and WGAN models. One big difference is that generated images do not have scale invariant mean power spectrum magnitude, which indicates existence of extra structures in these images caused by deconvolution operations.
%
%
%Another paper was basically our own TIP submission which was comparing the statistics of images
%A model is better which can generate more natural stats [we can also train models over aerial images and sketches and compare stats on them!]

%Building on the intuition that a successful generator should produce samples whose local image structure
%is similar to the training set over all scales, 

Low-level image statistics can be used for regularizing GANs to optimize the discriminator to inspect whether the generator's output matches expected statistics of the real samples (\aka feature matching~\cite{salimans2016improved}) using the loss function: $||\mathbb{E}_{\bx \sim P_\mathit{r}} f(\bx) - \mathbb{E}_{\bz \sim P_\mathit{\bz}(\bz)} f(G(\bz))||_2^2 $, where $f(.)$ represents the statistics of features. Karras \etal~\cite{karras2017progressive} investigated the multi scale statistical similarities between distributions of local image patches drawn from the Laplacian pyramid~\cite{burt1987laplacian} representations of generated and real images. %starting at a low-pass resolution of $16 \times 16$ pixels. As per standard practice, the pyramid progressively doubles until the full resolution is reached, each successive level encoding the difference to an up-sampled version of
%the previous. Then, 
They used the Wasserstein distance to compare the distributions of patches\footnote{This measure is known as the sliced Wasserstein distance (SWD)}. The multi-scale pyramid allows a detailed comparison of statistics. The distance between the patch sets extracted from the lowest resolution
%$16 \times 16$ images 
indicates similarity in large-scale image structures, while the finest-level
patches encode information about pixel-level attributes such as sharpness of edges and noise.

\item {\bf Precision, Recall and $F_1$ Score.} 
%Precision, recall and F1 score are proven and widely
%adopted techniques for quantitatively evaluating the quality
%of discriminative models. Precision measures the fraction of
%relevant retrieved instances among the retrieved instances,
%while recall measures the fraction of the retrieved instances
%among relevant instances. F1 score is the harmonic average
%of precision and recall
Lucic \etal~\cite{lucic2017gans} proposed to compute precision, recall and $F_1$ score to 
quantify the degree of overfitting in GANs. Intuitively precision measures the quality of the generated samples, whereas recall measures the proportion of the  reference distribution covered by the learned distribution.
They argue that IS only captures precision as it does not penalize
a model for not producing all modes of the data distribution. Rather, it only penalizes the model for not producing
all classes. FID score, on the other hand, captures both precision and recall. %Indeed, a model which fails to recover different modes of the data distribution will suffer in terms of FID.

To approximate these scores for a model, Lucic \etal proposed to use toy datasets for which the data manifold is known and distances of generated samples to the manifold can be computed. An example of such dataset is the manifold of convex shapes (See Fig.~\ref{fig:recall}). To compute these scores, first the latent representation $\bz$ of each test sample $\bx$ is estimated, through gradient descent, by inverting the generator $G$. Precision is defined as the fraction of the generated samples whose distance to the manifold is below a certain threshold. Recall, on the other hand, is given by the fraction of test samples whose $L_2$ distance to $G(\bz)$ is below the threshold.  If the samples from the model distribution $P_g$ are (on average) close to the manifold (see~\cite{lucic2017gans} for details), its precision is high. Simlarly, high recall implies that the generator can recover (\ie~generate something close to) any sample from the manifold, thus capturing most of the manifold.

The major drawback of these scores is that they are impractical for real images where the data manifold is unknown, and their use is limited to evaluations on synthetic data. In a recent effort, Sajjadi et al.~\cite{sajjadi2018assessing} 
introduced a novel definition of precision and recall to address this limitation. % and is theoretically sound and has desirable properties

%Recall is computed by considering a set of test samples.
% rephrase this

%evaluate GANs over a  a simple triangle dataset. 

%They proposed an approximation to precision and recall for GANs and how that it can be used to 

\begin{figure}[t]
\centering
\includegraphics[width=1\linewidth]{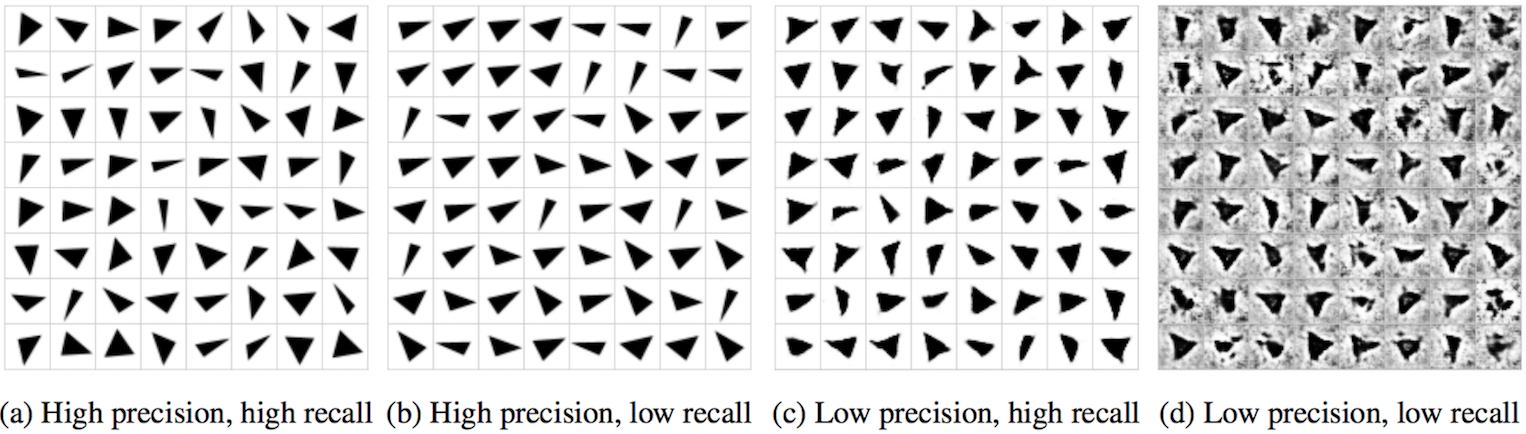}
\caption{Samples from a model trained on gray-scale triangles. These triangles belong to a low-dimensional
manifold embedded in $\mathbb{R}^{d \times d}$. A good generative model should be able to capture
the factors of variation in this manifold (\eg~rotation, translation, minimum angle size). a) high recall and precision, b) high precision, but low recall (lacking in diversity), c) low
precision, but high recall (can decently reproduce triangles, but
fails to capture convexity), and d) low precision and low recall. Figure from~\cite{lucic2017gans}.} 
\label{fig:recall}
\end{figure}

% r
%
%Intuitively,
%the coordinate system of this manifold represents the axes
%of variation (e.g. rotation, translation, minimum angle size,
%etc.). A good generative model should be able to capture
%these factors of variation and recover the training samples.
%Furthermore, it should recover any sample from this manifold
%from which we can efficiently sample which is illustrated
%in Figure 3.
%
%We perform a search over the wide range of hyperparameters
%and compute precision and recall by considering
%n = 1024 samples. In particular, we compute the precision
%of the model by computing the fraction of generated samples
%with distance below a threshold δ = 0.75. We then
%consider n samples from the test set and invert each sample
%
%squared Euclidean distance
%below δ

%Notice that these measures consider efficiency of evaluation measures and not models. Nevertheless, we men

\end{enumerate} % end of quantitative measures

% ------------------------------------------------------------------------------------------------------------------------------------------------

\subsection{Qualitative Measures}

Visual examination of samples by human raters%Human judgment of sample quality 
is one of the common and most intuitive ways to evaluate GANs (\eg~\cite{denton2015deep,salimans2016improved,millerhuman}). %Onsight can be gained by visually examining samples. 
While it greatly helps inspect and tune models, it suffers from the following drawbacks.
First, evaluating the quality of generated images with human vision is expensive and cumbersome, 
biased (\eg~depends on the structure and pay of the task, community reputation of the experimenter, etc in crowd sourcing setups~\cite{silberman2015stop}) difficult to reproduce, and does not fully reflect the capacity of models. Second, human inspectors may have high variance which makes it necessary to average over a large number of subjects. Third, an evaluation based on samples could be biased towards models that overfit and therefore a poor indicator of a good density model in a log-likelihood sense~\cite{theis2015note} For instance, it fails to tell whether a model drops modes.
%A subjective evaluation based on visual fidelity of samples is still clearly appropriate.
%Behavioral. User studies. User judgments. human perception!! although some quantitative measures such as SSIM also evaluates image quality, here we are interested in explicit human judgments.
%In practice this metric does not take into account mode dropping if the number of modes is greater than the number of samples one is visualizing. 
In fact, mode dropping generally helps visual sample quality as the model can choose to focus on only few common modes that correspond to typical samples. 

In what follows, I discuss the ways that have been followed in the literature to qualitatively inspect the quality of generated images by a model and explore its learned latent space.

%cite this one
%\cite{millerhuman}

\begin{enumerate}

\item {\bf Nearest Neighbors.} 
%Was common; maybe put a figure of this from somewhere xxx
To detect overfitting, traditionally some samples are shown next to their nearest neighbors in the training set (\eg~Fig.~\ref{fig:nn}). There are, however, two concerns regarding this manner of evaluation:

%[The most widely used method for evaluating GANs is to visually inspect the quality of generated samples. In particular, overfitting is detected by sampling from the GAN and then finding the closest training samples to see if they are sufficiently different, and by interpolating latent codes.

%<One test checks the similarity of each generated image to the nearest images in the training set. >

%There are several concerns regarding this. Theis et al.,~\cite{} argue that measuring overfitting of a model by taking samples from the model and making sure their training set nearest neighbors are different is ineffective, since it is actually trivial to generate samples that are each visually almost identical to a training example, but that yet each have large euclidean distance with their corresponding (visually similar) training example.  [[It is well-established that pixel representations of images do not induce meaningful Euclidean distances (Forsyth \& Ponce, 2011). Small translations, rotations, or changes in illumination can increase distances dramatically with little effect on the image content. To quantity the similarity between distributions of images, it is therefore desirable to use distances invariant to such transformations. ]]
% https://openreview.net/pdf?id=Sy1f0e-R-

\begin{enumerate}
%\item A qualitative assessment based on samples can be biased towards models which overfit (Breuleux et al., 2009), thus violating the whole purpose of 

\item Nearest neighbors are typically determined based on the Euclidean distance which is very sensitive to minor 
perceptual perturbations. This is a well known phenomenon in the psychophysics literature (See Wang and Bovik~\cite{wang2009mean}). It is trivial to generate samples that are visually almost identical to a training image, but have large Euclidean distances with it~\cite{theis2015note}. See Fig.~\ref{fig:theis}.A for some examples.

% <Even shifting pixels around can generate huge differences!> This is why L1 norm is not good enough for training generative models and why having a discriminator that uses high level features does better!!! <<Put the eucldean sensitivity fig here.>>

\item A model that stores (transformed) training images (\ie~memory GAN) can trivially pass the nearest-neighbor overfitting test~\cite{theis2015note}. This problem can be alleviated by choosing nearest neighbors based on perceptual measures, and by showing more than one nearest neighbor.

% from Reducing Mode Collapse in GANs using Implicit Variational Learning
\begin{figure}[t]
\centering
%\begin{subfigure}
  \includegraphics[width=0.474\linewidth]{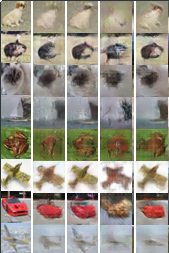}
 \includegraphics[width=0.482\linewidth]{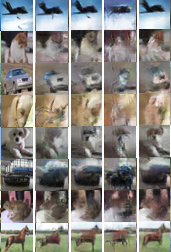}
\caption{Generated samples nearest to real images from CIFAR-10. In each of the two panels, the first column shows real images, followed by the nearest image generated by DCGAN~\cite{radford2015unsupervised}, ALI~\cite{dumoulin2016adversarially}, Unrolled GAN~\cite{metz2016unrolled}, and VEEGAN~\cite{srivastava2017veegan}, respectively. Figure compiled from~\cite{srivastava2017veegan}.}
\label{fig:nn}
%\end{subfigure}
\end{figure}

\end{enumerate}

\begin{figure}[htbp]
\includegraphics[width=1\linewidth]{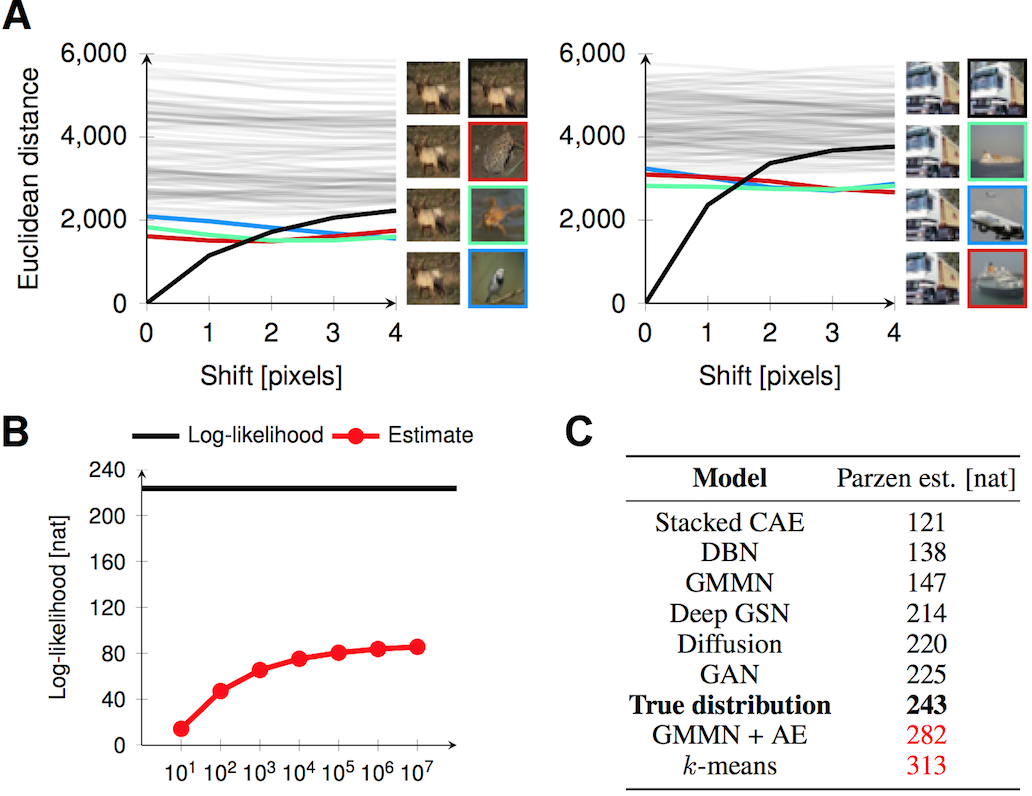}
\caption{A) Small changes to an image can lead to large changes
in Euclidean distance affecting the choice of the nearest neighbor. The left column shows a 
query image shifted 1 to 4 pixels (top to bottom). The right column shows the corresponding
nearest neighbor from the training set. The gray lines indicate Euclidean distance of
the query image to 100 randomly picked images from the training set. B) Parzen window estimates for a Gaussian
evaluated on 6 by 6 pixel image patches from the CIFAR-10 dataset. Even for small patches and a very large number of samples, the Parzen window estimate is far from the true log-likelihood. 
C) Using Parzen window estimates to evaluate various models trained on MNIST,
samples from the true distribution perform worse than samples from a simple model trained
with k-means. Figure compiled from~\cite{theis2015note}.} 
\label{fig:theis}
\end{figure}

\item {\bf Rapid Scene Categorization.} These measures are inspired by prior studies who have shown that humans are
capable of reporting certain characteristics of scenes in a short glance (\eg~scene category, visual layout ~\cite{oliva2005gist,serre2007feedforward}). To obtain a quantitative measure of quality of samples, Denton \etal~\cite{denton2015deep} asked volunteers to distinguish their generated samples from real images. The subjects were presented with the user interface shown in Fig.~\ref{fig:denton}(right) and were asked to click the appropriate button to indicate if they believed the image was real or generated. They varied the viewing time from  50ms to 2000ms (11 durations). Fig.~\ref{fig:denton}(left) shows the results over samples generated by three GAN models. They concluded that their model was better than the original GAN~\cite{goodfellow2014generative} since it did better in fooling the subjects (lower bound here is 0\% and upper bound is 100\%). See also Fig.~\ref{fig:salimas} for another example of fake vs. real experiment but without time constraints (conducted by Salimans \etal~\cite{salimans2016improved}).

%shown at random four different types of image: samples drawn from three different GAN models trained on CIFAR10. 

%After being presented with an image, the subject clicked the appropriate button to indicate if they believed the image was real or generated. Since accuracy is a function of viewing time, we also randomly pick the presentation time from one of 11 durations ranging from 50ms to 2000ms, after which a gray mask image is displayed. Before the experiment commenced, they were shown examples of real images from CIFAR10. After collecting 10k samples from the volunteers, we plot in Fig. 6 the fraction of images believed to be real for the four different data sources, as a function of presentation time. 
%
%<<Good news: One intuitive metric of performance can be obtained by having human annotators judge the visual quality of samples [2]. an be obtained by having human annotators judge the visual quality of samples [2]. We automate this process using Amazon Mechanical Turk (MTurk), using the web interface in figure Fig. 2  which we use to ask annotators to distinguish between generated data and real data.

This ``Turing-like'' test is very intuitive and seems inevitable to ultimately answer the question of whether generative models are as good as the nature in generating images. However, there are several concerns in conducting such a test in practice (especially when dealing with models that are far from perfect; See Fig.~\ref{fig:denton}(left)). Aside from experimental conditions which are hard to control in crowd-sourced platforms (\eg~presentation time, screen size, subject's distance to the screen, subjects' motivations, age, mood, feedback, etc) and high cost, these tests fall short in evaluating models in terms of diversity of generated samples and may be biased towards models that overfit to training data. % or drop the modes.  <This can be done in further detail categories similar to % GIST>

%Well, rapid scene categorization might not be the best since finding artifacts may take more viewing time and detailed inspection. A downside of using human annotators is that the metric varies depending on the setup of the task and the motivation of the annotators. XX et al., find that results change drastically when we give annotators feedback about their mistakes: By learning from such feedback, annotators are better able to point out the flaws in generated images, giving a more pessimistic quality assessment. The left column of Fig. 2 presents a screen from the annotation process, while the right column shows how we inform annotators about their mistakes. My guess: The problem with visual fidelity is that it does not test the variety in the sampled distribution. For example, if a models does mode collapse and then all the samples are sampled from that one mode, while the quality is good and model can fool a person more than another model!!>> 

%The curves show our models produce samples that are more realistic than those from standard GAN [11]. 

%I think the behavioral task by Emily Denton has also some problems
%Question: why humans are not so good in telling whether natural scenes are fake or perfect!! 

\begin{figure}[t]
\includegraphics[width=\linewidth]{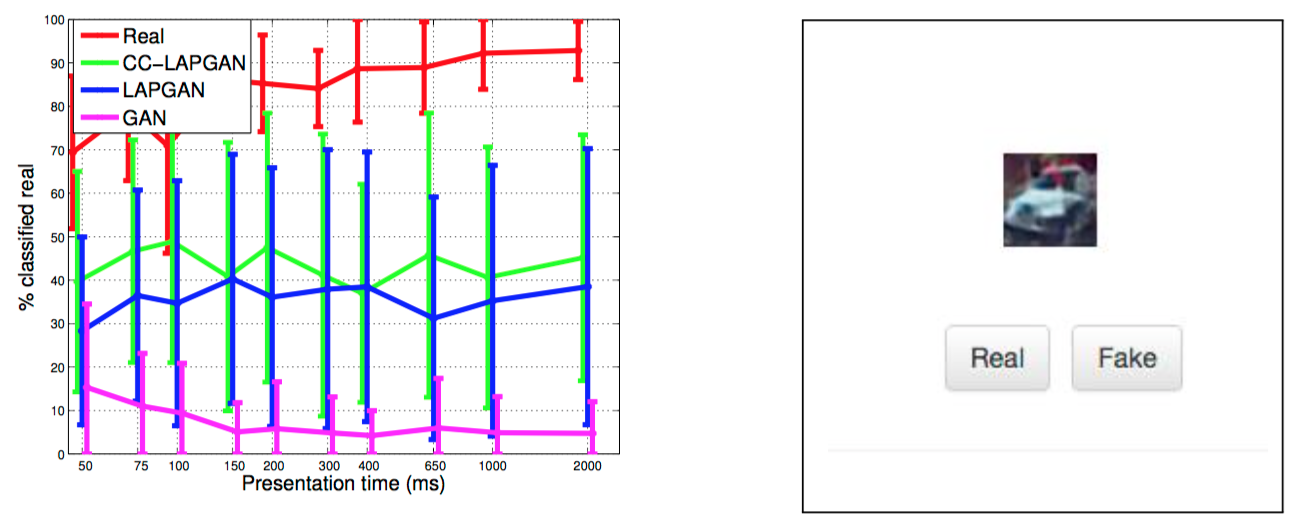}
\caption{Left: Human evaluation of real CIFAR10 images (red) and samples from Goodfellow
\etal~\cite{goodfellow2014generative} (magenta), and LAPGAN~\cite{denton2015deep} and a class conditional LAPGAN (green). Around 40\% of the samples generated by the class conditional LAPGAN model are realistic enough to fool a human into thinking they are real images. This compares with $\leq$ 10\% of images from the standard GAN model, but is still a lot lower than the > 90\% rate for real images. Right: The user-interface presented to the subjects. Figure from~\cite{denton2015deep}.} 
\label{fig:denton}
\end{figure}

\begin{figure}[t]
\includegraphics[width=\linewidth]{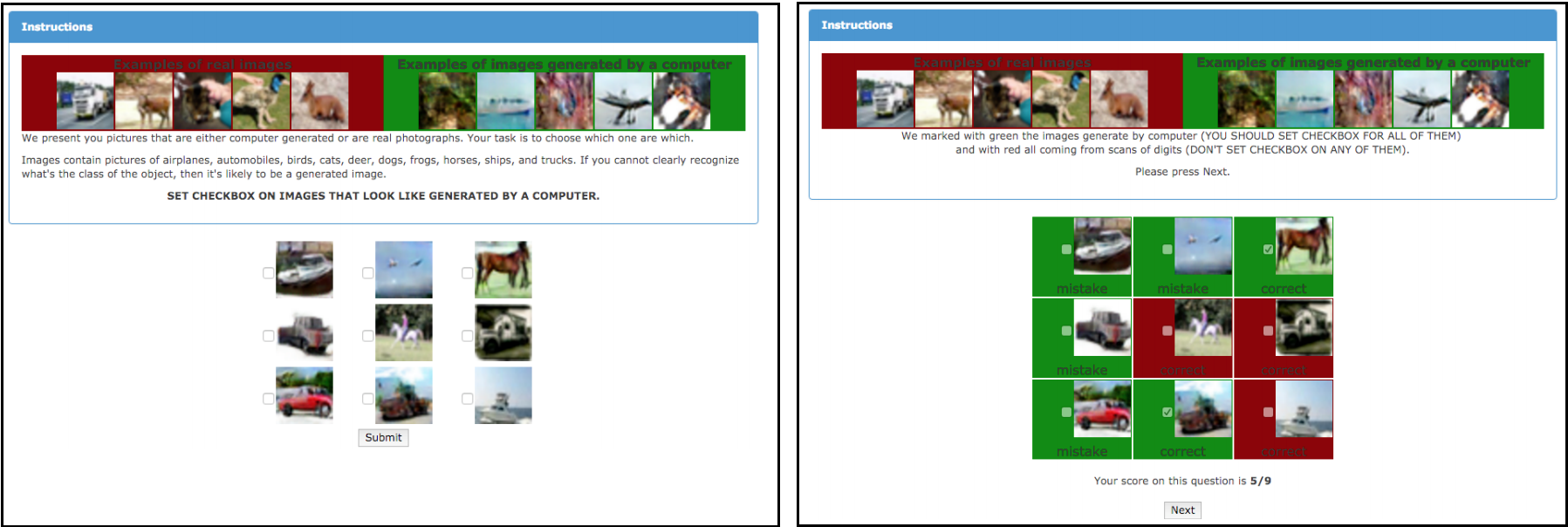}
\caption{Web interface given to annotators in the experiments conducted by Salimans~\etal~\cite{salimans2016improved}. Annotators are asked to distinguish
computer generated images from real ones (left) and are provided with feedback (right). Figure compiled from~\cite{salimans2016improved}.} 
\label{fig:salimas}
\end{figure}

\item {\bf Rating and Preference Judgment.} These types of experiments ask subjects to rate models in terms of the fidelity of their generated images. For example, Snell et al.,~\cite{snell2015learning} 
%From [[Learning to Generate Images With Perceptual Similarity Metrics]]
studied whether observers prefer reconstructions produced by perceptually-optimized networks or by the pixelwise-loss optimized networks. Participants were shown image triplets with the original (reference) image in the center and the SSIM- and MSE-optimized reconstructions on either side with the locations counterbalanced. Participants were instructed to select which of the two reconstructed images they preferred (See Fig.~\ref{fig:Snell}). Similar approaches have been followed in~\cite{huang2017stacked,zhang2017stackgan,xiao2018generating,yi2017dualgan,zhang2016colorful,upchurch2016deep,donahue2017semantically, liu2017auto, lu2017sketch}. Often the first few trials in these experiments are spared for practice.

\begin{figure}[t]
\includegraphics[width=.5\linewidth]{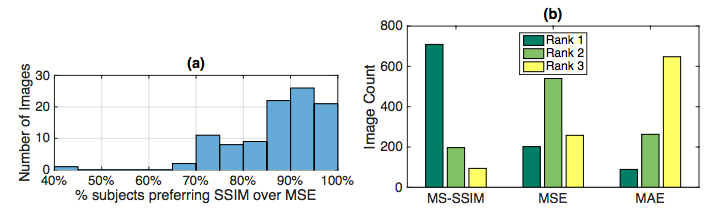}
\includegraphics[width=.5\linewidth]{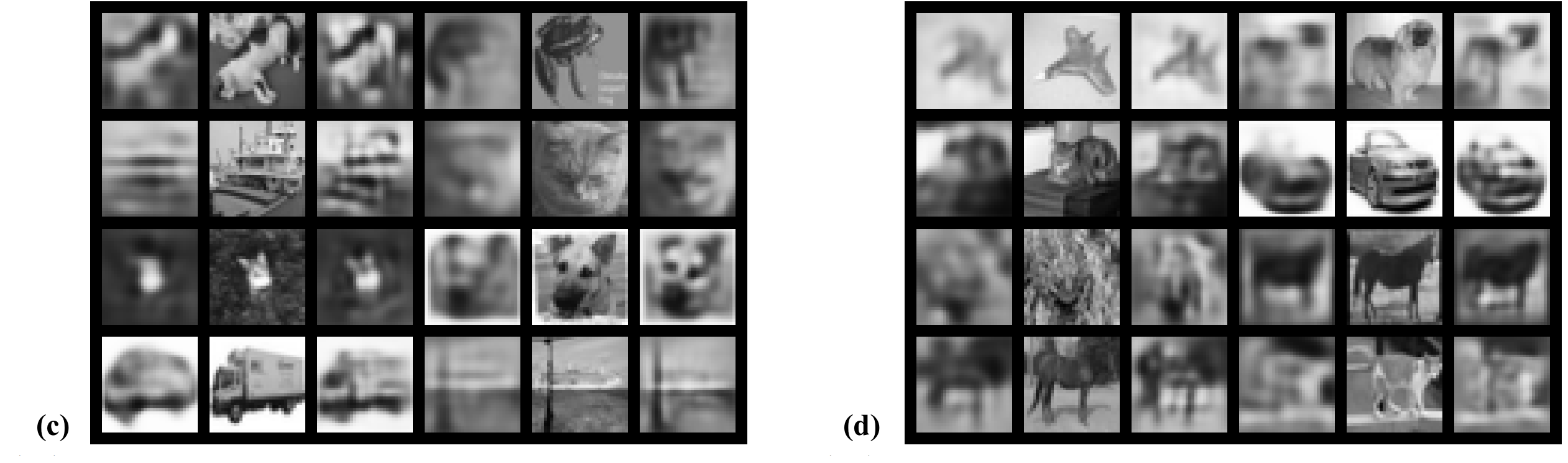}
\caption{An example of a user judgment study by Snell \etal~\cite{snell2015learning}. Left) Human judgments of generated images (a) Fully connected network: Proportion of participants preferring SSIM to
MSE for each of 100 image triplets. (b) Deterministic conv. network:
Distribution of image quality ranking for MS-SSIM, MSE,
and MAE for 1000 images from the STL-10 hold-out set. 
Right) Image triplets consisting of—from left to right—the MSE
reconstruction, the original image, and the SSIM reconstruction.
Image triplets are ordered, from top to bottom and left to right,
by the percentage of participants preferring SSIM. (c) Eight images
for which participants strongly preferred SSIM over MSE.
(d) Eight images for which the smallest proportion of participants
preferred SSIM. Figure compiled from~\cite{snell2015learning}.} 
\label{fig:Snell}
\end{figure}

\item{\bf Evaluating Mode Drop and Mode Collapse.} %https://openreview.net/pdf?id=Sy1f0e-R- 
GANs have been repeatedly criticized for failing to model the entire data distribution, while being able to generate realistically looking images. Mode collapse, \aka the Helvetica scenario, is the phenomenon when the generator learns to map several different input $z$ vectors to the same output (possibly due to low model capacity or inadequate optimization~\cite{arora2017gans}). It causes lack of diversity in the generated samples as the generator assigns low probability mass to significant subsets of the data distribution's support. %, and hence losing some modes. 
%This also includes the phenomenon of trained generators mapping two latent vectors that are far apart to the same or similar data samples. Mode collapse is a byproduct of poor generalization>
%In realistic settings, $P_r$ is usually very diverse since natural images are inherently multimodal. Many have conjectured that $P_g$ differs from $P_r$ by reducing diversity, possibly due to the lack of model capacity or inadequate optimization~\cite{arora2017gans}. This is often manifested itself for generative models in a mix of two ways: 
Mode drop occurs when some hard-to-represent modes of $P_r$ are simply ``ignored'' by $P_g$. This is different than mode collapse where several modes of $P_r$ are ``averaged'' by $P_g$ into a single mode, possibly located at a midpoint. An ideal GAN evaluation measure should be sensitive to these two phenomena. %[[xx et al., showed how these can be quantitavely measured for image generation in high D]].

Detecting mode collapse in GANs trained on large scale image datasets is very challenging\footnote{See~\cite{srivastava2017veegan,huang2018an} for analysis of mode drop and mode collapse over real datasets.}. However, it can be accurately measured on synthetic datasets where the true distribution and its modes are known (\eg~Gaussian mixtures). 
%In this section we compare all four competing methods on three synthetic datasets of increasing difficulty: 
%a mixture of eight 2D Gaussian distributions arranged in a ring, a mixture of twenty-five 
%2D Gaussian distributions arranged in a grid \footnote{Experiment follows \cite{ali2016}. Please note that for certain settings of parameters, vanilla GAN can also recover all 25 modes, as was pointed out to us by Paulina Grnarova.} and a mixture of ten 
%700 dimensional Gaussian distributions embedded in a 1200 dimensional space.  This mixture arrangement was chosen to mimic the higher dimensional manifolds of natural images.
%All of the mixture components were isotropic Gaussians.
%For a fair comparison of the different learning methods
%for GANs, we use the same 
%network architectures for the reconstructors and the generators for all methods,
%namely, fully-connected MLPs with two hidden layers. For the discriminator we use a two layer MLP without  dropout or normalization layers. %\acronym method works for both deterministic and stochastic generator networks. To allow for the generator to be a stochastic map we add an extra dimension of noise to the generator input that is not reconstructed. 
Srivastava \etal~\cite{srivastava2017veegan} proposed a measure to quantify mode collapse behavior as follows:
\begin{enumerate}
\item First, some points are sampled from the generator. A sample is counted as \emph{high quality}, if it is within a certain distance of its nearest mode center (\eg~$3\sigma$ over a 2D dataset, or $10\sigma$ over a 1200D dataset).

\item Then, the \emph{number of modes captured} is the number of mixture components whose mean is nearest to at least one high quality sample. Accordingly, a mode is considered lost if there is no sample in the generated test data within a certain standard deviations from the center of that mode. This is illustrated in Fig.~\ref{fig:mode}. 
\end{enumerate}

%[[from A Classification–Based Study of Covariate Shift in GAN Distributions
Santurkar et al.~\cite{santurkar2018classification}, to investigate mode distribution/collapse over natural datasets, propose to train GANs over a well-balanced dataset (\ie~a dataset that contains equal number of samples from each class) and then test whether generated data also generates a well-balanced dataset. Steps are as follows:
\begin{enumerate}
\item{Train the GAN unconditionally (without class labels) on the chosen balanced multi-class dataset D.}
\item{Train a multi-class classifier on the same dataset D (to be used as an annotator).} 
\item{Generate a synthetic dataset by sampling N images from the GAN. Then use the classifier trained in Step 2 above to obtain labels for this synthetic dataset.}
\end{enumerate}

An example is shown in Fig.~\ref{fig:covariate2}. It reveals that GANs often exhibit mode collapse.

\begin{figure}[t]
\centering
\includegraphics[width=.95\linewidth]{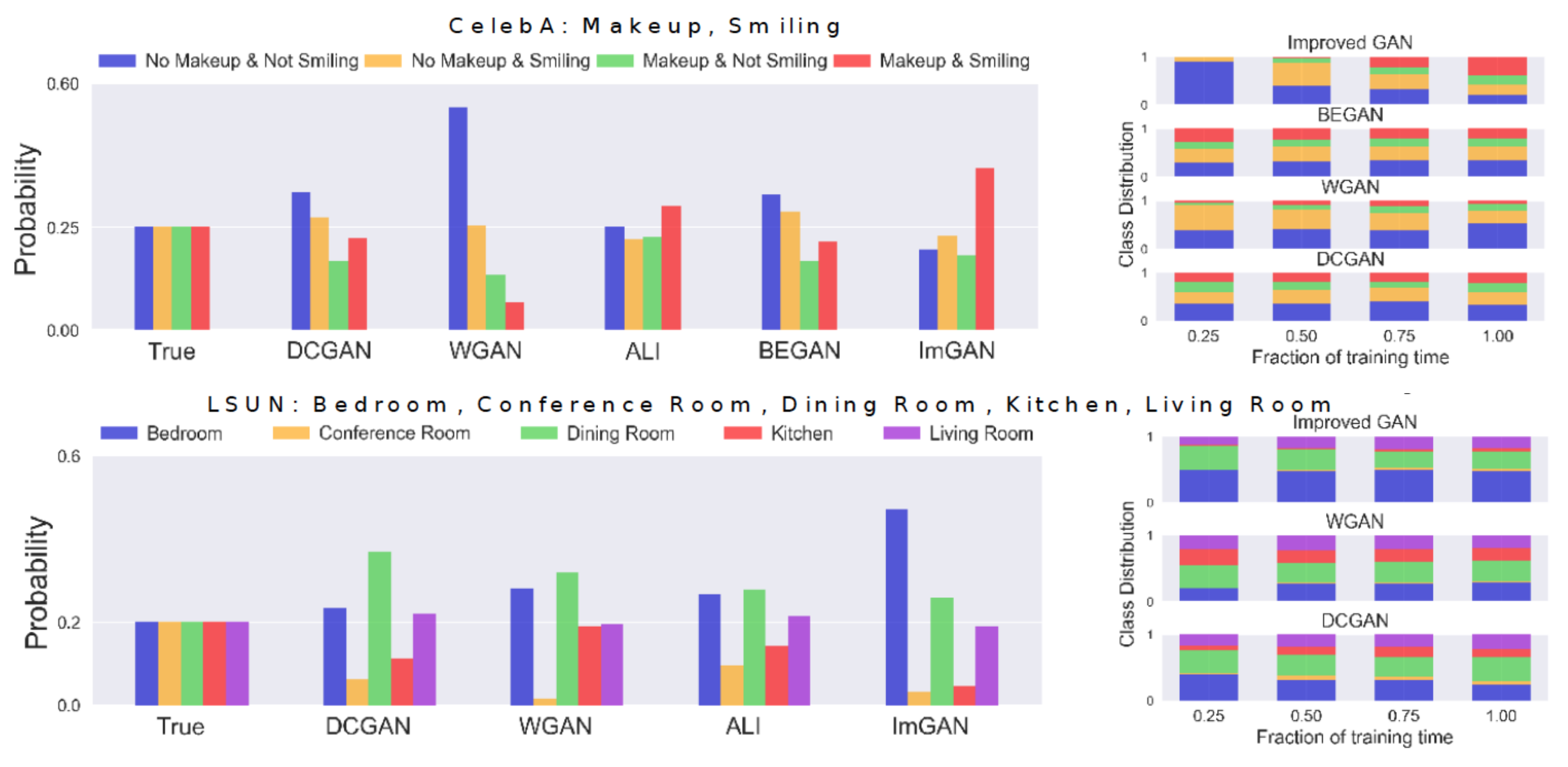}
\caption{Illustration of mode collapse in GANs trained on select subsets of CelebA and LSUN datasets using the technique in~\cite{santurkar2018classification}. Left panel shows the relative distribution of modes in samples drawn from the GANs, and compares is to the true data distribution (leftmost plots). Right panel shows the evolution of class distributions in different GANs over the course of training. It can be seen that these GANs introduce covariate shift through mode collapse. Figure compiled from~\cite{santurkar2018classification}.} 
\label{fig:covariate2}
\end{figure}

The reverse KL divergence over the modes has been used in~\cite{lin2017pacgan} to measure the quality of mode collapse as follows. Each generated sample is assigned to its closest mode. This induces an empirical, discrete
distribution with an alphabet size equal to the number of observed modes in the generated samples.
A similar induced discrete distribution is computed from the real data samples. The reverse KL
divergence between the induced distribution from generated samples and the induced distribution from the real samples is used as a measure.

The shortcoming of the described measures is that they only work for datasets with known modes (\eg~synthetic or labeled datasets). Overall, it is hard to quantitatively measure mode collapse and mode drop since they are poorly understood. Further, finding nearest neighbors and nearest mode center is non-trivial in high dimensional spaces is non-trivial. Active research is ongoing in this direction.

\begin{figure}[t]
\centering
\includegraphics[width=.9\linewidth]{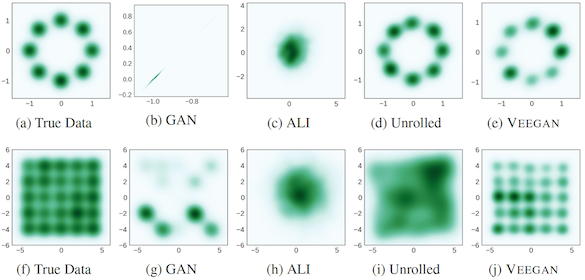}
\caption{Density plots of the true data and generator distributions from different GAN methods
trained on mixtures of Gaussians arranged in a ring (top) or a grid (bottom). Figure from~\cite{srivastava2017veegan}.} 
\label{fig:mode}
\end{figure}

%As shown in \autoref{tab:syn},  VEGAN  captures the greatest number of modes on all the synthetic datasets, while consistently generating higher quality samples.  This is visually apparent in Figure \ref{fig:ring}, which plot the generator distributions for each  method; the generators learned by VEGAN are sharper and closer to the true distribution. This figure also shows why it is important to measure sample quality and mode collapse simultaneously, as either alone can be  misleading. For instance, the GAN on the 2D ring has $99.3\%$ sample quality, but this is simply because the GAN collapses all of its samples onto one mode (Figure \ref{fig:gan_ring}). On the other extreme, the unrolled GAN on the 2D grid captures almost all the modes in  the true distribution, but this is simply because  that it is generating highly dispersed samples (Figure \ref{fig:unrolled_grid2}) that do not accurately represent the true distribution, hence the low sample quality. All methods had approximately the same running time, except for unrolled GAN,
%which is a few orders of magnitude slower due to the unrolling overhead.]]

%[[ from~\cite{che2016mode} MODE REGULARIZED GENERATIVE ADVERSARIAL NETWORKS
%In addition to the metric regularizer, we propose a mode regularizer to further penalize missi
%Png modes. ]]

\item {\bf Investigating and Visualizing the Internals of Networks.}
%  Do GANs actually learn the distribution? An empirical study. 
Other ways of evaluating generative models are studying how and what they learn, exploring their internal dynamics, and  understanding the landscape of their latent spaces. While this is a broad topic and many papers fall under it, here I bring few examples to give the reader some insights. 
%Visual imagination is the act of creating a latent representation of some concept. [To understand the landscape of the latent space.]
%One test checks the similarity of each generated image to the nearest images in the training set. 

\begin{enumerate}

\item \textit{Disentangled representations}. ``Disentanglement'' regards the alignment of ``semantic'' visual concepts to axes in the latent space. Some tests can check the existence of semantically meaningful directions in the latent space, meaning that varying the seed along those directions leads to predictable changes (\eg~changes in facial
hair, or pose). Some others (\eg~\cite{chen2016infogan,higgins2016beta,mathieu2016disentangling,lipton2017precise}) assess the quality of internal representations by checking whether they satisfy certain properties, such as being ``disentangled''. A measure of disentanglement proposed in~\cite{higgins2016beta} checks whether the latent space captures the true factors of variation in a simulated dataset where parameters are known by construction (\eg~using a graphics engine). Radford \etal~\cite{radford2015unsupervised} investigated their trained generators and discriminators in a variety of ways. 
%We do not do any kind of nearest neighbor search on the training set. Nearest neighbors in pixel or feature space are
%trivially fooled (Theis et al., 2015) by small image transforms. We also do not use log-likelihood
%metrics to quantitatively assess the model, as it is a poor (Theis et al., 2015) metric.
They proposed that walking on the learned manifold can tell us about signs of memorization (if there are sharp transitions) and about the way in which the space is hierarchically collapsed. If walking in this latent space
results in semantic changes to the image generations (such as objects being added and removed), one
can reason that the model has learned relevant and interesting representations. They also showed interesting results of performing vector arithmetic on the $z$ vectors of sets of exemplar samples for visual concepts (\eg~smiling woman - neutral woman + neutral man = smiling man; using $z$'s averaged over several samples).

\item \textit{Space continuity}. 
%[[ from DCGAN https://arxiv.org/pdf/1511.06434.pdf
Related to above, the goal here it to study the level of detail a model is capable of extracting. For example, given two random seed vectors $z_1$ and $z_2$ that generated
two realistic images, we can check the images produced using seeds lying on the line joining $z_1$ and $z_2$. If
such “interpolated” images are reasonable and visually appealing, then this may be taken as a sign
that a model can produce novel images rather than simply memorizing them (\eg~\cite{berthelot2017began}; See Fig.~\ref{fig:began}). Some other examples include~\cite{donahue2017semantically,nguyen2016synthesizing}. White~\cite{white2016sampling} suggests that replacing linear interpolation with spherical linear interpolation prevents diverging from a model’s prior distribution and produces sharper samples. Vedantam \etal~\cite{vedantam2017generative} studied ``visually grounded semantic imagination'' and proposed several ways to evaluate their models in terms of the quality of the learned semantic latent space.

\begin{figure}[htbp]
\centering
\includegraphics[width=1\linewidth]{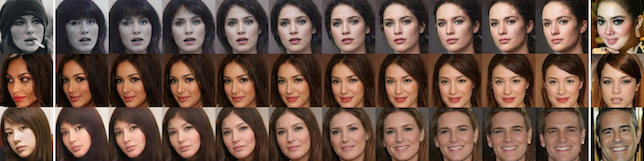}
\includegraphics[width=1\linewidth]{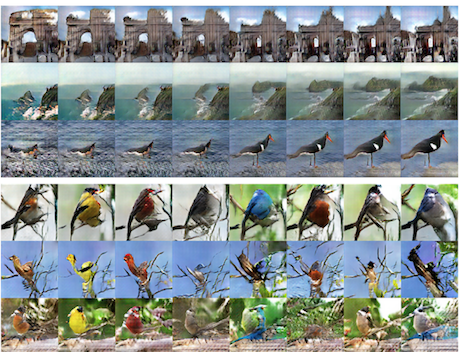}
\caption{Top: Interpolations on $z_r$ between real images at $128 \times 128$ resolution (from BEGAN~\cite{berthelot2017began}). These images
were not part of the training data. The first and last columns contain the real images to be represented
and interpolated. The images immediately next to them are their corresponding approximations
while the images in between are the results of linear interpolation in $z_r$. Middle: Latent space interpolations for three ImageNet classes. Left-most and right-columns show three pairs of image
samples - each pair from a distinct class. Intermediate columns
highlight linear interpolations in the latent space between these
three pairs of images (From~\cite{odena2016conditional}). Bottom: Class-independent information
contains global structure about the synthesized image. Each column
is a distinct bird class while each row corresponds to a fixed
latent code $z$ (From~\cite{odena2016conditional}).} 
\label{fig:began}
\end{figure}

%See also Semantically Decomposing the Latent Spaces of
%Generative Adversarial Networks
%https://arxiv.org/pdf/1705.07904.pdf

\item \textit{Visualizing the discriminator features}. Motivated by previous studies on investigating the representations and features learned by convolutional neural networks trained for scene classification  (\eg~\cite{zeiler2014visualizing,bau2017network,zhou2014object}), some works have attempted to visualize the internal parts of generators and discriminators in GANs.
%Previous work has demonstrated that supervised training of CNNs on large image datasets results in very powerful learned features (Zeiler \& Fergus, 2014). Additionally, supervised CNNs trained on scene classification learn object detectors (Oquab et al., 2014). We demonstrate that an unsupervised
For example, Radford \etal~\cite{radford2015unsupervised} showed that DCGAN trained on a large image dataset can also learn a hierarchy of interesting features. Using guided backpropagation~\cite{springenberg2014striving}, they showed that the features learned by the discriminator fire on typical parts of a bedroom, such as beds and windows (See Fig. 5 in~\cite{radford2015unsupervised}). The t-SNE method~\cite{maaten2008visualizing} has also been frequently used to project the learned latent spaces in 2D.

\end{enumerate}

% \item A QUANTITATIVE MEASURE OF GENERATIVE ADVERSARIAL NETWORK DISTRIBUTIONS  yet another measure 

% https://openreview.net/pdf?id=SJgabgBFl

\end{enumerate}

%[[from Deep Generative Filter for Motion Deblurring.
%Other perceptual metrics
%PSNR (dB) 
%SSIM 
%MS-SSIM 
%F-SIM 
%UIQI 
%IFC 
%VIF 
%Norm-NR ]]

%[[Perceptual Adversarial Networks for Image-to-Image Transformation.
%To illustrate the performance of image-to-image transformation tasks, we used qualitative and quantitative
%metrics to evaluate the performance of the transformed images. For the qualitative experiments, we directly
%showed the input and transformed images. Meanwhile, we used quantitative measures to evaluate the
%performance over the test sets, such as Peak Signal to Noise Ratio (PSNR), Structural Similarity Index
%(SSIM) [42], 

%Universal Quality Index (UQI)~\cite{WangBovik} [41] and Visual Information Fidelity (VIF)~\cite{sheikh2006image}
%%1706.09138.pdf
%%]]

%more xx
%From Schrier

%Another approach:
%Text to Image -> then build a caption -> and then compare the generated caption with the original text!!

%<<Application purpose Evaluation of GANs; like saliency but for a special purpose!!>>

\section{Discussion}
\label{disc}
%[https://arxiv.org/pdf/1705.10762.pdf]

\subsection{Other Evaluation Measures}
In addition to measures discussed above, there exist some other non-trivial or task-specific ways to evaluate GANs. Vedantam \etal~\cite{vedantam2017generative} proposed a model for visually grounded imagination to create images of novel semantic concepts. To evaluate the quality of the generated images, they proposed three measures including \textit{a) correctness:} fraction of attributes for each generated image that match those specified in the concept’s description, \textit{b) coverage:} diversity of values for the unspecified or missing attributes, measured as the difference between the empirical distributions of attribute values in the generated set and the true distribution for this attribute induced by the training set, and \textit{c) compositionality:} correctness
of generated images in response to test concepts that differ in at least one attribute from the training concepts. 
%[[from  Toward Multimodal Image-to-Image Translation~\cite{zhu2017toward}
To measure diversity of generated samples, Zhu \etal~\cite{zhu2017toward} randomly sampled from their model and computed the average pair-wise distance in a deep feature space using cosine distance and compared it with the same measure calculated from ground truth real images. This is akin to the image retrieval performance measure described above. 
%https://openreview.net/pdf?id=SJQHjzZ0-
Im \etal~\cite{jiwoong2018quantitatively} proposed to evaluate GANs by exploring the
divergence and distance measures that were used during training GANs. They showed that rankings produced
by four measures 
%our proposed metrics are consistent and robust across metrics. They used four metrics 
including 1) Jensen-Shannon Divergence, 2) Constrained Pearson $\chi^2$, 3) Maximum Mean Discrepancy, and 4)  Wasserstein Distance, are consistent and robust across measures. 
\begin{table}
\begin{center}
%\hspace{50pt}
\renewcommand{\tabcolsep}{.9mm}
\renewcommand{\arraystretch}{1.2}%

\begin{footnotesize}
\begin{tabular}{ l   ccccccc     | } 
%\hline
%\cline{2-7}
 & \multicolumn{7}{c}{\bf Desiderata } \\ \cline{2-8}
\multicolumn{1}{c|}{\bf Measure } 
 &   \rotatebox[origin=l]{90}{Discriminability} & \rotatebox[origin=l]{90}{Detecting Overfitting} &  \rotatebox[origin=l]{90}{Disentangled Latent Spaces \ }&   \multicolumn{1}{c}{\rotatebox[origin=l]{90}{Well-defined Bounds}} &   \rotatebox[origin=l]{90}{Perceptual Judgments}  & \rotatebox[origin=l]{90}{Sensitivity to Distortions } & \rotatebox[origin=l]{90}{Comp. \& Sample Efficiency }\\ 
%\hline
\hline
%\multirow{18}{*}{\rotatebox{90}{\bf Quantitative Measure}} & Average Log-
\multicolumn{1}{|l}{1. Average Log-
likelihood}~\cite{goodfellow2014generative,theis2015note} & low & low & - & [-$\infty$, $\infty$] & low & low  & low\\
\multicolumn{1}{|l}{2. Coverage Metric}~\cite{tolstikhin2017adagan} & low & low & - & [0, 1] &  low &  low & - \\
\multicolumn{1}{|l}{3. Inception Score (IS)}~\cite{salimans2016improved} & high & moderate & - &  [1, $\infty$] & high & moderate  & high \\
\multicolumn{1}{|l}{4. Modified Inception Score (m-IS)}~\cite{gurumurthy2017deligan} & high & moderate & - &  [1, $\infty$] & high & moderate & high \\
\multicolumn{1}{|l}{5. Mode Score (MS)}~\cite{che2016mode}  & high & moderate & - &  [0, $\infty$] & high & moderate & high \\
\multicolumn{1}{|l}{6. AM Score}~\cite{zhou2018activation}  & high & moderate & - &  [0, $\infty$] & high & moderate  & high\\
\multicolumn{1}{|l}{7. Fr\'echet Inception Distance (FID)}~\cite{heusel2017gans}  & high & moderate & - &  [0, $\infty$] & high & high  & high \\
\multicolumn{1}{|l}{8. Maximum Mean Discrepancy (MMD)}~\cite{gretton2012kernel} & high & low & - &  [0, $\infty$] & - & -  & -\\
\multicolumn{1}{|l}{9. The Wasserstein Critic}~\cite{arjovsky2017wasserstein} & high & moderate & - &  [0, $\infty$] & - & -  & low\\
\multicolumn{1}{|l}{10. Birthday Paradox Test}~\cite{arora2017gans}  & low & high & - &  [1, $\infty$] & low & low  & - \\
\multicolumn{1}{|l}{11. Classifier Two Sample Test (C2ST)}~\cite{lehmann2006testing}  & high & low & - &  [0, 1] & - & -  & - \\
\multicolumn{1}{|l}{12. Classification Performance}~\cite{radford2015unsupervised,isola2017image} & high & low & - &  [0, 1] & low & -  & - \\
\multicolumn{1}{|l}{13. Boundary Distortion}~\cite{santurkar2018classification} & low & low & - &  [0, 1] & - & -  & - \\
\multicolumn{1}{|l}{14. NDB}~\cite{Richardson2018} & low & high & - &  [0, $\infty$] & - & low  & - \\
\multicolumn{1}{|l}{15. Image Retrieval Performance}~\cite{wang2016ensembles}  & moderate & low & - & * & low & -  & - \\
\multicolumn{1}{|l}{16. Generative Adversarial Metric (GAM)}~\cite{im2016generating} & high & low & - & * & - & -  & moderate \\
\multicolumn{1}{|l}{17. Tournament Win Rate and Skill Rating}~\cite{olsson2018skill} & high & high & - & * & - & -  & low \\
\multicolumn{1}{|l}{18. NRDS}~\cite{zhang2018decoupled} & high & low & - &  [0, 1] & - & -  &  poor  \\
\multicolumn{1}{|l}{19. Adversarial Accuracy \& Divergence}  \cite{yang2017lr}  & high & low & - & [0, 1],  [0, $\infty$] & - & -  & - \\
\multicolumn{1}{|l}{20. Geometry Score}~\cite{khrulkov2018geometry} & low & low & - &  [0, $\infty$] & - & low  & low \\
\multicolumn{1}{|l}{21. Reconstruction Error}~\cite{xiang2017effects} & low & low & - &  [0, $\infty$] & - & moderate  & moderate \\
\multicolumn{1}{|l}{22. Image Quality Measures}~\cite{wang2004image,ridgeway2015learning,juefei2017gang} & low & moderate & - & * & high & high  & high\\
\multicolumn{1}{|l}{23. Low-level Image Statistics}~\cite{zeng2017statistics,karras2017progressive}  & low & low & - & * & low & low  & - \\
\multicolumn{1}{|l}{24. Precision, Recall and $F_1$ score}~\cite{lucic2017gans}  & low & high & \cmark  & [0, 1]  & - & -  & -  \\

%& Mode Drop and Collapse~\cite{srivastava2017veegan,lin2017pacgan} & \xmark & \cmark & \xmark & \cmark & \xmark & \xmark ! & \cmark & \xmark & \xmark & \cmark  ! \\
%\hline  
\hline

%\multirow{4}{*}{\rotatebox{90}{\bf Qualit.  }} & Nearest Neighbors  &  & & & & & & &  & +& +\\
%& Rapid Scene Categorization~\cite{goodfellow2014generative}  &   & & & & & & & & +& +\\
%& Preference Judgment~\cite{huang2017stacked,zhang2017stackgan,xiao2018generating,yi2017dualgan} &  & & & & & & &  & +& +\\
%& Network Internals~\cite{radford2015unsupervised,chen2016infogan,higgins2016beta,mathieu2016disentangling,zeiler2014visualizing,bau2017network} &  & & & & & & &  & +& +\\
%\hline
\end{tabular}
\caption{Meta measure of GAN quantitative evaluation scores. Notice that the ratings are relative. ``-'' means unknown (hence warranting further research). ``*'' indicates that several bounds for several scores in that family measure are available. Also, notice that tighter bounds for some of the measures might be possible. It seems that most of the measures do not systematically evaluate disentanglement in the latent space.  } 
%Maybe I can make a table summarizing the pros and cons of model!!
%e.g., needs labels, can it identify mode collapse? 
%How to classify the metrics?! scores? similar to saliency research?!!}
\label{tab:meta}
\end{footnotesize}
\end{center}

\end{table}
%\end{sidewaystable}

\subsection{Sample and Computational Efficiencies}
Here, I provide more details on two items in the list of desired properties of GAN evaluation measures. They will be used in the next subsection for assessing the measures. Huang \etal~\cite{huang2018an} argue that a practical GAN evaluation measure should be computed using a reasonable
number of samples and within an affordable computation cost. This is particularly important during monitoring the training process of models. They proposed the following ways to assess evaluation measures:

\begin{enumerate}
\item \textit{Sample efficiency:} It regards the number of samples needed for a measure to discriminate a set of generated samples $S_g$ from a set of real samples $S^{'}_r$. To do this, a reference set $S_r$ is uniformly sampled from the real training data (but disjoint with $S^{'}_r$). All three sets have
the same size (\ie~$|S_r| = |S^{'}_r| = |S_g| = n$). An ideal measure $\rho$ is expected to correctly score
$\rho(S_r, S^{'}_r)$ lower than $\rho(S_r, S_g)$ with a relatively small $n$. In other words, the number of samples $n$ needed for a measure to distinguish $S^{'}_r$ and $S_g$ can be viewed as its sample complexity. 

\item \textit{Computational efficiency:} Fast computation of the empirical measure is of practical concern as it
helps researchers monitor the training process and diagnose problems early on (\eg~for early
stopping). This can be measured in terms of seconds per number of evaluated samples.

\subsection{What is the Best GAN Evaluation Measure?} 
To answer this question, lets first take a look at how well the measures perform with respect to the desired properties mentioned in Section~\ref{desiderata}. Results are shown in Table~\ref{tab:meta}. I find that:
\begin{enumerate}
\item only two measures are designed to explicitly address overfitting, 
\item the majority of the measures do not consider disentangled representations, 
\item few measures have both lower and upper bounds, 
\item the agreement between the measures and human perceptual judgments is less clear, 
\item several highly regarded measures have high sample and computational efficiencies, and
\item the sensitivity of measures to image distortions is less explored.
\end{enumerate}

A detailed discussion and comparison of GAN evaluation measures comes next. 

As of yet, there is no consensus regarding the best score. Different scores assess various aspects of the image generation process, and it is unlikely that a single score can cover all aspects. Nevertheless, 
%There is no trivial answer to this question and there seems not to be a unique one, although 
some measures seem more plausible than others (\eg~FID score). %[[discussion 
Detailed analyses by Theis \etal~\cite{theis2015note} showed that average likelihood is not a good measure. Parzen windows estimation of likelihood favors trivial models and is irrelevant to visual fidelity of samples. Further, it fails to approximate the true likelihood in high dimensional spaces or to rank models (Fig.~\ref{fig:theis}). Similarly, the Wasserstein distance between generated samples and the training data is also intractable in high dimensions~\cite{karras2017progressive}. Two widely accepted scores, Inception Score and Fr\'echet Inception Distance, rely on pre-trained deep networks to represent and statistically compare original and generated samples. This brings along two significant drawbacks. First, the deep network is trained to be invariant to image transformations and artifacts making the evaluation method also insensitive to those distortions. Second, since the deep network is often trained on large scale natural scene datasets (\eg~ImageNet), applying them to other domains (\eg~faces, digits) is questionable. Some evaluation methods (\eg~MS-SSIM~\cite{odena2016conditional}, Birthday Paradox Test) aim to assess the diversity of the generated samples, regardless of the data distribution. While being able to detect severe
cases of mode collapse, these methods fall short in measuring how well a generator captures the true data distribution~\cite{karras2017progressive}.

%We stress that the diversity should only be taken into account together with the FID and IS metrics. 
%\cite{kurach2018gan}

%[[From this paper we have:  https://arxiv.org/pdf/1806.07755.pdf
%Most existing works attempt to justify their proposed metrics by showing a strong correlation with human evaluation (Salimans et al., 2016; Lopez-Paz & Oquab, 2016). However, human evaluation tends to be biased towards the visual quality of generated samples and neglect the overall distributional characteristics, which are important for unsupervised learning. [[put under discussion!!!]]
%]]

Quality measures such as nearest neighbor visualizations or rapid categorization tasks may favor models that overfit. Overall, it seems that the main challenge is to have a measure that evaluates both \textit{diversity} and \textit{visual fidelity} simultaneously. The former implies that all modes are covered while the latter implies that the generated samples should have high likelihood. Perhaps due to these challenges, Theis \etal~\cite{theis2015note} argued against evaluating models for task-independent image generation and proposed to evaluate GANs with respect to specific applications. For different applications then, different measures might be more appropriate. 
%extrapolation from one criterion to another is not warranted and 
%generative models need to be evaluated directly with respect to the application(s) they were intended. 
For example, the likelihood is good for measuring compression methods~\cite{theis2017lossy} while psychophysics and user ratings are fit for evaluating image reconstruction and synthesis methods~\cite{ledig2016photo,gerhard2013sensitive}. Some measures are suitable for evaluating generic GANs (when input is a noise vector), while some others are suitable for evaluating conditional GANs (\eg~FCN score) where correspondences are available (\eg~generating an image corresponding to a segmentation map).

Despite having different formulations, several scores are based on similar concepts. C2ST, adversarial accuracy, and classification performance employ classifiers to determine how separable generated images are from real images (on a validation dataset).  %to distinguish generated samples and real data from a validation set. 
FID, Wasserstein and MMD measures compute the distance between two distributions. Inception score and its variants including m-IS, Mode and AM scores use conditional and marginal distributions over generated data or real data to evaluate diversity and fidelity of samples. Average log likelihood and coverage metric estimate the probability distributions. Reconstruction error and some quality measures determine how dissimilar generated images are from their corresponding (or closest) images in the train set. Some measures use individual samples (\eg~IS) while others need pairs of samples (\eg~MMD). One important concern regarding many measures is that they are sensitive to the choice of the feature space (\eg~different CNNs) as well as the distance type (\eg~$L_2$ vs. $L_1$).

Fidelity, diversity and controllable sampling are the main aspects of a model that a measure should capture.
A good score should have well defined bounds and also be sensitive to image distortions and transformations (See Fig.~\ref{fig:robustness} and \ref{fig:FID}). One major problem with qualitative measures such as SSIM and PSNR is that they only tap visual fidelity and not diversity of samples. Humans are also often biased towards the visual quality of generated images and are less affected by the lack of image diversity. 
%neglect the overall distributional characteristics, which are arguably just as important for unsupervised learning. This also applies to some quantitative measures such as SSIM, and PSNR. 
On the other hand, some quantitative measures mostly concentrate on evaluating diversity (\eg~Birthday Paradox Test) and discard fidelity. Ideally, a good measure should take both into account.

%\item {\bf Sample and Runtime Efficiency.}  
% https://openreview.net/pdf?id=Sy1f0e-R-

\begin{figure}[t]
\includegraphics[width=\linewidth]{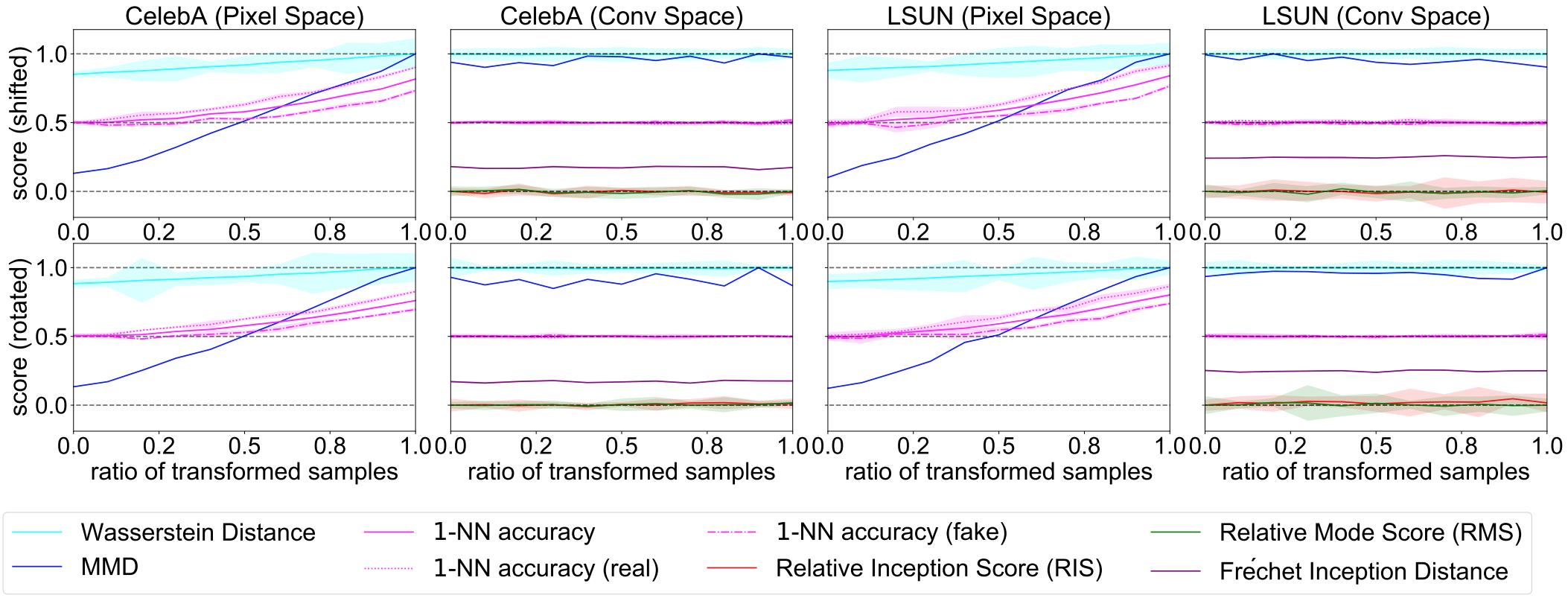}
\caption{Robustness analysis of different GAN evaluation measures to small image transformations (rotations and translations). A good measure is expected to remain constant across all mixes of real and transformed real samples, since the transformations do not alter semantics of the image. Some measures are more susceptible to changes in the pixel space than the convolutional space. Figure from~\cite{huang2018an}.} 
\label{fig:robustness}
\end{figure}

Fig.~\ref{fig:runtime} shows a comparison of GAN evaluation measures in terms of sample and computational efficiency. 
While some measures are practical to compute for a small sample size (about 2000 images), some others (\eg Wasserstein distance) do not scale to large sample sizes. Please see~\cite{huang2018an} for further details.

 % for an example from~\cite{huang2018an}).

\end{enumerate}

\begin{figure}[t]
\centering
\includegraphics[width=.6\linewidth]{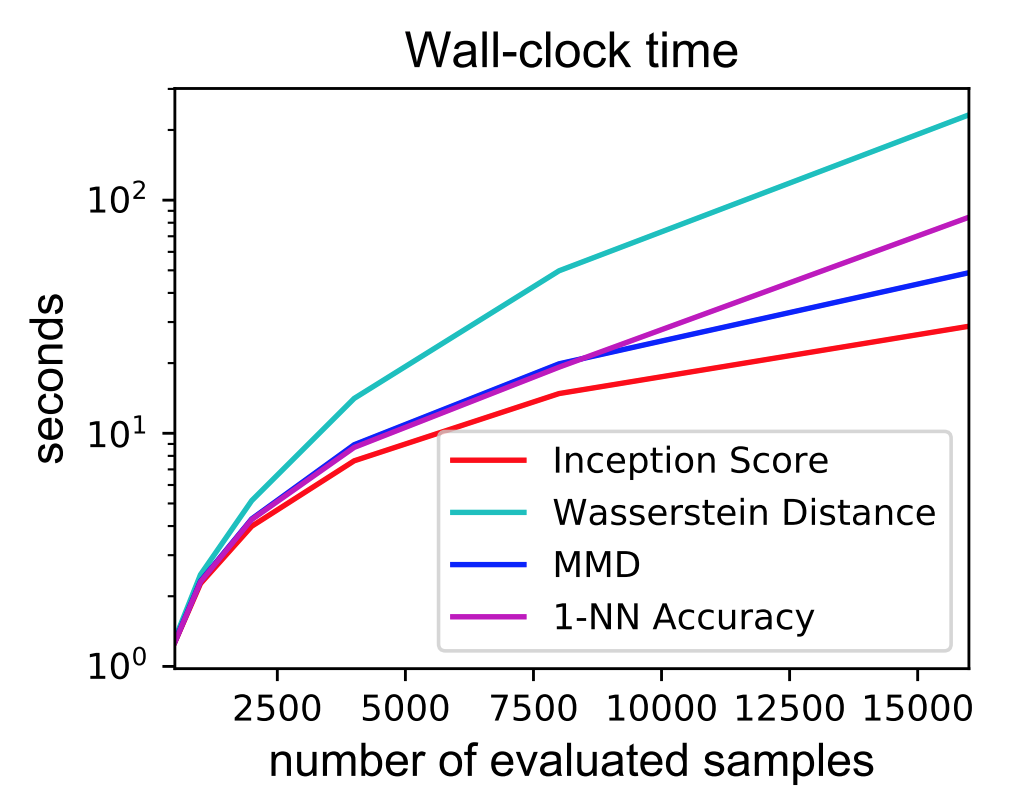}
\caption{Measurement of wall-clock time for computing various measures as a function
of the number of samples. As it shows, all measures are practical to compute for a sample of size
2000, but Wasserstein distance does not scale to large sample sizes. Figure from~\cite{huang2018an}.} 
\label{fig:runtime}
\end{figure}

\section{Summary and Future Work} 

In this work, I provided a critical review of the strengths and limitations of 24 quantitative and 5 qualitative measures that have been introduced so far for evaluating GANs. Seeking appropriate measures for this purpose continues to be an important open problem, not only for fair model comparison but also for understanding, improving, and developing generative models. Lack of a universal powerful measure can hinder the progress. In a recent benchmark study, Lucic \etal~\cite{lucic2017gans} found no empirical evidence in favor of GAN models who claimed superiority over the original GAN. In this regard, borrowing from other fields such as natural scene statistics and cognitive vision can be rewarding.
For example, understanding how humans perceive symmetry~\cite{driver1992preserved,funk2017beyond} or image clutter~\cite{rosenholtz2007measuring} in generated images versus natural scenes can give clues regarding the plausibility of the generated images.

%[[[[add this to the discussion]]
%We believe that truly measuring the distribution learning capabilities of GANs, in particular the learned diversity, inherently requires using multiple metrics. Given that we aim to capture some very complex behaviour in terms of simple, concise metrics, each of these individually will inevitably conflate some aspects. A GAN could, for instance, perform well on the proposed classification-based tests if it just memorized the training set. However, based on prior work, it is unlikely that state-of-theart GANs have this deficiency as is discussed in Section 1.
%Our examination shows that mode collapse is indeed a prevalent issue for current GANs. Also, we observe that GANgenerated datasets have significantly reduced diversity in the periphery of the support.
%]]

%
%} 
%
%
%For example, while many algorithms have claimed superiority over the original GAN model, xx \etal found that 

%no empirical evidence which supports such claims, across all data sets. In fact, the NS GAN performs on par with most other models and achieves the best overall FID on MNIST.

%to encourage more evaluation studies in this area. 

Ultimately, I suggest the following directions for future research in this area: 
\begin{enumerate}
\item creating a code repository of evaluation measures, 
\item conducting detailed comparative empirical and analytical studies of available measures, and 
\item benchmarking models under the same conditions (\eg~architectures, optimization, hyperparameters, computational budget) using more than one measure. 
\end{enumerate}

\bibliography{refs}
\bibliographystyle{abbrv}

\end{document}